\documentclass[11pt]{article}

\usepackage[preprint]{acl}
\usepackage{float}
\usepackage{fancyhdr}

\usepackage{times}
\usepackage{latexsym}
\usepackage[T1]{fontenc}
\usepackage[utf8]{inputenc}
\usepackage{microtype}
\usepackage{inconsolata}
\usepackage{graphicx}
\graphicspath{{asset/images/}}

\usepackage{amsmath,amssymb}
\usepackage{booktabs}
\usepackage{multirow}
\usepackage{xcolor}
\definecolor{PosGreen}{HTML}{1A7F37}   % MOOR positive delta highlight
\definecolor{NegRed}{HTML}{C50000}     % MOOR negative delta highlight
\definecolor{DropL1}{HTML}{FEF6F6}     % drop 0–5pt   (near-white pink)
\definecolor{DropL2}{HTML}{FCE5E5}     % drop 5–10pt  (very light pink)
\definecolor{DropL3}{HTML}{F7CACA}     % drop 10–15pt (light pink)
\definecolor{DropL4}{HTML}{EFB0B0}     % drop ≥15pt   (light salmon — black text remains readable)
\usepackage{enumitem}                % \begin{itemize}[leftmargin=…]
\usepackage{bm}                      % \bm{} bold math
\usepackage{gensymb}                 % \degree in corruption table
\usepackage{pifont}                  % \cmark / \xmark
\newcommand{\cmark}{\ding{51}}
\newcommand{\xmark}{\ding{55}}
\usepackage{listings}                % prompt / code boxes in appendix
\usepackage[skins,breakable]{tcolorbox} % styled prompt boxes
\usepackage{longtable}               % multi-page tables (Tab.8 ablation)
\usepackage{makecell}                % multi-line table headers
\usepackage{subcaption}              % subfigure captions in appendix
\usepackage{titletoc}                % appendix mini-TOC (\startcontents, \printcontents)
\usepackage{array}                   % column spec extensions in appendix tables
\usepackage{colortbl}                % \cellcolor in appendix corruption table
\usepackage[capitalise,noabbrev]{cleveref}  % \Cref{}
\usepackage{placeins}                % \FloatBarrier in appendix
\usepackage{float}                 % [H] to pin floats with their text

\usepackage{adjustbox}
\newcommand{\benchname}{SnapBench}

\title{SnapBench: Benchmarking Snap-and-Ask Multimodal Retrieval for Mobile Interactions}

\author{%
  Zirong Chen\textsuperscript{1,2,3},
  Fuda Ye\textsuperscript{1},
  Kuan Zhang\textsuperscript{3},
  Enjun Du\textsuperscript{1,2,5},
  Junfu Pu\textsuperscript{4},
  Xinlei Wang\textsuperscript{2},
  Xinyu Zuo\textsuperscript{2},
  Lisheng Duan\textsuperscript{2},
  Jin Ma\textsuperscript{2},
  Yongqi Zhang\textsuperscript{1}
}

\begin{document}
\pagestyle{fancy}
\fancyhf{}
\lhead{\small Yuanbao Technical Report}
\rhead{\small Tencent}
\cfoot{\thepage}
\setlength{\headheight}{14pt}
\thispagestyle{fancy}

\twocolumn[{
\begin{minipage}{\textwidth}
  \noindent
  \includegraphics[width=0.18\linewidth]{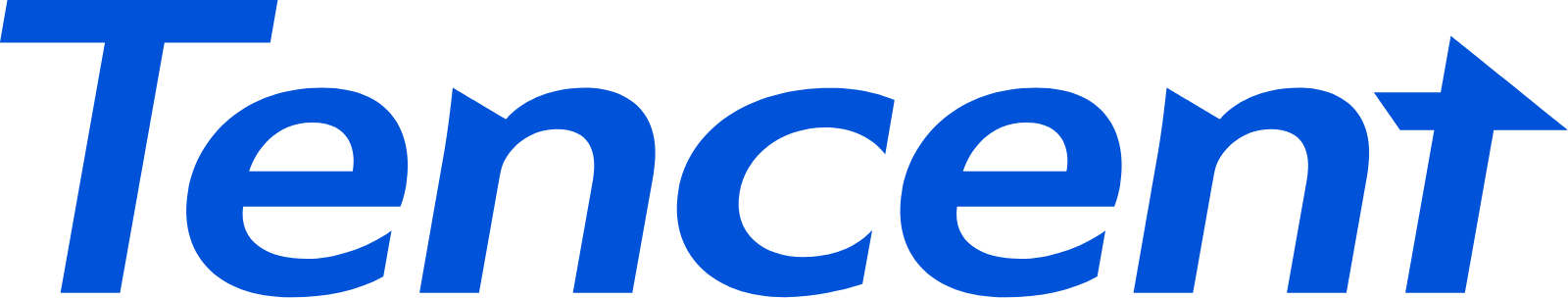}
  \hfill
  \includegraphics[height=1.25cm]{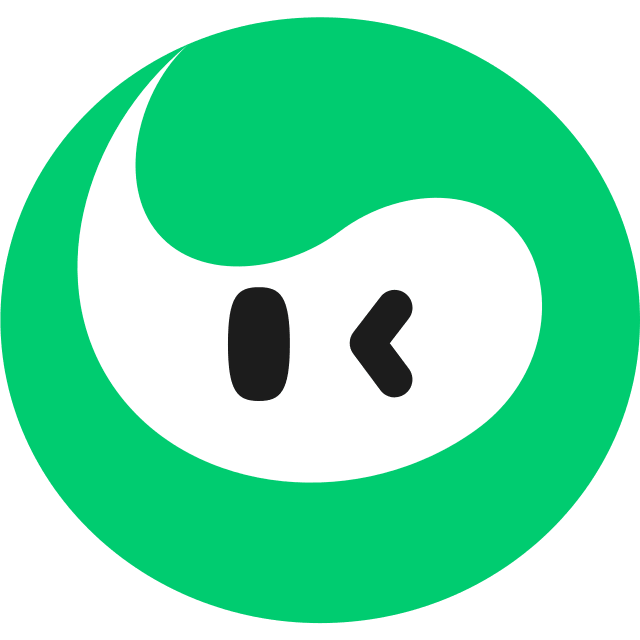}

  \vspace{0.45cm}
  \begin{center}
  {\Huge\bfseries
  SnapBench: Benchmarking Snap-and-Ask\\[0.2em]
  Multimodal Retrieval for Mobile Interactions\par}

  \vspace{0.5cm}
  {\large
  Zirong Chen\textsuperscript{1,2,3,\textdagger,\textdaggerdbl} \quad
  Fuda Ye\textsuperscript{1,\textdaggerdbl} \quad
  Kuan Zhang\textsuperscript{3} \quad
  Enjun Du\textsuperscript{1,2,5,\textdagger} \quad
  Junfu Pu\textsuperscript{4}\\[0.25em]
  Xinlei Wang\textsuperscript{2} \quad
  Xinyu Zuo\textsuperscript{2} \quad
  Lisheng Duan\textsuperscript{2} \quad
  Jin Ma\textsuperscript{2} \quad
  Yongqi Zhang\textsuperscript{1,*}\par}

  \vspace{0.3cm}
  {\small
  \textsuperscript{1}The Hong Kong University of Science and Technology (Guangzhou)\\[0.1em]
  \textsuperscript{2}Tencent Yuanbao \qquad
  \textsuperscript{3}Tsinghua University \qquad
  \textsuperscript{4}ARC Lab, Tencent \qquad
  \textsuperscript{5}The University of Hong Kong\\[0.18em]

  \texttt{\href{mailto:imzrchen@gmail.com}{imzrchen@gmail.com}, \href{mailto:yongqizhang@hkust-gz.edu.cn}{yongqizhang@hkust-gz.edu.cn}}\par}
  \end{center}
  \vspace{0.25cm}
\end{minipage}
}]
\thispagestyle{fancy}

% After the full-width title block, page-1's left column is packed
% (abstract + intro), so a normal \footnotetext has no room there and
% LaTeX dumps the notes under the right column. Pin them to the left
% column with a bottom float instead.
\begin{figure}[b]
\footnotesize\raggedright
\setlength{\parskip}{0.2ex}
\noindent\rule{0.4\columnwidth}{0.4pt}\\[0.35em]
{\textdagger}Work done during an internship at Tencent.\\
{\textdaggerdbl}Zirong Chen and Fuda Ye contributed equally.\\
{*}Corresponding author.
\vspace{-0.6em}
\end{figure}

\begin{abstract}

Mobile AI acts as a visual oracle, empowering users to \textbf{\textit{snap}} a picture of something and \textbf{\textit{ask}} for information. Snap-and-ask retrieval is now one of the most common entry points for mobile AI, yet photos are often blurry, while text questions may be short or mistyped. Existing benchmarks only test on clean inputs or do not isolate paired robustness in snap-and-ask retrieval. Therefore, we introduce SnapBench, the first paired benchmark for robust snap-and-ask multimodal retrieval, spanning $1{,}145$ queries, $9{,}085$ gallery items under 53 controlled corruption conditions with human annotations. We evaluate 16 multimodal retrievers, covering dual-tower encoders and embedding-based VLMs. Results show that image corruptions substantially degrade retrieval, while text corruptions mainly affect text-only retrieval and have limited impact on joint retrieval. Clean image-only retrieval often outperforms joint retrieval, indicating the coarse-text drag and the lack of cross-modal fallback under noisy inputs. SnapBench provides a controlled testbed for evaluating robust retrieval in snap-and-ask scenarios. We further propose MOOR (\textbf{M}odality-anchored, \textbf{O}utlier-aware, \textbf{O}ptimal \textbf{R}eweighting), a simple adaptive fusion approach, highlighting the need for reliability-aware modality calibration in snap-and-ask retrieval. The code and dataset are available at \url{https://github.com/zrchen03/SnapBench}.\looseness-1

% Mobile AI acts as a visual oracle, empowering users to snap a picture of something and ask for information. Snap-and-ask retrieval is now one of the most common entry points for mobile AI, yet photos and text questions are usually blurry and mistyped. Existing benchmarks only test on clean inputs or do not isolate paired robustness in snap-and-ask retrieval. Therefore, we introduce SnapBench, the first paired benchmark for robust snap-and-ask multimodal retrieval, spanning 1,145 queries, 9,085 gallery items under 53 controlled corruption conditions with human annotations. We evaluate 16 multimodal retrievers, covering dual-tower encoders and embedding-based VLMs. Results show that image corruptions substantially degrade retrieval, while text corruptions mainly affect text-only retrieval and have limited impact on joint retrieval. Clean image-only retrieval often outperforms joint retrieval, indicating the coarse-text drag and the lack of cross-modal fallback under noisy inputs. SnapBench provides a controlled testbed for evaluating robust retrieval in snap-and-ask scenarios. We further propose MOOR (Modality-anchored, Outlier-aware, Optimal Reweighting), a simple adaptive fusion approach, highlighting the need for reliability-aware modality calibration in snap-and-ask retrieval. The code and dataset are available at https://anonymous.4open.science/r/SnapBench/.

\end{abstract}

\section{Introduction}
\label{sec:intro}
\begin{figure}[t]
  \centering
  \includegraphics[width=1\linewidth]{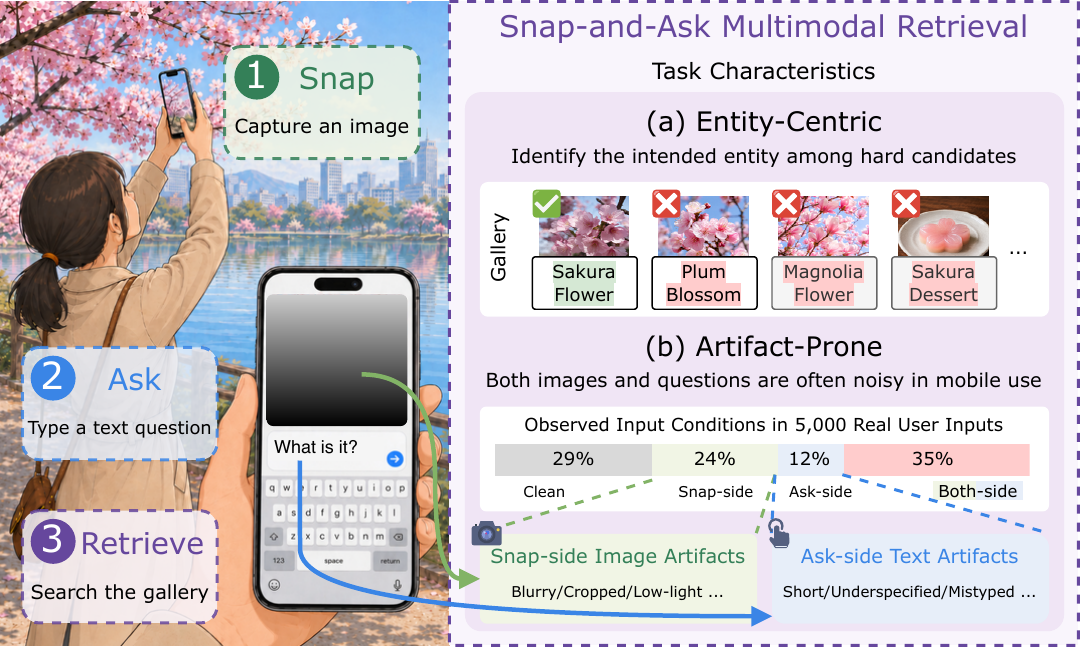}
  \caption{
    Illustration of the mobile snap-and-ask retrieval scenario.
    Given a snapped image and a short text question, the system retrieves items from a gallery. The pilot analysis setting is detailed in Appendix~\ref{sec:appendix_prevalence}.
    }
    
    \label{fig:snap_ask_overview}
\end{figure}

Multimodal AI assistants are increasingly used as mobile visual search interfaces. In a snap-and-ask interaction, a user points a phone camera at an object and asks a short question about it~\cite{Fan2005PhototosearchUM, Chu2024MobileVLMVF}. We study the retrieval step behind this interaction: given the snapped image and text question, the system must rank gallery items that match the target entity intended by the user, i.e., the object identity or fine-grained category to be retrieved from the gallery. This abstraction focuses on visual-language grounding: whether a model can surface the intended target from visually and semantically related candidates~\cite{Wang2011JIGSAWIM, Zhang2012SnapAsk}.

Snap-and-ask retrieval differs from standard multimodal retrieval in two ways.
First, it is \textbf{entity-centric}. As shown in Figure~\ref{fig:snap_ask_overview}, a question such as ``What is it?'' specifies only a coarse intent; it does not by itself distinguish sakura, magnolia, plum blossom, or other hard candidates. The snapped image carries the discriminative evidence, and success depends on ranking the intended entity above hard negatives, not merely matching the image to a generic caption~\cite{Fan2005PhototosearchUM, Weyand2020GoogleLD}. 
Second, the setting is \textbf{artifact-prone}. Snapped photos may be blurry, cropped, occluded, watermarked, or affected by interface overlays, while mobile questions are often brief, category-level, underspecified, or mistyped. These artifacts are intrinsic to everyday mobile use rather than rare edge cases~\cite{Chiu2020AssessingIQ}. The two modalities are therefore complementary but unevenly reliable. A pilot analysis of real mobile uploads (\Cref{fig:snap_ask_overview}) confirms such artifacts are common in practice, motivating controlled evaluation.\looseness-1

Existing benchmarks miss either the query form or the paired robustness requirement. Image-caption retrieval datasets evaluate clean caption-style alignment rather than short questions grounded in user-captured images~\cite{lin2014microsoft, Plummer2015Flickr30kEC, Sharma2018ConceptualCA}. Composed retrieval benchmarks use text as a modification, feedback signal, or similarity constraint over a reference image; snap-and-ask queries instead ask about the snapped entity itself~\cite{wu2021fashion, liu2021image}. Robustness benchmarks perturb images or text, but they rarely keep the target entity, gallery, and labels fixed across clean and corrupted inputs~\cite{hendrycks2019benchmarking, Morris2020TextAttackAF, Qiu2022BenchmarkingRO}. Thus, they cannot isolate how a mobile artifact changes retrieval behavior for the same snap-and-ask intent.

We introduce SnapBench, a paired benchmark for entity-centric snap-and-ask retrieval under mobile interactions. Its design follows a simple principle: clean and corrupted inputs should differ only in the observed input condition, not in the retrieval task being evaluated. Each clean instance contains a snapped image, a user question, and an entity-aware gallery containing positives and hard negatives. We then instantiate controlled image-side and text-side artifacts while preserving the same retrieval intent, ground-truth positives, and gallery. This paired design makes performance changes attributable to the artifact itself rather than to changes in the target entity, candidate pool, or annotation set. SnapBench contains $1{,}145$ queries and $9{,}085$ gallery items with human annotations, covering 53 controlled corruption conditions: 45 image-side conditions and 8 text-side conditions. We further evaluate representative joint image-text artifact settings to test whether single-modality robustness transfers to jointly degraded inputs.\looseness-1

Using SnapBench, we evaluate 16 vision-language retrieval models, including dual-tower encoders and embedding-based VLMs. The evaluation reveals three consistent failures. First, image artifacts cause the largest degradation, especially when they remove or obscure entity-level evidence; clean retrieval accuracy does not reliably predict robustness to mobile capture conditions. Second, text artifacts have limited effect in joint image-text retrieval, but this apparent robustness is misleading: text-only retrieval remains sensitive to text corruption, while many joint systems underuse the text question. More surprisingly, adding the text question can hurt retrieval. Coarse questions such as ``What flower is this?'' raise the scores of many same-category candidates and dilute a strong visual ranking, a failure mode we term coarse-text drag. Third, joint image-text artifacts are non-additive: robustness under image-only and text-only perturbations does not reliably predict robustness when both modalities are corrupted. These findings show that snap-and-ask robustness cannot be assessed by clean retrieval accuracy or by independent single-modality stress tests alone.

These failures expose a modality-calibration bottleneck: the reliable signal varies by query, but fixed fusion cannot adapt. We therefore introduce MOOR (\textbf{M}odality-anchored, \textbf{O}utlier-aware, \textbf{O}ptimal \textbf{R}eweighting), a lightweight adaptive fusion baseline that estimates query signal reliability from the retrieval score distributions and reweights modality paths without modifying the architecture. MOOR serves as a diagnostic probe for fixed-fusion failures: its gains suggest that user text is not inherently harmful, but that current retrieval systems lack reliable calibration for deciding when text should guide, refine, or be ignored relative to vision.\looseness-1

Overall, our contributions are as follows:
\begin{itemize}[nosep,leftmargin=*]
    \item We introduce SnapBench, a paired benchmark for mobile snap-and-ask multimodal retrieval. SnapBench combines entity-centric user questions, dense same-category hard negatives, clean-corrupted retrieval comparisons, and controlled image-side and text-side artifacts.
    \item We provide a systematic evaluation of 16 vision-language retrieval models and identify key failure modes in snap-and-ask retrieval, including image-artifact sensitivity, text underuse, coarse-text drag, and non-additive joint artifact effects.
    \item We propose MOOR, a simple adaptive fusion approach that estimates modality reliability from retrieval scores. It improves over fixed fusion and demonstrates the importance of modality calibration for robust snap-and-ask retrieval.
\end{itemize}

The remainder of this paper is organized as follows.
Section~\ref{sec:related} reviews related work.
Section~\ref{sec:bench} introduces SnapBench.
Section~\ref{sec:exp} presents the main experimental findings.
Section~\ref{sec:method} describes MOOR, and Section~\ref{sec:conclusion} concludes.

\section{Related Work}
\label{sec:related}
\subsection{Multimodal Retrieval Benchmarks}

Image--text retrieval is commonly evaluated on MS COCO~\cite{lin2014microsoft}, Flickr30K~\cite{Plummer2015Flickr30kEC}, and Conceptual Captions~\cite{Sharma2018ConceptualCA}, which assess cross-modal alignment through paired captions or alt text. Despite their broad adoption, these benchmarks do not evaluate whether a model can infer the intended entity from a short and underspecified text question grounded in a user-captured image. More fine-grained datasets, such as Google Landmarks Dataset v2~\cite{Weyand2020GoogleLD} and OVEN~\cite{Hu2023OpenDomainVE}, focus on instance-level or open-domain visual entity recognition. However, they are not designed for mobile snap-and-ask retrieval or for paired image--text corruptions. Composed retrieval benchmarks, including FashionIQ~\cite{wu2021fashion}, CIRR~\cite{liu2021image}, CIRCO~\cite{Baldrati2023ZeroShotCI}, and GeneCIS~\cite{Vaze2023GeneCIS}, retrieve target images using a reference image together with textual feedback or conditions. These tasks require multimodal reasoning, but their textual inputs typically describe modifications or similarity constraints~\cite{Song2025CIRSurvey}. In contrast, snap-and-ask text questions are brief and specific to the snapped entity itself.

\subsection{Robustness in Snap-and-Ask Retrieval}

Mobile visual search enables users to express intent through camera input, often combined with short textual or spoken queries~\cite{Fan2005PhototosearchUM,Wang2011JIGSAWIM}, as well as in visual e-commerce search~\cite{Dagan2021AnII}. Real-world visual question answering datasets, such as VizWiz~\cite{Gurari2018VizWizGC} and InfoSeek~\cite{Chen2023CanPV}, further show that user-captured images are often noisy, and that user text questions are frequently conversational, underspecified, and grounded in visual entities. However, these datasets primarily focus on answer generation rather than retrieving the intended entity from a gallery. Robustness benchmarks examine visual corruptions, textual corruptions, and multimodal distribution shifts~\cite{hendrycks2019benchmarking,Morris2020TextAttackAF,Qiu2022BenchmarkingRO}, but they typically do not keep the target entity and gallery fixed when comparing clean and corrupted image--text pairs. In contrast, \benchname{} introduces a paired protocol that isolates the effects of realistic image and text artifacts on snap-and-ask retrieval.

\Cref{tab:related_work_positioning} positions \benchname{} against two representative single-modality robustness benchmarks. Two properties are distinct to \benchname{}. First, the paired protocol: because the target entity, gallery, and labels stay fixed across clean and corrupted conditions, every score delta cleanly measures corruption impact, whereas ImageNet-C~\cite{hendrycks2019benchmarking} and TextAttack~\cite{Morris2020TextAttackAF} swap in a new test instance per corruption, conflating label difficulty with corruption sensitivity. Second, only \benchname{} corrupts both modalities jointly, which is what makes the super-additivity finding in \Cref{sec:exp_joint_artifacts} measurable at all---no single-modality benchmark could reveal it.

\subsection{Multimodal Fusion and Re-ranking}

A complementary line of work improves multimodal retrieval quality by refining how candidates are combined or re-ordered after initial encoding. EviRank~\citep{du2026evirank} re-ranks image candidates using structured relevance evidence extracted as an explicit intermediate signal. MOOR (\Cref{sec:method_overview}) targets a related but distinct problem surfaced by \benchname{}: rather than introducing additional evidence for re-ranking, it adaptively reweights the four native similarity paths already produced by a single frozen encoder, using only rank consistency and score variance, without extracting extra evidence or learning any parameters.

\begin{table}[t]
\centering\small
\setlength{\tabcolsep}{4pt}

\resizebox{\linewidth}{!}{%
\begin{tabular}{lccc}
\toprule
\textbf{Dimension} & \textbf{ImageNet-C} & \textbf{TextAttack} & \textbf{\benchname{}} \\
\midrule
Modality corrupted & image only & text only & \makecell{image, text,\\or both} \\
Paired protocol & \xmark & \xmark & \cmark \\
Joint multimodal corruption & \xmark & \xmark & \cmark \\
\bottomrule
\end{tabular}%
}
\caption{\benchname{} against existing robustness benchmarks. \emph{Paired}: target entity, gallery, and labels are fixed across clean and corrupted conditions.}
\label{tab:related_work_positioning}
\end{table}

\section{\benchname{}}
\label{sec:bench}

\begin{figure*}[t]
  \centering
  \includegraphics[width=1\textwidth]{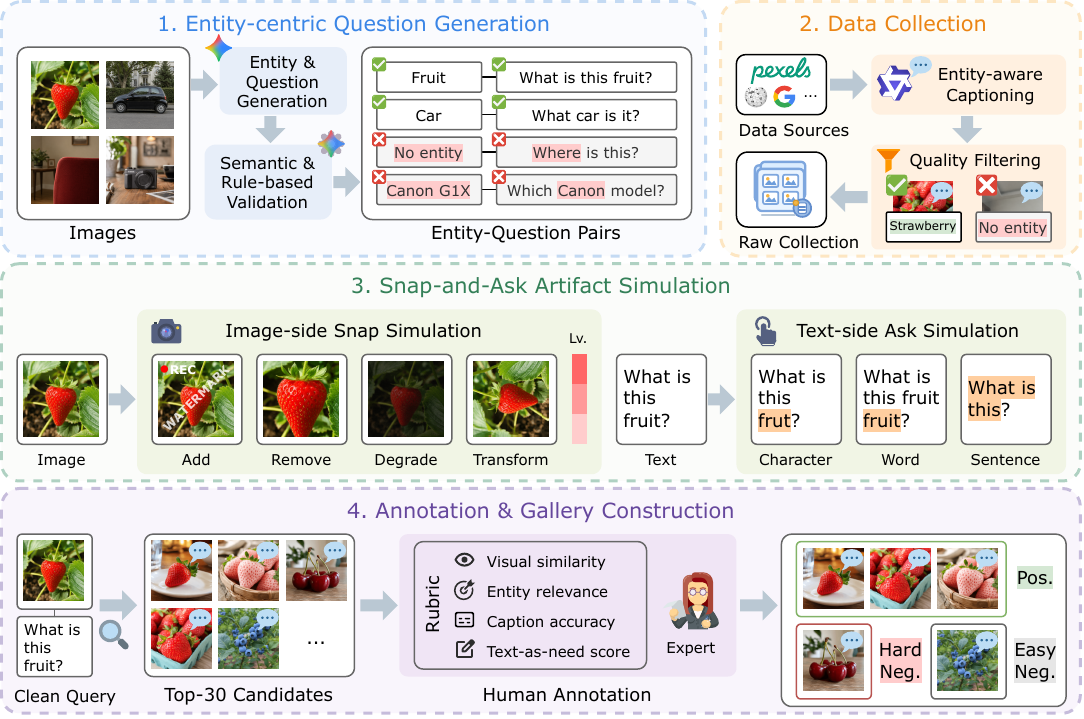}%
  % \vspace{-0.5em}
  \caption{
  Overview of the SnapBench construction pipeline. 
  This pipeline creates entity-centric snap-and-ask retrieval instances and simulates two sources of mobile interaction artifacts through controlled image and text question corruptions. 
  \texttt{Lv.}, \texttt{Pos.}, and \texttt{Neg.} denote level, positive, and negative, respectively.
  }
  \label{fig:bench_overview}
\end{figure*}

SnapBench is designed for the retrieval step in mobile \emph{snap-and-ask} interactions, where a snapped image provides concrete visual evidence and a short user question specifies the retrieval intent. We operationalize two properties identified in Section~\ref{sec:intro}: it is entity-centric and artifact-prone. To capture the entity-centric nature of the task, we use coarse questions together with a fixed gallery containing same-category hard negatives, so successful retrieval requires identifying the intended entity rather than matching a broad category. To reflect artifact-prone inputs in a controlled manner, we instantiate image-side and text-side artifacts while preserving the same retrieval intent, gallery, and labels. This paired design allows retrieval behavior to be compared across different conditions.
% A user snaps an image and types a short text; the system must retrieve gallery items that match the intended entity. In this setting, the image provides concrete visual evidence, while the text specifies the retrieval intent. The two modalities are complementary but are also produced under everyday mobile conditions, where photos may contain capture artifacts and queries may be short, noisy, or underspecified.

% This motivates three design choices. First, SnapBench is entity-centric: queries are intentionally coarse and require the snapped image to disambiguate the target entity from related gallery items. Second, the benchmark uses a fixed gallery with same-category hard negatives, so successful retrieval requires entity-level discrimination rather than broad category matching. Third, perturbations are applied to the snapped image and text question as a simulation of snap-and-ask inputs. All perturbed variants preserve the same retrieval intent, ground-truth set, and gallery, allowing retrieval behavior to be compared across clean, image-artifact, text-artifact, and joint-artifact conditions.

% ------------------------------------------------------------
\subsection{Construction Pipeline}
\label{sec:bench_pipeline}
% ------------------------------------------------------------
Figure~\ref{fig:bench_overview} summarizes the construction pipeline.

\paragraph{Stage 1: Entity-centric question construction.}
We start from a private image pool crawled from publicly accessible web sources, followed by copyright filtering and deduplication. Candidate query images are manually screened to ensure that each image contains a visually identifiable primary entity and avoids severe clutter, low visibility, or leakage-prone overlays. An independent audit on $200$ randomly sampled images yields $96\%$ three-way agreement among raters. For the retained images, Gemini-3-Flash~\cite{geminiteam2024gemini} identifies a coarse primary entity and generates a short English question, such as ``What kind of plant is this?'' or ``What type of building is this?''. We then apply four rule-based syntactic checks, followed by a read-only GPT-5.4-mini semantic check that rejects no-entity cases, malformed questions, overly fine-grained entity tags, and queries that expose brand names, person names, or place names. Failed cases are repaired for up to three rounds and rechecked. This process yields \textbf{$1{,}145$} clean image--question pairs from $1{,}150$ candidates. Full details are provided in Appendix~\ref{sec:appendix_query}.
\paragraph{Stage 2: Data collection.}
Candidate gallery items are sampled from the same private image pool under strict non-overlap with the query set. Query images are excluded before captioning. Each remaining image is paired with an entity-centric English caption generated by a private Qwen-VL-based captioning model fine-tuned for factual entity description. The caption names the primary entity and salient visual attributes without adding external knowledge. We remove entries without an identifiable entity, entries with low-information captions (e.g., single-word or templated outputs), and duplicate entries identified via caption normalization (Appendix~\ref{sec:appendix_gallery}). The raw collection contains nearly $10^8$ unique images for subsequent gallery construction. Additional collection details are given in Appendix~\ref{sec:appendix_gallery}.

\paragraph{Stage 3: Snap-and-ask artifact simulation.}
Snap-and-ask artifacts arise from two sources: the snap side, where mobile capture and interface context distort the image, and the ask side, where text questions introduce noise. SnapBench models these artifacts with deterministic corruption operators applied only to the query, leaving gallery items unchanged. Each perturbed variant therefore represents a different observation of the same snap-and-ask intent, with the candidate gallery fixed. The perturbation operators are detailed in Section~\ref{sec:bench_perturbation}.
% Starting from each clean image--question pair, we instantiate controlled image-side and text-side conditions that simulate mobile capture artifacts, interface overlays, typing noise, and underspecified questions.

\paragraph{Stage 4: Annotation and gallery construction.}
For each clean image-question query, we pre-retrieve the top-$30$ gallery candidates from the image collection using private vision-encoder-based and caption-based text retrieval systems, yielding approximately $34{,}000$ query--candidate pairs for annotation. We then convert the annotated pairs into retrieval labels according to the rubric in Section~\ref{sec:bench_annotation}: pairs with aggregated fitness ${=}0$ are excluded as easy negatives, pairs with fitness ${=}1$ are treated as hard negatives, and pairs with fitness ${\geq}2$ are used as ground-truth positives. Importantly, only positive and hard-negative candidates are merged into the final retrieval gallery, ensuring that evaluation is performed over labeled gallery items rather than unlabeled samples outside the annotation pool. These labels are shared by all perturbed variants of the same clean query, so performance changes reflect the simulated input condition rather than changes in the target set.

% ------------------------------------------------------------
\subsection{Snap-and-Ask Artifact Simulation}
\label{sec:bench_perturbation}
% ------------------------------------------------------------

\paragraph{Image-side Snap Simulation.}
Image perturbations are organized into four primitive operations that reflect common mobile snap-and-ask artifacts: \textsc{Add}, \textsc{Remove}, \textsc{Degrade}, and \textsc{Transform}. \textsc{Add} introduces visual overlays such as watermarks, scribbles, interface elements, and mosaics. \textsc{Remove} reduces visible content through cropping and downscaling. \textsc{Degrade} changes image quality through blur, compression, lighting changes, and resolution loss. \textsc{Transform} changes geometry through rotation, perspective shift, and lens distortion. We instantiate $15$ image operators at three severity levels, yielding $45$ image-side conditions. All perturbations are deterministic and seeded for reproducibility. The full parameter grid is given in Appendix~\ref{sec:appendix_corruption_tables}.

\paragraph{Text-side Ask Simulation.}
Text perturbations target the entity or category cue in the user text question. They cover three granularities with $8$ deterministic rule-based operators. At the character level, \texttt{char\_add}, \texttt{char\_delete}, \texttt{char\_change}, and \texttt{char\_swap} simulate mobile typing errors through insertion, deletion, QWERTY-adjacent replacement, and local character swaps. At the word level, \texttt{word\_repeat} duplicates the target cue and \texttt{word\_swap} exchanges it with an adjacent word. At the sentence level, \texttt{sent\_add} prepends or appends a short conversational phrase, while \texttt{sent\_replace} reduces the question to ``What is this?''. Text perturbations do not use severity levels.

% ------------------------------------------------------------
\subsection{Human Annotation Rubric}
\label{sec:bench_annotation}
% ------------------------------------------------------------

Each candidate item is scored on four dimensions: visual similarity, entity relevance, caption accuracy, and relevance to the query intent. Each dimension uses a $0$--$3$ scale. The scores are aggregated into a single intent-conditioned fitness score. Since all SnapBench queries require both the snapped image and the text question, the image+text aggregation rule is used throughout. Items with fitness ${\geq}2$ are labeled as positives. Items with fitness ${=}1$ are labeled as hard negatives. Items with fitness ${=}0$ are treated as easy negatives and excluded from the gallery. Thus, only positive and hard-negative candidates are included in the final gallery.

Finally, SnapBench contains $5{,}462$ positives, averaging $4.77$ positives per query, and $4{,}005$ hard negatives, averaging $3.50$ hard negatives per query. Ten annotators participate in the annotation. Each annotator must pass a $3{,}000$-item calibration round with at least $90\%$ accuracy. During live annotation, $20$--$30\%$ of submissions are rechecked by quality-control reviewers and project leads; incorrect audited items are re-annotated. The final audit pass rate is \textbf{$95\%$}. Full annotation guidelines and audit details are provided in Appendix~\ref{sec:appendix_annotation}.

\section{Experiments}
\label{sec:exp}
Snap-and-ask retrieval is inherently artifact-prone: snapped images can be blurred, cropped, or poorly lit, and user text questions are often coarse or underspecified. We evaluate robustness by comparing model performance on clean inputs against performance under controlled corruption conditions. Three findings characterize how corruption affects retrieval under IT$\rightarrow$IT fusion: image-side artifacts, text-side artifacts, and joint image-text artifacts.

% ------------------------------------------------------------
\subsection{Experimental Setup}
\label{sec:exp_setup}
% ------------------------------------------------------------

% ------------------------------------------------------------
% ── Tab: Main Results — Baseline / Text / Image top-level labels ──
\begin{table*}[t]
  \centering\footnotesize\setlength{\tabcolsep}{3pt}\renewcommand{\arraystretch}{1.05}

  \begin{adjustbox}{max width=\textwidth}
  \begin{tabular}{l c | c c c  c c c  c c c | c c c  c c c  c c c  c c c | c}
  \toprule
    \multirow{2}{*}[0.1ex]{\textbf{Model}} & \multirow{2}{*}[0.1ex]{\textbf{Clean}} & \multicolumn{9}{c|}{\textbf{Text Corruption}} & \multicolumn{12}{c|}{\textbf{Image Corruption}} & \multirow{2}{*}[0.1ex]{\textbf{Overall}} \\
  \cmidrule(lr){3-11}\cmidrule(lr){12-23}
                                           & & \multicolumn{3}{c}{Char} & \multicolumn{3}{c}{Word} & \multicolumn{3}{c|}{Sent} & \multicolumn{3}{c}{Degrade} & \multicolumn{3}{c}{Transform} & \multicolumn{3}{c}{Remove} & \multicolumn{3}{c|}{Add} & \\
  \midrule
  \rowcolor{gray!12} \multicolumn{24}{c}{\textit{Dual-Encoder}} \\
  SigLIP2-SO400M & 39.8 & \multicolumn{3}{>{\columncolor{DropL1}}c}{35.1} & \multicolumn{3}{>{\columncolor{DropL1}}c}{38.6} & \multicolumn{3}{>{\columncolor{DropL2}}c|}{32.4} & \cellcolor{DropL1}38.3 & \cellcolor{DropL2}32.3 & \cellcolor{DropL4}20.3 & \cellcolor{DropL1}37.3 & \cellcolor{DropL2}31.4 & \cellcolor{DropL3}28.1 & \cellcolor{DropL2}34.6 & \cellcolor{DropL3}25.7 & \cellcolor{DropL4}13.0 & \cellcolor{DropL1}35.8 & \cellcolor{DropL2}30.1 & \cellcolor{DropL4}21.3 & 30.3 \\
  CLIP-ViT-L/14 & 49.5 & \multicolumn{3}{>{\columncolor{DropL1}}c}{45.0} & \multicolumn{3}{c}{49.9} & \multicolumn{3}{>{\columncolor{DropL2}}c|}{43.1} & \cellcolor{DropL1}49.1 & \cellcolor{DropL1}45.9 & \cellcolor{DropL3}37.8 & \cellcolor{DropL1}48.6 & \cellcolor{DropL1}44.7 & \cellcolor{DropL2}41.7 & \cellcolor{DropL1}49.1 & \cellcolor{DropL2}44.5 & \cellcolor{DropL4}30.7 & \cellcolor{DropL1}48.3 & \cellcolor{DropL2}44.3 & \cellcolor{DropL4}34.2 & 43.8 \\
  BLIP-ITM-L & 57.3 & \multicolumn{3}{c}{60.2} & \multicolumn{3}{c}{58.0} & \multicolumn{3}{>{\columncolor{DropL1}}c|}{56.4} & \cellcolor{DropL1}57.2 & \cellcolor{DropL1}55.4 & \cellcolor{DropL2}48.1 & 58.5 & \cellcolor{DropL1}55.5 & \cellcolor{DropL2}52.1 & 57.7 & \cellcolor{DropL1}54.6 & \cellcolor{DropL3}43.7 & \cellcolor{DropL1}56.2 & \cellcolor{DropL2}52.1 & \cellcolor{DropL3}44.3 & 54.0 \\
  SigLIP-SO400M & 67.0 & \multicolumn{3}{c}{68.5} & \multicolumn{3}{c}{67.1} & \multicolumn{3}{c|}{68.2} & \cellcolor{DropL1}66.3 & \cellcolor{DropL1}63.9 & \cellcolor{DropL3}53.4 & 67.1 & \cellcolor{DropL1}62.3 & \cellcolor{DropL2}60.2 & \cellcolor{DropL1}66.6 & \cellcolor{DropL2}61.5 & \cellcolor{DropL4}41.7 & \cellcolor{DropL1}64.3 & \cellcolor{DropL2}61.7 & \cellcolor{DropL4}48.6 & 61.4 \\
  \midrule
  \rowcolor{gray!12} \multicolumn{24}{c}{\textit{VLM Embedding}} \\
  VLM2Vec-Full & 57.6 & \multicolumn{3}{c}{57.8} & \multicolumn{3}{c}{58.0} & \multicolumn{3}{c|}{61.3} & \cellcolor{DropL1}57.4 & \cellcolor{DropL1}54.9 & \cellcolor{DropL2}48.0 & \cellcolor{DropL1}56.8 & \cellcolor{DropL1}53.7 & \cellcolor{DropL2}50.3 & \cellcolor{DropL1}55.9 & \cellcolor{DropL2}52.6 & \cellcolor{DropL4}42.4 & \cellcolor{DropL1}56.5 & \cellcolor{DropL1}53.9 & \cellcolor{DropL3}45.9 & 53.7 \\
  E5-V & 59.1 & \multicolumn{3}{c}{59.2} & \multicolumn{3}{c}{59.1} & \multicolumn{3}{>{\columncolor{DropL1}}c|}{58.7} & \cellcolor{DropL1}58.6 & \cellcolor{DropL1}57.2 & \cellcolor{DropL2}52.2 & \cellcolor{DropL1}57.4 & \cellcolor{DropL1}54.4 & \cellcolor{DropL2}52.7 & \cellcolor{DropL1}57.3 & \cellcolor{DropL1}55.1 & \cellcolor{DropL3}44.6 & \cellcolor{DropL1}58.1 & \cellcolor{DropL1}55.9 & \cellcolor{DropL2}50.1 & 55.4 \\
  UME-R1-7B & 60.2 & \multicolumn{3}{>{\columncolor{DropL1}}c}{59.7} & \multicolumn{3}{c}{60.2} & \multicolumn{3}{c|}{62.5} & \cellcolor{DropL1}58.1 & \cellcolor{DropL1}58.0 & \cellcolor{DropL3}49.4 & \cellcolor{DropL1}58.5 & \cellcolor{DropL1}55.9 & \cellcolor{DropL1}55.3 & \cellcolor{DropL2}54.7 & \cellcolor{DropL2}53.1 & \cellcolor{DropL4}40.1 & \cellcolor{DropL1}57.3 & \cellcolor{DropL1}57.4 & \cellcolor{DropL2}50.6 & 55.4 \\
  VLM2Vec-V2 & 62.5 & \multicolumn{3}{c}{63.8} & \multicolumn{3}{c}{64.8} & \multicolumn{3}{c|}{66.8} & 62.7 & \cellcolor{DropL1}61.7 & \cellcolor{DropL2}52.6 & 62.8 & \cellcolor{DropL1}59.7 & \cellcolor{DropL1}58.6 & 63.1 & \cellcolor{DropL1}58.3 & \cellcolor{DropL4}45.4 & 63.0 & \cellcolor{DropL1}61.1 & \cellcolor{DropL2}54.5 & 59.9 \\
  GME-2B & 61.8 & \multicolumn{3}{>{\columncolor{DropL1}}c}{60.6} & \multicolumn{3}{>{\columncolor{DropL1}}c}{61.5} & \multicolumn{3}{>{\columncolor{DropL1}}c|}{58.9} & \cellcolor{DropL1}61.2 & \cellcolor{DropL1}59.7 & \cellcolor{DropL2}53.4 & \cellcolor{DropL1}61.5 & \cellcolor{DropL1}59.7 & \cellcolor{DropL1}58.8 & 65.3 & 62.2 & \cellcolor{DropL2}53.2 & 64.2 & 62.6 & \cellcolor{DropL1}57.5 & 60.0 \\
  Qwen3-VL-Emb-2B & 64.6 & \multicolumn{3}{c}{66.1} & \multicolumn{3}{c}{66.4} & \multicolumn{3}{c|}{71.0} & \cellcolor{DropL1}64.3 & \cellcolor{DropL1}62.8 & \cellcolor{DropL3}53.4 & 64.7 & \cellcolor{DropL1}61.5 & \cellcolor{DropL1}61.1 & \cellcolor{DropL1}63.3 & \cellcolor{DropL2}57.9 & \cellcolor{DropL4}42.3 & \cellcolor{DropL1}64.2 & \cellcolor{DropL1}62.9 & \cellcolor{DropL2}55.4 & 61.2 \\
  Jina-V4 & 65.8 & \multicolumn{3}{c}{69.5} & \multicolumn{3}{c}{65.9} & \multicolumn{3}{c|}{71.0} & 66.1 & \cellcolor{DropL1}64.3 & \cellcolor{DropL3}55.3 & \cellcolor{DropL1}65.4 & \cellcolor{DropL1}62.4 & \cellcolor{DropL1}61.9 & \cellcolor{DropL1}64.3 & \cellcolor{DropL2}57.9 & \cellcolor{DropL4}44.9 & 65.9 & \cellcolor{DropL1}63.6 & \cellcolor{DropL3}54.4 & 62.2 \\
  Qwen3-VL-Emb-8B & 65.5 & \multicolumn{3}{c}{66.9} & \multicolumn{3}{c}{67.2} & \multicolumn{3}{c|}{69.6} & 65.7 & \cellcolor{DropL1}63.9 & \cellcolor{DropL2}55.7 & \cellcolor{DropL1}64.3 & \cellcolor{DropL1}62.7 & \cellcolor{DropL1}62.3 & \cellcolor{DropL1}63.1 & \cellcolor{DropL2}58.7 & \cellcolor{DropL4}46.8 & \cellcolor{DropL1}65.4 & \cellcolor{DropL1}64.3 & \cellcolor{DropL2}58.4 & 62.3 \\
  GME-7B & 68.0 & \multicolumn{3}{c}{68.9} & \multicolumn{3}{c}{68.2} & \multicolumn{3}{c|}{69.7} & \cellcolor{DropL1}67.5 & \cellcolor{DropL1}65.5 & \cellcolor{DropL2}58.5 & \cellcolor{DropL1}66.3 & \cellcolor{DropL1}63.3 & \cellcolor{DropL1}63.6 & \cellcolor{DropL1}67.0 & \cellcolor{DropL1}63.7 & \cellcolor{DropL3}54.1 & \cellcolor{DropL1}67.6 & \cellcolor{DropL1}66.0 & \cellcolor{DropL2}59.8 & 64.6 \\
  RzenEmbed-7B & 75.2 & \multicolumn{3}{c}{76.8} & \multicolumn{3}{c}{76.5} & \multicolumn{3}{c|}{77.1} & \cellcolor{DropL1}73.5 & \cellcolor{DropL1}72.1 & \cellcolor{DropL4}56.2 & \cellcolor{DropL1}73.0 & \cellcolor{DropL2}66.4 & \cellcolor{DropL1}72.8 & \cellcolor{DropL1}75.9 & \cellcolor{DropL1}74.5 & \cellcolor{DropL2}68.6 & \cellcolor{DropL1}75.6 & \cellcolor{DropL1}74.4 & \cellcolor{DropL2}66.8 & 72.0 \\
  Ops-MM-2B & 77.6 & \multicolumn{3}{c}{78.0} & \multicolumn{3}{c}{77.6} & \multicolumn{3}{c|}{78.6} & \cellcolor{DropL1}76.8 & \cellcolor{DropL1}76.9 & \cellcolor{DropL4}57.7 & \cellcolor{DropL1}75.8 & \cellcolor{DropL2}68.2 & \cellcolor{DropL1}73.3 & \cellcolor{DropL1}76.7 & \cellcolor{DropL1}76.1 & \cellcolor{DropL2}68.1 & \cellcolor{DropL1}77.4 & \cellcolor{DropL1}76.3 & \cellcolor{DropL3}66.7 & 73.6 \\
  Ops-MM-7B & 79.1 & \multicolumn{3}{c}{79.2} & \multicolumn{3}{c}{79.4} & \multicolumn{3}{>{\columncolor{DropL1}}c|}{77.8} & \cellcolor{DropL1}77.9 & \cellcolor{DropL1}76.1 & \cellcolor{DropL4}58.0 & \cellcolor{DropL1}77.6 & \cellcolor{DropL2}70.1 & \cellcolor{DropL1}74.8 & \cellcolor{DropL1}79.4 & \cellcolor{DropL1}77.6 & \cellcolor{DropL2}71.0 & \cellcolor{DropL1}78.8 & \cellcolor{DropL1}77.6 & \cellcolor{DropL3}67.4 & 74.8 \\
  \bottomrule
  \end{tabular}
  \end{adjustbox}
\caption{%
    Main results on \benchname{} (Recall@1, \%). Text columns report joint retrieval under character-, word-, and sentence-level query perturbations. Image columns report joint retrieval under four perturbation primitives at severity levels $s_1$, $s_2$, and $s_3$. Darker cells denote larger performance degradation from the clean score.
    }
  \label{tab:main_results}
\end{table*}

% ------------------------------------------------------------

\paragraph{Models.}
We evaluate 16 multimodal retrieval models. The first group contains dual-encoder or dual-score baselines with late fusion: CLIP-ViT-L/14~\cite{radford2021clip}, SigLIP-SO400M~\cite{zhai2023siglip}, SigLIP2-SO400M~\cite{tschannen2025siglip2}, and BLIP-ITM-L~\cite{li2022blip}. The second group contains VLM-based embedding models: Jina-V4~\cite{jina2025v4}, Qwen3-VL-Embedding-2B/8B~\cite{Li2026Qwen3VLEmbeddingAQ}, E5-V~\cite{jiang2024e5v}, VLM2Vec-V2/Full~\cite{jiang2024vlm2vec}, UME-R1-7B~\cite{ume2024}, GME-2B/7B~\cite{zhang2024gme}, Ops-MM-2B/7B~\cite{opsmm2025}, and RzenEmbed-7B~\cite{Jian2025RzenEmbedTC}.

\paragraph{Retrieval modes.}
Each query consists of a snapped image and a user text question, and each gallery item consists of an image and an entity-aware caption. We evaluate 5 retrieval modes, denoted \textit{query\,$\rightarrow$\,gallery}: joint-to-joint ({IT$\rightarrow$IT}), image-to-image ({I$\rightarrow$I}), image-to-joint ({I$\rightarrow$IT}), text-to-text ({T$\rightarrow$T}), and text-to-joint ({T$\rightarrow$IT}). Following~\citet{wei2024uniir} and \citet{radford2021clip}, for dual-encoder models, any joint-side score is computed via late fusion: $s_{\mathrm{joint}}=\frac{1}{2}\left(s_{\mathrm{img}}+s_{\mathrm{text}}\right)$, where $s_{\mathrm{img}}$ and $s_{\mathrm{text}}$ denote image--image and text--text similarities. For VLM-based embedding models, we use their released multimodal encoding interface, which combines image and text within a single frozen forward pass rather than an externally-imposed weight. We use the term \textbf{fixed fusion} to cover both cases: an explicit, fixed-weight combination for dual-encoder models, and an implicit combination baked into a frozen forward pass for VLM-based embedding models. In neither case can the image--text weighting adapt to per-query modality reliability, which is the property MOOR (\Cref{sec:method_overview}) targets. The main paper focuses on the {IT$\rightarrow$IT} mode; full five-mode results are reported in Appendix~\ref{app:additional_results}.

\paragraph{Metrics.}
We use Recall at rank $k$ (R@$k$) as the evaluation metric. The main paper reports R@1, as the first returned item matters most in mobile snap-and-ask interactions. All results are macro-averaged over the full query set. Since clean and corrupted variants share the same target set and gallery, performance changes can be interpreted as paired robustness effects. We use a fixed gallery order and seeded corruptions. Gallery embeddings are pre-computed once per model and reused across all corruption conditions, following standard retrieval evaluation practice~\cite{jiang2024vlm2vec, wei2024uniir, zhang2024gme}. Further implementation details are given in Appendix~\ref{sec:appendix_setup}.

\paragraph{Conditions.}
We evaluate clean inputs, text corruptions, image corruptions, and joint corruptions. Text corruptions contain $8$ deterministic operators grouped into character-, word-, and sentence-level conditions. Image corruptions contain $15$ operators grouped into degrade, transform, remove, and add, each evaluated at three severity levels $s_1$, $s_2$, and $s_3$---yielding $45$ image-side conditions. For joint artifacts, we use six representative image--text operator pairs listed in \Cref{tab:super_additivity}.

% ------------------------------------------------------------
\subsection{Main Results}
\label{sec:exp_main}
% ------------------------------------------------------------

\Cref{tab:main_results} reports clean R@1 across all 16 models. Several observations can be drawn from the results.
(1) Clean R@1 ranges from $39.8$ to $79.1$, with no model exceeding $80$---showing that SnapBench retains discriminative power despite a moderate gallery size, and that substantial room for improvement remains. VLM-based embedding models outperform dual-encoder baselines by a $13$-point margin on average, with Ops-MM-7B, Ops-MM-2B, and RzenEmbed-7B at the top.
(2) Clean performance does not predict robustness. Image corruption drops mean R@1 from $63.2$ to $57.9$, averaged across all corruption types and severities. Both paradigms degrade, though dual encoders drop more than VLM-based models. Rank order also shifts: SigLIP-SO400M places high on clean but loses heavily at severe corruption, while GME-2B scores lower on clean yet drops less on average. Clean accuracy and corruption robustness must be evaluated separately.
(3) Text corruption shows a different pattern. Character-, word-, and sentence-level corruptions yield mean R@1 of $63.5$, $63.7$, and $63.9$---less than $1$ point from the clean mean---while image corruption causes a $5.3$-point drop under the same averaging. \Cref{sec:exp_text_artifacts} examines why this stability does not indicate robustness.
(4) Joint corruption---simultaneously corrupting both the image and text of the query---reveals the most severe failure. The resulting R@1 drop exceeds the sum of the two single-component drops by $+7.6$ points on average, a super-additive interaction that cannot be inferred from single-modality evaluations alone. 
% \Cref{sec:exp_joint_artifacts} gives further analysis.

% ------------------------------------------------------------
\subsection{Snap-and-Ask Artifact Analysis}
\label{sec:exp_robustness}
% ------------------------------------------------------------

\begin{table*}[t]
\centering

\resizebox{\linewidth}{!}{%
\setlength{\tabcolsep}{2pt}\renewcommand{\arraystretch}{1.05}\scriptsize
\begin{tabular}{l c | ccc | ccc | ccc | ccc | ccc | ccc | ccc | c}
\toprule
\multirow{2}{*}[0.1ex]{\textbf{Model}} & \multirow{2}{*}[0.1ex]{\textbf{Base.}} & \multicolumn{3}{c|}{\textbf{C+C}} & \multicolumn{3}{c|}{\textbf{D+D}} & \multicolumn{3}{c|}{\textbf{L+W}} & \multicolumn{3}{c|}{\textbf{M+S}} & \multicolumn{3}{c|}{\textbf{E+W}} & \multicolumn{3}{c|}{\textbf{R+S}} & \multicolumn{3}{c|}{\textbf{Mean}} & \multirow{2}{*}[0.1ex]{\textbf{Overall}} \\
\cmidrule(lr){3-20}\cmidrule(lr){21-23}
 & & Exp. & Obs. & $\delta$ & Exp. & Obs. & $\delta$ & Exp. & Obs. & $\delta$ & Exp. & Obs. & $\delta$ & Exp. & Obs. & $\delta$ & Exp. & Obs. & $\delta$ & Exp. & Obs. & $\delta$ & \\
\midrule
\rowcolor{gray!10} \multicolumn{24}{c}{\textit{Dual-Encoder}} \\
SigLIP2\mbox{-}SO400M & 39.8 & 5.8 & 18.9 & \textbf{13.0} & 27.6 & 35.5 & \textbf{7.9} & 11.3 & 23.8 & \textbf{12.5} & 10.3 & 23.3 & \textbf{13.0} & 8.3 & 16.9 & \textbf{8.6} & 23.5 & 32.6 & \textbf{9.1} & 14.5 & 25.2 & \textbf{10.7} & 25.2 \\
CLIP\mbox{-}ViT\mbox{-}L/14 & 49.5 & -0.8 & 21.8 & \textbf{22.6} & 4.8 & 29.5 & \textbf{24.7} & 6.7 & 23.5 & \textbf{16.7} & 8.5 & 24.0 & \textbf{15.5} & 5.7 & 22.4 & \textbf{16.7} & 6.8 & 38.7 & \textbf{31.9} & 5.3 & 26.6 & \textbf{21.4} & 26.6 \\
BLIP\mbox{-}ITM\mbox{-}L & 57.3 & 0.2 & 6.1 & \textbf{5.9} & 5.6 & 13.0 & \textbf{7.3} & 7.2 & 12.9 & \textbf{5.7} & 6.0 & 11.8 & \textbf{5.8} & 6.8 & 13.4 & \textbf{6.6} & 9.2 & 24.8 & \textbf{15.6} & 5.8 & 13.7 & \textbf{7.8} & 13.7 \\
SigLIP\mbox{-}SO400M & 67.0 & -3.3 & 2.2 & \textbf{5.5} & 9.5 & 13.4 & \textbf{3.9} & 8.0 & 11.5 & \textbf{3.4} & 5.7 & 11.7 & \textbf{6.0} & 7.5 & 9.0 & \textbf{1.6} & -0.1 & 7.0 & \textbf{7.1} & 4.5 & 9.1 & \textbf{4.6} & 9.1 \\
\midrule
\rowcolor{gray!10} \multicolumn{24}{c}{\textit{VLM Embedding}} \\
GME\mbox{-}7B & 68.0 & 0.9 & 48.5 & \textbf{47.7} & 4.7 & 50.5 & \textbf{45.7} & 6.9 & 45.1 & \textbf{38.2} & 7.9 & 50.7 & \textbf{42.9} & 5.3 & 45.1 & \textbf{39.9} & 1.5 & 6.1 & \textbf{4.6} & 4.5 & 41.0 & \textbf{36.5} & 41.0 \\
GME\mbox{-}2B & 61.8 & 1.2 & 48.2 & \textbf{47.1} & -3.8 & 48.2 & \textbf{52.1} & 5.2 & 45.4 & \textbf{40.2} & 3.5 & 47.4 & \textbf{43.9} & 2.8 & 42.0 & \textbf{39.2} & 0.1 & -4.6 & -4.7 & 1.5 & 37.8 & \textbf{36.3} & 37.8 \\
UME\mbox{-}R1\mbox{-}7B & 60.2 & 0.9 & 12.8 & \textbf{11.9} & 13.4 & 25.6 & \textbf{12.2} & 9.9 & 18.9 & \textbf{9.1} & 9.0 & 18.0 & \textbf{9.0} & 6.5 & 14.6 & \textbf{8.1} & 2.7 & 14.7 & \textbf{12.0} & 7.1 & 17.5 & \textbf{10.4} & 17.5 \\
Qwen3\mbox{-}VL\mbox{-}Emb\mbox{-}2B & 64.6 & 0.1 & 13.4 & \textbf{13.3} & 13.0 & 27.7 & \textbf{14.6} & 7.2 & 18.1 & \textbf{10.9} & 6.9 & 12.0 & \textbf{5.1} & 4.8 & 15.2 & \textbf{10.3} & 1.3 & 15.7 & \textbf{14.4} & 5.6 & 17.0 & \textbf{11.4} & 17.0 \\
Qwen3\mbox{-}VL\mbox{-}Emb\mbox{-}8B & 65.5 & -0.3 & 12.8 & \textbf{13.1} & 11.0 & 25.5 & \textbf{14.5} & 5.7 & 16.7 & \textbf{11.1} & 5.8 & 15.9 & \textbf{10.0} & 3.4 & 14.9 & \textbf{11.5} & -1.2 & 13.5 & \textbf{14.6} & 4.1 & 16.5 & \textbf{12.5} & 16.5 \\
Ops\mbox{-}MM\mbox{-}7B & 79.1 & 1.5 & 1.6 & \textbf{0.2} & 8.9 & 7.0 & -2.0 & 12.2 & 8.3 & -3.9 & 14.0 & 11.2 & -2.8 & 6.7 & 5.8 & -0.9 & 7.7 & 8.3 & \textbf{0.6} & 8.5 & 7.0 & -1.5 & 7.0 \\
Jina\mbox{-}V4 & 65.8 & -0.4 & -0.1 & \textbf{0.4} & 13.2 & 16.1 & \textbf{2.9} & 7.9 & 7.9 & 0.0 & 7.5 & 8.1 & \textbf{0.6} & 6.6 & 6.6 & -0.0 & 1.8 & 2.7 & \textbf{0.9} & 6.1 & 6.9 & \textbf{0.8} & 6.9 \\
E5\mbox{-}V & 59.1 & 0.1 & 2.2 & \textbf{2.0} & 4.5 & 6.9 & \textbf{2.4} & 3.8 & 3.6 & -0.2 & 1.9 & 11.1 & \textbf{9.1} & 4.4 & 5.3 & \textbf{0.9} & 2.9 & 7.8 & \textbf{4.9} & 2.9 & 6.1 & \textbf{3.2} & 6.1 \\
Ops\mbox{-}MM\mbox{-}2B & 77.6 & 0.6 & 1.4 & \textbf{0.8} & 8.5 & 6.9 & -1.6 & 12.2 & 7.3 & -4.9 & 13.2 & 8.5 & -4.7 & 6.9 & 4.3 & -2.5 & 6.1 & 5.1 & -1.1 & 7.9 & 5.6 & -2.3 & 5.6 \\
RzenEmbed\mbox{-}7B & 75.2 & 0.4 & -0.5 & -0.9 & 7.5 & 6.0 & -1.5 & 13.4 & 7.1 & -6.3 & 11.3 & 8.6 & -2.7 & 6.3 & 4.0 & -2.3 & 5.5 & 6.2 & \textbf{0.7} & 7.4 & 5.2 & -2.2 & 5.2 \\
VLM2Vec\mbox{-}V2 & 62.5 & -0.4 & -12.4 & -12.0 & 7.5 & -0.3 & -7.8 & 4.1 & -5.4 & -9.4 & 6.0 & -2.7 & -8.7 & 0.8 & -11.8 & -12.5 & 5.4 & -14.5 & -20.0 & 3.9 & -7.8 & -11.7 & -7.8 \\
VLM2Vec\mbox{-}Full & 57.6 & 0.8 & -12.4 & -13.2 & 1.4 & -7.6 & -9.0 & 2.8 & -10.3 & -13.1 & 6.8 & -8.6 & -15.3 & 4.1 & -6.9 & -11.1 & 21.5 & -11.8 & -33.2 & 6.2 & -9.6 & -15.8 & -9.6 \\
\midrule
\textit{Mean} & & 0.5 & 10.3 & \textbf{9.8} & 8.6 & 19.0 & \textbf{10.4} & 7.8 & 14.6 & \textbf{6.9} & 7.8 & 15.7 & \textbf{7.9} & 5.4 & 12.6 & \textbf{7.1} & 5.9 & 9.5 & \textbf{3.6} & 6.0 & 13.6 & \textbf{7.6} & 13.6 \\
\bottomrule
\end{tabular}}
\caption{%
    Interaction between image-side and text-side artifacts.
  {Exp.}~$=\Delta_I+\Delta_T$ (naive additive prediction);
  {Obs.}~= observed joint drop;
  $\boldsymbol{\delta}$~$=$~Obs.$-$Exp.\ (${>}0{=}$ super-additive, \textbf{bold}).
  {Overall} = mean Obs.\ across all six joint conditions.
  C+C = Compression$+$CharInsert;
  D+D = Downscale$+$CharDelete; L+W = LowLight$+$WordSwap;
  M+S = Mosaic$+$SentAdd; E+W = Overexposure$+$WordRepeat;
  R+S = Rotation$+$SentReplace.}
\label{tab:super_additivity}
\end{table*}

\paragraph{Image-side artifacts damage entity evidence.}
\label{sec:exp_image_artifacts}

\begin{figure}[t]
  \centering
  \includegraphics[width=\columnwidth]{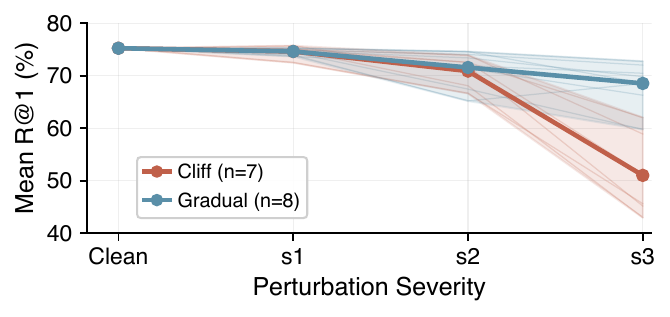}
  \caption{%
    {Severity response of image-side corruptions.}
    Thick lines show group means;
    faint lines are individual operators.
    \emph{Cliff} ($n{=}7$): Mosaic, Low Light, Downscale, Crop, Watermark,
    Low Res, Defocus.
    \emph{Gradual} ($n{=}8$): Perspective, UI Elements, Scribble, Motion Blur,
    JPEG, Rotation, Lens Distortion, Overexposure.}
  \label{fig:dose_response}
\end{figure}

Image-side corruptions produce a clear dose--response effect. Averaged over the four image primitives, the R@1 drop is only $0.8$ points at $s_1$ severity, increases to $3.8$ points at $s_2$ severity, and reaches $11.2$ points at $s_3$ severity. Thus, models tolerate mild artifacts but fail once corruption removes or obscures discriminative evidence. The four primitives affect retrieval unequally: at the highest severity, remove is most damaging ($16.3$ R@1 points), followed by degrade ($12.5$) and add ($10.9$), while transform is milder ($5.2$). This ordering matches SnapBench's entity-centric design: cropping, downscaling, overlays, and severe low-light directly obscure the target entity, whereas geometric transformations largely preserve entity identity. Figure~\ref{fig:dose_response} shows robustness is not always smooth: some operators degrade gradually, while others exhibit cliff behavior, changing little until collapsing at $s_3$ severity (low light, mosaic, watermark, downscaling, and cropping are the strongest cliff-prone operators)---average scores can therefore hide high-risk regimes common in mobile use. The failure-mode analysis in Appendix~\ref{sec:appendix_failure_analysis} confirms this pattern: under mild corruption, models retrieve a visually similar but wrong entity, while under high-severity remove and degrade, the target entity becomes unrecognizable and retrieval collapses entirely.
% Under mild corruption, models retrieve a visually similar but wrong entity. Under high-severity remove and degrade, the target entity becomes unrecognizable and retrieval collapses entirely.
% 

% ------------------------------------------------------------
% ------------------------------------------------------------

\paragraph{Text-side artifacts reveal text underuse and coarse-text drag.}
\label{sec:exp_text_artifacts}

As reported in \Cref{sec:exp_main}, textual question corruption barely changes joint R@1. This stability is not a sign of robustness---it reflects that the text channel was already contributing little in this task. In snap-and-ask interactions, users formulate queries around what they see in the image: questions such as ``What species is this flower?'' or ``What brand is this sneaker?'' express broad retrieval intent rather than identifying the specific target entity. In a gallery with same-category hard negatives, fusing such coarse text raises scores for many related candidates, diluting the strong visual ranking toward category-level confusions. We term this ranking-dilution effect \textbf{coarse-text drag}. The weak response to text corruption follows as a consequence: when the text channel is already miscalibrated under fixed fusion, corrupting it further changes joint R@1 only slightly.

The retrieval-mode ablation (\Cref{tab:ablation_modes_dual,tab:ablation_modes_vlm}) confirms this directly. Joint retrieval ({IT$\rightarrow$IT}) underperforms image-only ({I$\rightarrow$I}) for the majority of models across clean and all corruption conditions, with a mean drag of $-11.21$ R@1 across all 16 models (full per-model table in Appendix~\ref{sec:appendix_more_results}). The text channel is not inherently harmful---it is assigned more weight than its signal quality justifies under fixed fusion.

A causal intervention isolates coarse granularity, rather than the text channel itself, as the cause of the drag: on GME-7B, replacing the coarse question (``What flower is this?'') with entity-label text (``A tulip.'') reduces the drag from $-7.68$ to $-0.17$ R@1 points, a $97.8\%$ elimination. Conversely, a text-essential subset ($n{=}69/1{,}145$, queries with discriminative modifiers and genuine visual ambiguity; Appendix~\ref{sec:appendix_more_results}) shows drag shrinking on $14$ of $16$ models when text does carry discriminative signal, confirming that the effect is calibration-scoped rather than a general failure to exploit user text.

\paragraph{Joint artifacts are non-additive.}
\label{sec:exp_joint_artifacts}

We next ask whether corruption on the image and text components of a query can be composed: if image corruption alone drops R@1 by $d_I$ and text corruption alone by $d_T$, does joint corruption drop by roughly $d_I+d_T$? Each query pairs a snapped image with a user text question; we hold the gallery fixed and corrupt only the query side. Let $r_0$ be clean R@1 under {IT$\rightarrow$IT} retrieval, and let $r_I$, $r_T$, and $r_{IT}$ denote R@1 when only the image, only the text, or both are corrupted. We define drops $d_I=r_0-r_I$, $d_T=r_0-r_T$, $d_{IT}=r_0-r_{IT}$,
% \begin{equation}
% ,\quad
% ,\quad
% ,
% \end{equation}
and interaction with $\gamma_{\mathrm{int}}=d_{IT}-\left(d_I+d_T\right)$.
% \begin{equation}
% .
% \label{eq:interaction_gap}
% \end{equation}
A positive $\gamma_{\mathrm{int}}$ means joint corruption hurts more than the sum of the two single-component drops (super-additive); a negative value means sub-additive interaction.

\Cref{tab:super_additivity} shows that composition fails. Averaged over six paired image--text corruption conditions and all 16 models, the mean interaction gap is $+7.6$ R@1 points: observed joint drops ($13.6$ on average) exceed the na\"ive sum of single-component drops ($6.0$). The split is model-dependent (full per-model values in the table): GME-2B/7B show the largest super-additive gaps (mean $\delta\!>\!+30$), while Ops-MM-2B/7B and VLM2Vec-Full/V2 are sub-additive on average. Robustness under snap-and-ask corruption therefore requires joint evaluation of both query components; separate image-corruption and text-corruption tests cannot predict the paired regime. This paired setting is also the hardest test for fixed fusion: both signals degrade at once, yet the model must still set their relative weight.

\section{Adaptive Modality Fusion with MOOR}
\label{sec:method}

\subsection{Modality-Anchored Reweighting via Outlier-Aware Gating}
\label{sec:method_overview}

Section~\ref{sec:exp} traces a single chain of failure under fixed \textbf{IT$\rightarrow$IT} fusion. Finding~1 shows that image corruption can collapse retrieval, but the same global weights prevent leaning on text-involving paths when the photo fails. Finding~2 shows the converse on clean inputs: coarse text is over-weighted, so joint mode trails image-only ({coarse-text drag}) even before corruption. Finding~3 compounds both---joint drops are super-additive because fusion still cannot rebalance whichever channel remains more reliable. The three findings are not three separate bugs; they are one \emph{signal-blind} fusion rule failing whenever per-query modality reliability shifts. Snap-and-ask retrieval therefore needs adaptive calibration, not a fixed global weight.

We introduce MOOR (\textbf{M}odality-anchored, \textbf{O}utlier-aware, \textbf{O}ptimal \textbf{R}eweighting): a training-free adapter that reweights four gallery score paths per query---$\mathbf{s}^{II}$, $\mathbf{s}^{TT}$, $\mathbf{s}^{IT}$, $\mathbf{s}^{TI}$---using only the frozen encoder's outputs. Embeddings are gallery-whitened; $\mathbf{s}^{II}$ anchors the rank. Each text-involving path is gated by Pearson correlation with $\mathbf{s}^{II}$ through a bell function (suppress unreliable and redundant paths), then scaled by score variance. The fused score is the normalized weighted sum. No learned parameters; full details in Appendix~\ref{app:raft_detail}.

\begin{figure}[t]
    \centering
    \includegraphics[width=1\linewidth]{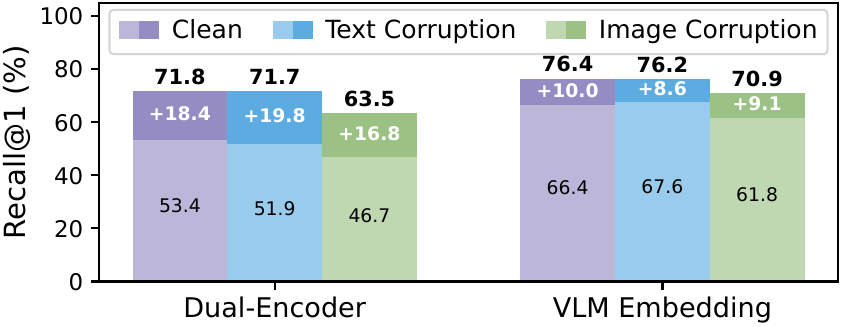}
    \caption{
    Average performance of MOOR in joint retrieval. Darker regions denote gains over fixed fusion.
    }
    \label{fig:moor_clean_text_image_gain}
\end{figure}

\subsection{Evaluation on \benchname{}}
\label{sec:method_eval}

We evaluate MOOR on the same setup as Section~\ref{sec:exp}; only fusion changes. As shown in Figure~\ref{fig:moor_clean_text_image_gain}, MOOR improves over fixed fusion across clean, text-corrupted, and image-corrupted settings for both dual-encoder and VLM-embedding models. These gains suggest that adaptive fusion better calibrates the contribution of image and text signals when their reliability varies across snap-and-ask inputs. The results also support the coarse-text drag: user text is not inherently harmful, but fixed fusion over-weights the text channel relative to its signal quality under coarse snap-and-ask phrasing, weakening an otherwise strong visual ranking.

Additional results in Appendix~\ref{sec:appendix_raft_full} show that gains generally increase with image corruption severity, but diminish when artifacts remove the target entity itself, as discussed in Appendix~\ref{sec:appendix_failure_analysis}. 

\paragraph{Why MOOR generalizes across architectures.} MOOR operates solely on statistical properties of the frozen encoder's own retrieval scores---rank consistency against the image-only anchor path and score variance---without accessing model internals, so its mechanism does not depend on any particular backbone. The oracle analysis (Appendix~\ref{sec:appendix_raft_motivation}) explains why this helps universally: for all 16 evaluated models, the per-query optimal fusion weight is highly dispersed (std $0.086$--$0.138$ on a $[0,1]$ scale), meaning fixed-fusion miscalibration is a model-independent property of current snap-and-ask retrievers rather than an artifact of any single architecture.
% Under joint corruption, \methodname{} remains helpful, although the paired conditions in Table{tab:super_additivity} remain the hardest setting.

\section{Conclusion}
\label{sec:conclusion}

We introduced SnapBench, a paired benchmark for mobile snap-and-ask multimodal retrieval. SnapBench focuses on entity-centric interactions where users provide a snapped image and a text question, and it preserves the same target, gallery, and annotations across clean and perturbed inputs. Our evaluation of 16 retrieval models shows that current systems remain fragile in this setting. Image artifacts cause the largest failures, especially when they obscure or remove entity-level evidence. Text artifacts have weaker effects in joint retrieval. We further observe coarse-text drag, where adding a text question can reduce accuracy compared with image-only retrieval, and show that joint image--text artifacts interact non-additively. Finally, MOOR provides a lightweight diagnostic baseline whose gains suggest that modality calibration is important for future snap-and-ask retrieval systems.

% ARR: Limitations (required) and optional Ethical Considerations after
% Conclusion, before References, without page breaks. ARR recommends the
% ethics section be titled "Ethical Considerations". Acknowledgments go
% immediately before References. Both Limitations and Ethics are outside
% the content page limit (EMNLP CR: up to one extra page for them).
\section*{Limitations}
All findings reported in this paper are scoped to the snap-and-ask entity retrieval setting rather than to mobile AI more broadly. \benchname{} is a deliberately controlled benchmark and does not aim to substitute for in-the-wild deployment evaluation. We highlight three limitations. 
(1) Synthetic corruptions. The 53 corruption conditions are programmatically generated rather than crawled from real noisy uploads; this is a deliberate trade-off favouring factorial controllability and reproducibility over distributional fidelity. We argue in \Cref{sec:bench_perturbation} that the primitive-level taxonomy spans the operator space, but the joint distribution of corruptions in the wild may differ from our marginal distributions~\citep{VSI_Super_Wild}. 
(2) Scale. \benchname{} contains 1{,}145 queries and 9{,}085 gallery items; we prioritise dense ground-truth annotations and tight hard-negative coverage over raw scale, which yields stronger discriminative signal but smaller absolute numbers than web-scale benchmarks. 
(3) MOOR in in-distribution settings. MOOR's per-query adaptation is most effective when modal signal quality varies across queries; on queries where both modalities are uniformly informative or uniformly uninformative, MOOR reverts to fixed-fusion behavior and provides no advantage by construction.

\section*{Ethical Considerations}
SnapBench is constructed from a private image pool collected from publicly accessible web sources, followed by copyright filtering, deduplication, and content safety screening. Gallery items are sourced from the same library.
The private pool is used only as a source for constructing the benchmark; we do not release the full pool. 
The released benchmark contains the query and gallery items used for evaluation, together with their annotations and corruptions. Annotators were recruited through a third-party data annotation vendor, compensated above the local minimum wage, and informed of the research nature of the task; consent for the research use of their annotations was included as a standard term in the data-labeling agreement between our organization and the vendor. The annotation protocol was reviewed by our organization's internal ethics and data-compliance committee prior to data collection. Annotation guidelines deliberately exclude content involving violence, sexual material, or political sensitivity. 
We do not foresee direct negative societal impacts of the benchmark itself; we note however that MOOR, as a training-free fusion adapter, inherits the biases of whichever underlying encoder it is applied to.
The pilot analysis (\Cref{sec:appendix_prevalence}) was conducted on de-identified data under the production service's User Agreement and Privacy Policy for internal algorithm-optimization purposes; results are reported only as aggregate statistics, with no personally identifying information released. The underlying distribution statistics are derived entirely from automated signal-level metrics; a small subset of images (4\%) was reviewed solely to calibrate automated detection thresholds, with no content extracted or retained beyond a binary detection outcome per image. This analysis completed under our internal data-governance review prior to submission.
Web-sourced gallery and query images are released under a research-only license that prohibits commercial redistribution, together with a takedown mechanism (contact channel and removal workflow) on the release page.

\section*{Acknowledgments}
This work was supported by the CCF-Tencent Rhino-Bird Open Research Fund (No.~CCF-Tencent RAGR20250119), the Guangdong Basic and Applied Basic Research Foundation (No.~2025A1515010304), the Guangdong Province Project (No.~2024QN11X088), and the Guangzhou Science and Technology Planning Project (No.~2025A03J4491).

\bibliography{asset/reference}

\newpage
% NOTE: the appendix is two-column (ACL/ARR requires double-column appendices
% since July 2025); \input{chapter/appendix/00_cover} issues \twocolumn[...]
% for its heading block, and that mode persists through the rest of the
% appendix. Wide tables/figures below use table*/figure* to span both columns.

\appendix
\label{sec:appendix}

% % Appendix cover – auto-generated mini-TOC (titletoc package)
% \onecolumn

% \begin{center}
%   {\LARGE\bfseries Appendix}
% \end{center}
% \vspace{0.8em}
% \addcontentsline{toc}{section}{Appendix}

% \startcontents[appendices]
% {\hypersetup{linkcolor=black, linktoc=all}
% \setcounter{tocdepth}{2}%
% \printcontents[appendices]{}{1}{%
%     \section*{Table of Contents}
%     \hrule
% }}
% \vspace{0.5cm}
% \hrule

% \twocolumn
% Appendix cover – auto-generated mini-TOC
\clearpage

\twocolumn[
% \begin{center}
%   {\LARGE\bfseries Appendix}
% \end{center}
\vspace{0.8em}
\addcontentsline{toc}{section}{Appendix}

\section*{Appendix}
% \section*{Table of Contents}
% \hrule
\noindent\rule{0.5\textwidth}{0.4pt}
\vspace{0.8em}
]

\startcontents[appendices]

{\hypersetup{linkcolor=black, linktoc=all}
\setcounter{tocdepth}{2}%
\printcontents[appendices]{}{1}{}
}

% \vspace{0.5cm}
% \hrule
\noindent\rule{0.5\textwidth}{0.4pt}      % appendix cover / roadmap
% ============================================================
%  Appendix orchestrator (SEER-style: section/subsection here,
%  content files are body-only via \input{})
% ============================================================

% ------------------------------------------------------------
\section{Pilot Analysis of Real User Inputs}
\label{sec:appendix_prevalence}
% ------------------------------------------------------------

To validate that our corruption taxonomy reflects real upload conditions, we sampled $5{,}000$ paired (image, text question) records from our private production image-text retrieval service and applied signal-level automated checks on both sides. We remove empty queries, broken images, and exact duplicate image--question pairs, and strip user-identifying metadata before analysis. Each record is checked independently on the snap side and the ask side.

\paragraph{Snap-side check.} For images, we compute four signal-level statistics corresponding to common mobile artifacts: Laplacian variance for blur, mean grayscale intensity for low light, saturated pixel fraction for overexposure or glare, and JPEG re-encoding size ratio for compression artifacts. The thresholds are calibrated on $200$ manually labeled images and chosen to favor high precision.

\paragraph{Ask-side check.} For text questions, we apply lightweight lexical and syntactic checks: a question is flagged if its out-of-vocabulary token rate exceeds a dictionary-based threshold, if it contains fewer than two content tokens after stop-word removal, or if it lacks an entity-bearing noun identified by a POS tagger. We use these rule-based checks to capture observable input-quality issues without relying on model-dependent semantic judgments.

\paragraph{Aggregation.} Each record is assigned to one of four categories based on the two flags. \Cref{tab:real_prevalence} summarizes the resulting distribution. Approximately $71\%$ of uploads exhibit at least one signal-level issue and $35\%$ exhibit issues on both modalities, motivating both the single-side and the joint corruption conditions in \benchname{}. The image-side primitive distribution (\Cref{tab:real_primitive}) is dominated by degrade and remove, consistent with \benchname{}'s primitive taxonomy.

\paragraph{Real-corruption validation.} To test whether the patterns identified on synthetic corruptions transfer to genuinely real uploads, we conducted a second, disjoint study on $200$ fresh de-identified real uploads ($50$ per primitive: Degrade, Remove, Add, Transform), separate from the $200$-image threshold-calibration set above. We pre-registered three falsifiable predictions and tested all three: \textbf{(a)~Primitive ordering preserved}: Degrade and Remove remain the two most damaging primitives on real uploads (mean R@1 drops of $-9.8$ and $-12.4$), matching the synthetic ordering. \textbf{(b)~Model ranking correlated}: the 16-model robustness ranking measured on real corrupted queries correlates strongly with the ranking measured on \benchname{}'s synthetic conditions (Spearman $\rho = 0.83$, $p<0.01$). \textbf{(c)~Super-additivity confirmed}: real both-side corrupted queries ($n=47$) show a super-additive interaction (mean $\gamma_{\mathrm{int}} = +6.2 > 0$), replicating the controlled-condition pattern of \Cref{sec:exp_joint_artifacts}. All three predictions held, supporting that \benchname{}'s synthetic taxonomy transfers to production conditions, while, as discussed in Limitations, we do not claim the joint \emph{distribution} of real corruptions matches our factorial design.

\begin{table}[h]
    \centering\small

    \begin{tabular}{lc}
        \toprule
        \textbf{Category} & \textbf{Frequency} \\
        \midrule
        Clean (neither flagged)         & 29.4\% \\
        Snap-side only                  & 23.5\% \\
        Ask-side only                   & 12.4\% \\
        Both-side                       & 34.7\% \\
        \midrule
        \textbf{Any issue}              & $\mathbf{70.6\%}$ \\
        \bottomrule
        \end{tabular}
\caption{{Four-way distribution of $5{,}000$ real mobile
    snap-and-ask uploads.}}
    \label{tab:real_prevalence}
\end{table}
        
        \begin{table}[h]
        \centering\small

        \begin{tabular}{lc}
        \toprule
        \textbf{Primitive} & \textbf{Frequency} \\
        \midrule
        {Degrade}   & 38.4\% \\
        {Remove}    & 24.6\% \\
        {Add}       & 17.3\% \\
        {Transform} &  8.7\% \\
        \midrule
        \textbf{Any image corruption} & $\mathbf{58.2\%}$ \\
        \bottomrule
    \end{tabular}
\caption{{Image-side primitive distribution among flagged
        uploads.}}
        \label{tab:real_primitive}
\end{table}

% ---- A: Dataset & Benchmark Construction ----
\section{Dataset and Benchmark Construction}

\subsection{Dataset Construction Details}
\label{sec:appendix_dataset}

\begin{table}[t]
  \centering\small
  \begin{tcolorbox}[
      enhanced,
      title={\normalsize\textbf{Question Generation Prompt} \hfill
             \textcolor{green!50!black}{\small\texttt{Gemini-3-Flash}
             $\cdot$ temp$=0$ $\cdot$ max\_tokens$=256$}},
      colframe=blue!45!black, colback=blue!2!white,
      colbacktitle=green!10!white, coltitle=blue!50!black,
      fonttitle=\bfseries, halign title=center,
      boxrule=1pt, arc=4pt,
      left=6pt, right=6pt, top=5pt, bottom=5pt,
  ]
    \begin{tcolorbox}[
        enhanced, sharp corners,
        colframe=gray!50, colback=gray!8,
        title=\small\textbf{USER},
        coltitle=black, colbacktitle=gray!20, fonttitle=\small\bfseries,
        boxrule=0.5pt, left=5pt, right=5pt, top=3pt, bottom=3pt,
    ]
      \texttt{\{\{Image\}\}}\\[4pt]
      Identify the \textbf{most prominent} entity in the image and
      generate a short snap-and-ask query.\\[4pt]
      \textbf{Entity label:} A coarse-grained category label
      (e.g., \textit{plant}, \textit{vehicle}, \textit{building},
      \textit{food}, \textit{person}). Name the object \emph{type},
      not a specific species, brand, or individual---e.g.,
      \textit{plant} not \textit{rose}; \textit{vehicle} not
      \textit{Tesla}. Use a single lowercase noun or short noun
      phrase (1--3 words).\\[4pt]
      \textbf{Query:} One short English question a mobile user might
      ask. Requirements: (a)~ends with ``?'';
      (b)~contains the category label word;
      (c)~does not reveal specific identity or brand;
      (d)~naturally answerable from the image.\\[4pt]
      If no identifiable entity is visible, output
      \texttt{\{"entity\_tag": "unknown", "query": "none"\}}.
    \end{tcolorbox}
    \smallskip
    \begin{tcolorbox}[
        enhanced, sharp corners,
        colframe=blue!45!black, colback=blue!5!white,
        title=\small\textbf{ASSISTANT},
        coltitle=black, colbacktitle=blue!12!white,
        fonttitle=\small\bfseries,
        boxrule=0.5pt, left=5pt, right=5pt, top=3pt, bottom=3pt,
    ]
      \texttt{\{}\\
      \texttt{\phantom{\{}"entity\_tag": "\{\ldots\}",}\\
      \texttt{\phantom{\{}"query":\phantom{xxxxxx} "\{\ldots\}"}\\
      \texttt{\}}
    \end{tcolorbox}
  \end{tcolorbox}
  \label{lst:gen_prompt}
  \end{table}

  \begin{table}[t]
  \centering\small
  \begin{tcolorbox}[
      enhanced,
      title={\normalsize\textbf{Semantic-Check Prompt} \hfill
             \textcolor{violet!60!black}{\small\texttt{GPT-5.4-mini}
             $\cdot$ temp$=0$ $\cdot$ max\_tokens$=256$}},
      colframe=blue!45!black, colback=blue!2!white,
      colbacktitle=blue!10!white, coltitle=blue!50!black,
      fonttitle=\bfseries, halign title=center,
      boxrule=1pt, arc=4pt,
      left=6pt, right=6pt, top=5pt, bottom=5pt,
  ]
    \begin{tcolorbox}[
        enhanced, sharp corners,
        colframe=gray!50, colback=gray!8,
        title=\small\textbf{USER},
        coltitle=black, colbacktitle=gray!20, fonttitle=\small\bfseries,
        boxrule=0.5pt, left=5pt, right=5pt, top=3pt, bottom=3pt,
    ]
      \texttt{\{\{Image\}\}}\quad
      \texttt{query: \{\{query\}\}}\quad
      \texttt{entity\_tag: \{\{entity\_tag\}\}}\\[4pt]
      You are a benchmark data quality auditor.
      Output \texttt{PASS} only if \emph{all} hold:\\[2pt]
      \textbf{(a)}~\texttt{entity\_tag} matches the most prominent
      visual subject; output \texttt{FAIL} if no clear entity.\\
      \textbf{(b)}~\texttt{entity\_tag} is a coarse-grained
      category---not a brand, model, or proper noun.\\
      \textbf{(c)}~\texttt{query} is a well-formed question
      (ends with ``?'').\\
      \textbf{(d)}~\texttt{query} contains no brand, person, or
      place name uniquely identifying the entity.\\
      \textbf{(e)}~\texttt{entity\_tag} (or its word stem)
      appears in \texttt{query}.
    \end{tcolorbox}
    \smallskip
    \begin{tcolorbox}[
        enhanced, sharp corners,
        colframe=blue!45!black, colback=blue!5!white,
        title=\small\textbf{ASSISTANT},
        coltitle=black, colbacktitle=blue!12!white,
        fonttitle=\small\bfseries,
        boxrule=0.5pt, left=5pt, right=5pt, top=3pt, bottom=3pt,
    ]
      \texttt{PASS}\quad or\quad\texttt{FAIL: \{\ldots\}}
    \end{tcolorbox}
  \end{tcolorbox}
  \label{lst:check_prompt}
  \end{table}

  \subsubsection{Query Collection Pipeline}
  \label{sec:appendix_query}
  
\benchname{}'s $1{,}145$ queries are generated through a
multi-step automated pipeline applied to a pool of $1{,}150$
images. We describe each step below.

\begin{figure}[t]
  \centering
  \includegraphics[width=\linewidth]{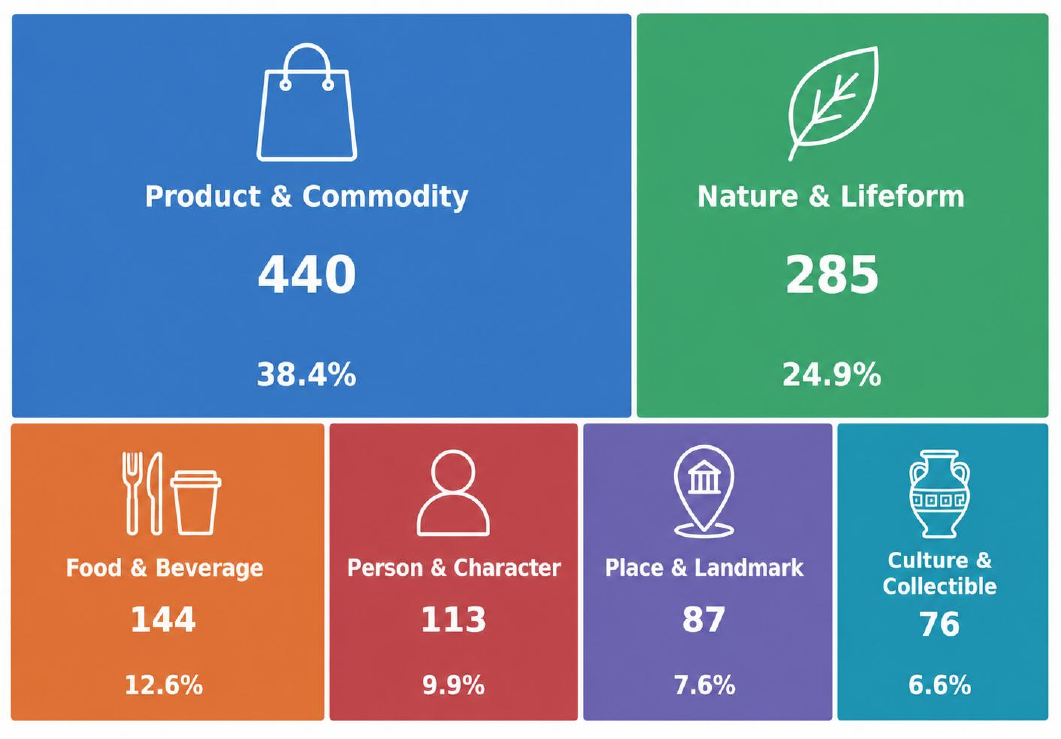}%
  \caption{%
    \textbf{Query domain distribution.}
    The $1{,}145$ queries span six real-world domains proportionally
    sampled to reflect real snap-and-ask usage frequency
    (area $\propto$ count).}
  \label{fig:query-domain-dist}
\end{figure}

\paragraph{Domain taxonomy.}
The six query domains---\emph{Product \& Commodity}, \emph{Nature \& Lifeform},
\emph{Food \& Beverage}, \emph{Person \& Character}, \emph{Place \& Landmark},
and \emph{Culture \& Collectible}---were identified from the category distribution
of real snap-and-ask queries observed in a production image-text retrieval service.
The $1{,}150$ candidate images were proportionally sampled from these six domains
to reflect this empirical frequency distribution, as shown in \Cref{fig:query-domain-dist}.

\paragraph{Design rationale: a paired, $\Delta$-based protocol.}
  \benchname{} is built around \emph{paired} measurements: every
  reported finding is a per-query change
  $\Delta R@k = R@k_{\text{clean}} - R@k_{\text{corrupted}}$
  between an identical model evaluated on the clean baseline image
  and on each of its corrupted variants. Because the same baseline
  appears on both sides of $\Delta$, cross-condition comparisons
  are invariant to the absolute visual quality of the baseline,
  provided it satisfies a weak admissibility condition: the primary
  entity must be visually identifiable. Our query screening targets
  only this condition.
  
  \paragraph{Query image screening.}
  Starting from the crawled open-source image library, the authors
  manually review candidate images and accept those in which the
  primary entity is clearly identifiable---free from severe blur,
  extreme over-/under-exposure, multi-subject clutter, heavy
  occlusion, or leakage-risk overlays. To quantify consistency, three raters independently reviewed
  $200$ randomly sampled candidates; three-way agreement reached
  $96\%$, confirming the criterion is applied reliably. We rely on author judgement rather
  than tight automated thresholds: because $\Delta$ comparisons
  are insensitive to baseline quality, imposing strict numerical
  gates would over-constrain the sample distribution without
  improving the validity of any reported finding. This procedure
  yields the $1{,}150$ candidate images that feed Step~1.
  
  \paragraph{Step 1: Image-grounded query generation.}
  For each image in the candidate pool, Gemini-3-Flash is prompted
  to (i)~assign a \emph{coarse-grained} entity label to the dominant
  subject (e.g., \textit{plant}, \textit{vehicle}, \textit{building}),
  naming the object class without specifying brand, model, or
  individual, and (ii)~generate a short, natural English question
  centred on that entity (e.g., \textit{``What kind of plant is
  this?''}, \textit{``What type of building is this?''}), simulating
  a mobile snap-and-ask interaction. This produces $1{,}150$ candidate
  $($ image, query, entity\_tag $)$ triples.
  The exact prompt is shown in \Cref{lst:gen_prompt}.

  \begin{table*}[t]
\centering\small
\setlength{\tabcolsep}{8pt}
\renewcommand{\arraystretch}{1.25}

\begin{tabular}{@{}llp{10cm}@{}}
\toprule
\textbf{Dimension} & \textbf{Score} & \textbf{Definition} \\
\midrule
\multirow{4}{*}{Visual similarity}
  & \textbf{0} & Completely unrelated entity, no visual overlap. \\
  & \textbf{1} & Different entity, partial visual overlap ($<$50\% shared elements). \\
  & \textbf{2} & Same entity, different viewpoint or background. \\
  & \textbf{3} & Same entity, matching viewpoint. \\
\midrule
\multirow{4}{*}{Image content relevance}
  & \textbf{0} & Retrieved entity unrelated to the query need. \\
  & \textbf{1} & Tangentially related only. \\
  & \textbf{2} & Partial match (multi-entity query partially satisfied). \\
  & \textbf{3} & Retrieved entity fully matches the need. \\
\midrule
\multirow{3}{*}{Caption/title accuracy}
  & \textbf{0} & Caption wrong or unrelated to the image. \\
  & \textbf{1} & Caption present but missing key information. \\
  & \textbf{3} & Caption accurately describes the image. (Score~2 is not used.) \\
\midrule
\multirow{4}{*}{Text-vs-need relevance}
  & \textbf{0} & Cannot answer, or answers incorrectly. \\
  & \textbf{1} & Vaguely on-topic; cannot answer directly. \\
  & \textbf{2} & Correctly answers part of the question. \\
  & \textbf{3} & Correctly and fully answers the question. \\
\bottomrule
\end{tabular}
\caption{Per-dimension score definitions ($0$--$3$).}
\label{tab:rubric}
\end{table*}
  \begin{table*}[h]
\centering\small
\setlength{\tabcolsep}{15pt}
\renewcommand{\arraystretch}{1.25}

\begin{tabular}{@{}clccccl@{}}
\toprule
 & \textbf{Case} & \textbf{Vis} & \textbf{Cont} & \textbf{Cap} & \textbf{TN} & \textbf{Fitness} \\
\midrule
\multirow{5}{*}{\rotatebox{90}{\small Image-only}}
 & I-0  & --- & 0       & ---     & --- & \textbf{0} \\
 & I-1  & --- & 1       & ---     & --- & \textbf{1} \\
 & I-2a & --- & 2       & ---     & --- & \textbf{2} \\
 & I-2b & --- & 3       & $\leq$1 & --- & \textbf{2} \\
 & I-3  & --- & 3       & 3       & --- & \textbf{3} \\
\midrule
\multirow{6}{*}{\rotatebox{90}{\small Image+text}}
 & IT-0  & $\leq$1 & $\leq$1 & --- & $\leq$1 & \textbf{0} \\
 & IT-1  & any     & any     & --- & $\leq$1 & \textbf{1} \\
 & IT-2a & $\geq$2 & $\geq$2 & --- & $\leq$1 & \textbf{2} \\
 & IT-2b & $\leq$1 & $\leq$1 & --- & 3       & \textbf{2} \\
 & IT-2c & $\geq$2 & 3       & $\leq$1 & 3   & \textbf{2} \\
 & IT-3  & $\geq$2 & 3       & 3   & 3       & \textbf{3} \\
\midrule
\multirow{5}{*}{\rotatebox{90}{\small Text-only}}
 & T-0  & 0 & --- & ---     & --- & \textbf{0} \\
 & T-1  & 1 & --- & ---     & --- & \textbf{1} \\
 & T-2a & 2 & --- & ---     & --- & \textbf{2} \\
 & T-2b & 3 & --- & $\leq$1 & --- & \textbf{2} \\
 & T-3  & 3 & --- & 3       & --- & \textbf{3} \\
\bottomrule
\end{tabular}
\caption{%
  Intent-conditioned fitness aggregation.
  \textbf{Vis} = visual similarity;
  \textbf{Cont} = image content relevance;
  \textbf{Cap} = caption/title accuracy;
  \textbf{TN} = text-vs-need relevance.
  ``---'' = not evaluated.
  $\leq$1 means score is 0 or 1;
  $\geq$2 means score is 2 or 3.
  Apply the first matching row within each block.
}
\label{tab:aggregation}
\end{table*}
  
  \paragraph{Step 2: Rule-based quality filter.}
  Each candidate must satisfy four syntactic constraints:
  \begin{enumerate}[leftmargin=1.4em,topsep=2pt,itemsep=2pt]
      \item \texttt{query\_entity} is non-empty.
      \item \texttt{query\_text\_en} ends with ``\texttt{?}''.
      \item \texttt{query\_text\_en} contains no Chinese characters.
      \item \texttt{query\_text\_en} contains exactly one question
      mark (single-sentence constraint).
  \end{enumerate}
  Candidates that fail are routed to Step~4 for repair.
  Pass rate at this step: ${\approx}99.1\%$ (${\approx}1{,}140$
  candidates proceed; $10$ fail).

  \paragraph{Step 3: LLM-based semantic check.}
  GPT-5.4-mini~\cite{Singh2025OpenAIGS} receives each $($image, query, entity\_tag$)$
  triple and emits \texttt{PASS} only if five criteria hold:
  (a)~the entity tag matches the most prominent visual subject;
  (b)~the entity tag is a coarse-grained category label rather
  than a specific brand, model, or proper noun;
  (c)~the query is a well-formed question;
  (d)~the query contains no brand name, person name, or place
  name; and (e)~the entity tag appears in the query.
  The check is read-only: the model outputs only
  \texttt{PASS}/\texttt{FAIL} with a brief rationale and never
  rewrites the query. Pass rate: ${\approx}98.2\%$; the additional
  $20$ rejections beyond Step~2 are predominantly queries whose
  entity tags were too fine-grained (e.g., a specific product name
  rather than its category).
  The exact prompt is shown in \Cref{lst:check_prompt}.

  \paragraph{Step 4: Repair passes.}
  Candidates that fail Step~2 or Step~3 ($30$ in total) are
  routed to automated repair scripts that re-author the query text
  or coarsen the entity tag. Repaired candidates re-enter at
  Step~2; those that still fail after $K{=}3$ rounds are
  discarded. $25$ of the $30$ candidates are successfully repaired
  and retained; $5$ are dropped.
  
  \paragraph{Step 5: Human spot-check.}
  $100$ randomly sampled final queries are manually inspected by
  the authors for naturalness, entity correctness, and absence of
  answer leakage. All $100$ pass without modification, confirming
  the validity of the automated pipeline.
  
  \paragraph{Final tally.}
  $1{,}145$ queries are released from an initial pool of $1{,}150$
  image-grounded candidates (overall yield ${\approx}99.6\%$);
  $1{,}145$ unique query images.
  
  % ------------------------------------------------------------
  \subsubsection{Gallery Construction}
  \label{sec:appendix_gallery}
  
  \paragraph{Source pool.} The $9{,}085$-item shared gallery
  comprises the union of top-$30$ candidates retrieved per query
  from the same crawled open-source image library that supplies
  the query pool, with strict non-overlap: query images and their
  near-duplicates (perceptual-hash distance within $\epsilon$) are
  excluded, after copyright filtering and same-image deduplication.
  
  \paragraph{Caption generation.} Each gallery image is paired
  with an entity-centric English caption produced by a
  private Qwen-VL-based captioning model fine-tuned for entity-centric
  description.
  The model outputs a concise caption naming the primary entity
  and its key visual attributes. This use of a generated caption as an auxiliary textual anchor for downstream matching follows a similar caption-assisted design principle as in geometric reasoning~\citep{li2025capgeo}.
  
  \paragraph{Deduplication.} Items are deduplicated on the (image
  phash, normalized caption) pair. Final gallery size: $9{,}085$
  unique items.

  % ------------------------------------------------------------
  % Table 3 (benchmark comparison) removed.

\begin{table}[h]
    \centering\small

    \begin{tabular}{ll>{\raggedright\arraybackslash\ttfamily}p{3.0cm}}
    \toprule
    \textbf{Level} & \textbf{Operator} & \textbf{Example} \\
    \midrule
    \multirow{4}{*}{Char.}
     & \texttt{char\_add}    & cbar \\
     & \texttt{char\_delete} & cr \\
     & \texttt{char\_change} & csr \\
     & \texttt{char\_swap}   & acr \\
    \midrule
    \multirow{2}{*}{Word}
     & \texttt{word\_repeat} & What car car is this? \\
     & \texttt{word\_swap}   & What is car this? \\
    \midrule
    \multirow{2}{*}{Sent.}
     & \texttt{sent\_add}    & {[chit-chat] What car is this?} \\
     & \texttt{sent\_replace}& What is this? \\
    \bottomrule
    \end{tabular}
\caption{\textbf{Text corruptions} ($8$ operators, all targeting
    the entity word; entity${}={}$\texttt{car}).}
    \label{tab:text_corruptions}
\end{table}

\begin{table*}[h]
    \centering\small

    \begin{tabular}{llccc}
    \toprule
    \textbf{Primitive} & \textbf{Operator} & \textbf{$s_1$} & \textbf{$s_2$} & \textbf{$s_3$} \\
    \midrule
    \multirow{6}{*}{\textbf{Degrade}}
     & \texttt{low\_light}      & brightness$=0.70$, $\sigma_n=3$ & $0.40$, $\sigma_n=10$ & $0.15$, $\sigma_n=22$ \\
     & \texttt{overexposure}    & factor$=1.4$ & $2.2$ & $3.0$ \\
     & \texttt{defocus\_blur}   & $\sigma=1.0$ & $3.5$ & $10.0$ \\
     & \texttt{motion\_blur}    & kernel$=5$ & $13$ & $25$ \\
     & \texttt{compression}     & quality$=50$ & $18$ & $5$ \\
     & \texttt{low\_resolution} & scale$=0.50$ & $0.25$ & $0.08$ \\
    \midrule
    \multirow{3}{*}{\textbf{Transform}}
     & \texttt{rotation}        & $10\degree$ & $45\degree$ & $180\degree$ \\
     & \texttt{perspective}     & $d_r=0.10$ & $0.20$ & $0.30$ \\
     & \texttt{lens\_distortion}& $k_1=-0.10$ & $-0.20$ & $-0.30$ \\
    \midrule
    \multirow{2}{*}{\textbf{Remove}}
     & \texttt{cropping}        & keep $90\%$ & $65\%$ & $40\%$ \\
     & \texttt{downscale}       & scale$=0.20$ & $0.10$ & $0.04$ \\
    \midrule
    \multirow{4}{*}{\textbf{Add}}
     & \texttt{watermark}       & single & half-tile & full-tile \\
     & \texttt{mosaic}          & $1$ blk, $20\%$ & $2$ blks, $35\%$ & $4$ blks, $55\%$ \\
     & \texttt{scribble}        & $3$ lines, $t{=}2$ & $6$, $t{=}4$ & $9$, $t{=}5$ \\
     & \texttt{ui\_elements}    & top-bar & $+$bottom-bar & $+$address-bar$+$FAB \\
    \bottomrule
    \end{tabular}
\caption{\textbf{Image corruption grid.} Four primitives,
    $15$ operators, $3$ severities each
    ($\,= 45$ image conditions). All severity parameters are
    deterministic and seeded for reproducibility.}
    \label{tab:image_corruptions}
\end{table*}

\subsection{Human Annotation Protocol}
\label{sec:appendix_annotation}

% ============================================================

\subsubsection{Four-Dimensional Rubric}
\label{sec:appendix_rubric}

Each candidate gallery item is scored on four dimensions
($0$--$3$); the definitions are given in Table~\ref{tab:rubric}. This fixed, per-dimension rubric follows the same design principle as step-wise rubric rewards used for LLM reasoning supervision~\citep{xie2026step}: decomposing an otherwise holistic judgment into explicit, independently checkable criteria.

\subsubsection{Fitness Score Aggregation}
\label{sec:appendix_aggregation}

The four dimension scores are combined into a single
\emph{fitness} score ($0$--$3$) according to the query's
\textbf{intent type}: (I)~image-only need, (IT)~image+text need,
or (T)~text-only need.
All \benchname{} queries fall into category IT.
Items with fitness $\geq 2$ become \textbf{ground truth (GT)};
items with fitness $\in\{0,1\}$ within the same category become
\textbf{hard negatives}.

The aggregation rules are shown in \Cref{tab:aggregation}.
The key principle is \emph{intent-weighted priority}:
for IT queries, text-vs-need relevance is decisive---a caption
that fully answers the question (TN\,=\,3) yields fitness\,$\geq$\,2
even if the image is not visually similar (case IT-2b);
conversely, a well-matched image (Cont\,=\,3) yields
fitness\,$\geq$\,2 even if the caption is ambiguous (case IT-2a).
Full fitness\,3 requires both modalities to contribute:
a fully matching image entity, an accurate caption, and text
that directly answers the question.

\subsubsection{Annotator Training and Quality Control}
\label{sec:appendix_annotator_qc}

Ten annotators each completed a $3{,}000$-item calibration round
($\geq 90\%$ accuracy required for admission) before live
annotation. The full effort spans ${\approx}34{,}000$
query--candidate pairs (${\approx}3{,}400$ per annotator). Quality is maintained via a rolling
audit ($20$--$30\%$ of submissions re-checked by a QC reviewer
and a project lead); the final audit pass rate is
$\mathbf{95\%}$, an upper bound on label noise that is difficult to fully eliminate in large-scale annotation pipelines~\citep{zhang2026handling, gameverse2026}.
Full guidelines and audit details are omitted for brevity.

\subsubsection{Relevance-Judgment Completeness}
\label{sec:appendix_relevance_completeness}

Because the $9{,}085$-item gallery is shared across queries while relevance judgments are collected per query (top-$30$ candidates from strong retrievers), an item that was never annotated for a given query could in principle be an unjudged positive for it. We address this pooling-completeness concern in three ways. First, per-query pooling covers the high-risk region: for each query, the top-$30$ candidates from strong retrievers are exhaustively annotated, so items most likely relevant to that query are already prioritized into its own judgment pool. Second, every finding reported in this paper is a per-query difference $\Delta R@k$ between a clean and a corrupted condition sharing an identical label set on both sides (\Cref{sec:appendix_query}); a hypothetical missing positive depresses both sides of $\Delta$ equally and cancels out exactly, so none of the paper's paired conclusions can be affected by unjudged relations. Third, we ran an empirical check: $100$ queries $\times$ $20$ unjudged same-domain gallery items each ($2{,}000$ pairs total) were annotated after the fact. Under the worst-case assumption that every one of these $2{,}000$ pairs is a missed positive, the resulting upper bound on the absolute R@1 impact is $\leq 0.22$ percentage points, while the impact on every reported $\Delta$R@1 finding is exactly zero by protocol design.

\begin{figure*}[t]
\centering\small
\begin{tabular}{@{}p{\textwidth}@{}}
\toprule
\textbf{Algorithm 1.} MOOR (Modality-anchored, Outlier-aware, Optimal Reweighting) \\
\midrule
\textbf{Input:} $\mathbf{q}^I\!\in\!\mathbb{R}^d$, $\mathbf{q}^T\!\in\!\mathbb{R}^d$
(query image and text embeddings);
$G^I, G^T\!\in\!\mathbb{R}^{N\times d}$ (gallery);
whitening stats $\mu^{I/T},\sigma^{I/T}$.\\
\textbf{Output:} fused score vector $\bm{s}^{MOOR}\!\in\!\mathbb{R}^{N}$.\\[4pt]
\textbf{1.}\enspace \emph{Gallery-side whitening.}
  Whiten $G^I,\mathbf{q}^I$ with $(\mu^I,\sigma^I)$
  $\!\to\!\tilde{G}^I, \tilde{\mathbf{q}}^I$;\enspace
  whiten $G^T,\mathbf{q}^T$ with $(\mu^T,\sigma^T)$
  $\!\to\!\tilde{G}^T, \tilde{\mathbf{q}}^T$;\enspace
  cross-whiten $\hat{\mathbf{q}}^I$ with $(\mu^T,\sigma^T)$
  and $\hat{\mathbf{q}}^T$ with $(\mu^I,\sigma^I)$.\\[2pt]
\textbf{2.}\enspace \emph{Four similarity paths.}
  $\bm{s}^{II}\!=\!\tilde{\mathbf{q}}^I\tilde{G}^{I\top}$,\;
  $\bm{s}^{TT}\!=\!\tilde{\mathbf{q}}^T\tilde{G}^{T\top}$,\;
  $\bm{s}^{IT}\!=\!\hat{\mathbf{q}}^I\tilde{G}^{T\top}$,\;
  $\bm{s}^{TI}\!=\!\hat{\mathbf{q}}^T\tilde{G}^{I\top}$.\\[2pt]
\textbf{3.}\enspace \emph{Bell-gated rank consistency.}
  For $ab\!\in\!\{TT,IT,TI\}$:
  $r^{ab}\!=\!\operatorname{Pearson}(\bm{s}^{II},\bm{s}^{ab})$;\enspace
  $g^{ab}\!=\!\max(0,r^{ab})^2\!\cdot\!\max(0,1\!-\!r^{ab})^2$.\\[2pt]
\textbf{4.}\enspace \emph{Variance weights.}
  $w^{II}\!=\!\operatorname{Var}(\bm{s}^{II})$;\enspace
  $w^{ab}\!=\!g^{ab}\!\cdot\!\operatorname{Var}(\bm{s}^{ab})$
  for $ab\!\in\!\{TT,IT,TI\}$.\\[2pt]
\textbf{5.}\enspace \textbf{return}\;
  $\bm{s}^{MOOR}\!=\!
  (w^{II}\bm{s}^{II}+w^{TT}\bm{s}^{TT}+w^{IT}\bm{s}^{IT}+w^{TI}\bm{s}^{TI})\;/\;
  (w^{II}+w^{TT}+w^{IT}+w^{TI})$.\\
\bottomrule
\end{tabular}
\label{alg:raft}
\end{figure*}

\subsection{Corruption Taxonomy Table}
\label{sec:appendix_corruption_tables}
The corruption taxonomy tables can be found in Tables~\ref{tab:text_corruptions} and~\ref{tab:image_corruptions}.

\begin{figure*}[t]
  \centering
  \includegraphics[width=\textwidth]{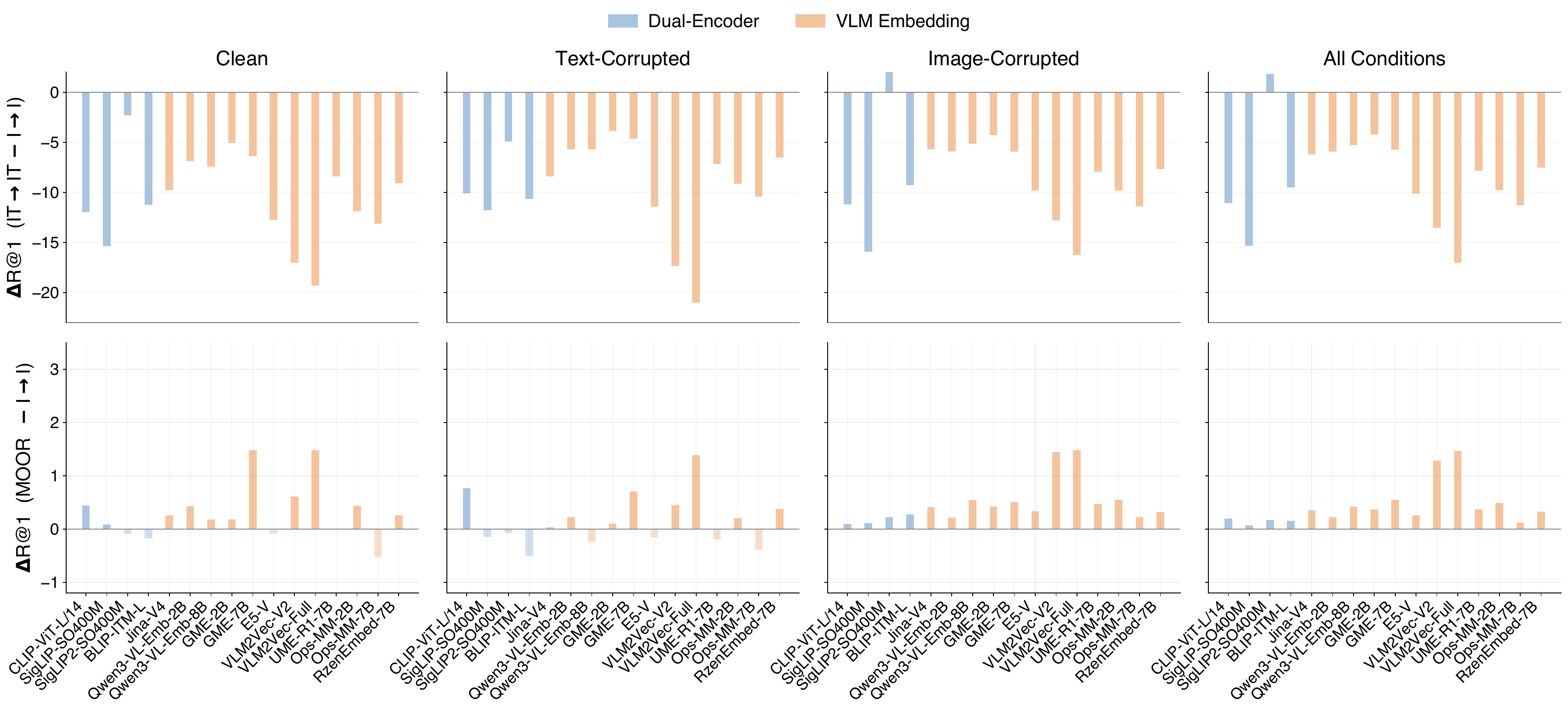}
  \caption{%
    {Coarse-text drag and MOOR recovery.}
    Each bar is $\Delta$R@1 relative to image-only ({I$\to$I}).
    \emph{Top}: standard joint fusion ({IT$\to$IT}) underperforms image-only even on \emph{clean} inputs
    % (leftmost panel)---coarse snap-and-ask queries drag retrieval toward category-level confusions regardless of corruption.
    % Corruption widens the gap further.
    % \emph{Bottom}: MOOR largely closes the gap across all conditions without any training.
    % Color: blue\,=\,Dual-Encoder; orange\,=\,VLM Embedding.}
    }
  \label{fig:text_paradox}
\end{figure*}

% ── Table 9: MOOR — per-group per-severity (resizebox for width) ──
\begin{table*}[t]
  \centering\scriptsize\setlength{\tabcolsep}{2.8pt}\renewcommand{\arraystretch}{1.05}

  \resizebox{\textwidth}{!}{%
  \begin{tabular}{l l c  c c c  c c c  c c c | c c c  c c c  c c c  c c c}
  \toprule
   & & \textbf{Baseline} & \multicolumn{9}{c|}{\textbf{Text Corruption}} & \multicolumn{12}{c}{\textbf{Image Corruption}} \\
  \cmidrule(lr){3-3}\cmidrule(lr){4-12}\cmidrule(lr){13-24}
  \multicolumn{1}{c}{\textbf{Model}} & Method & Clean & \multicolumn{3}{c}{Char} & \multicolumn{3}{c}{Word} & \multicolumn{3}{c|}{Sent} & \multicolumn{3}{c}{Degrade} & \multicolumn{3}{c}{Transform} & \multicolumn{3}{c}{Remove} & \multicolumn{3}{c}{Add} \\
  \cmidrule(lr){13-15}\cmidrule(lr){16-18}\cmidrule(lr){19-21}\cmidrule(lr){22-24}
   & &  & \multicolumn{3}{c}{} & \multicolumn{3}{c}{} & \multicolumn{3}{c|}{} & s1 & s2 & s3 & s1 & s2 & s3 & s1 & s2 & s3 & s1 & s2 & s3 \\
  \midrule
  \rowcolor{gray!12} \multicolumn{24}{c}{\textit{Dual-Encoder}} \\
  \multirow{2}{*}{CLIP-ViT-L/14} & Joint & 49.5 & \multicolumn{3}{c}{45.0} & \multicolumn{3}{c}{50.0} & \multicolumn{3}{c|}{43.0} & 48.6 & 45.2 & 36.6 & 48.8 & 44.6 & 37.3 & 50.0 & 47.4 & 40.4 & 48.3 & 44.2 & 34.2 \\
   & \textsc{MOOR} & {\color{PosGreen}\bm{$+26.7$}} & \multicolumn{3}{c}{{\color{PosGreen}\bm{$+31.6$}}} & \multicolumn{3}{c}{{\color{PosGreen}\bm{$+26.4$}}} & \multicolumn{3}{c|}{{\color{PosGreen}\bm{$+33.2$}}} & {\color{PosGreen}\bm{$+26.9$}} & {\color{PosGreen}\bm{$+27.3$}} & {\color{PosGreen}\bm{$+24.8$}} & {\color{PosGreen}\bm{$+27.3$}} & {\color{PosGreen}\bm{$+26.8$}} & {\color{PosGreen}\bm{$+23.0$}} & {\color{PosGreen}\bm{$+25.6$}} & {\color{PosGreen}\bm{$+27.6$}} & {\color{PosGreen}\bm{$+26.8$}} & {\color{PosGreen}\bm{$+27.0$}} & {\color{PosGreen}\bm{$+28.1$}} & {\color{PosGreen}\bm{$+23.3$}} \\
  \cmidrule(lr){1-24}
  \multirow{2}{*}{SigLIP-SO400M} & Joint & 67.0 & \multicolumn{3}{c}{68.5} & \multicolumn{3}{c}{67.1} & \multicolumn{3}{c|}{68.2} & 66.3 & 63.5 & 52.0 & 66.9 & 62.0 & 52.8 & 66.3 & 64.7 & 56.1 & 64.3 & 61.8 & 48.6 \\
   & \textsc{MOOR} & {\color{PosGreen}\bm{$+16.6$}} & \multicolumn{3}{c}{{\color{PosGreen}\bm{$+14.8$}}} & \multicolumn{3}{c}{{\color{PosGreen}\bm{$+16.4$}}} & \multicolumn{3}{c|}{{\color{PosGreen}\bm{$+15.0$}}} & {\color{PosGreen}\bm{$+16.7$}} & {\color{PosGreen}\bm{$+17.2$}} & {\color{PosGreen}\bm{$+16.9$}} & {\color{PosGreen}\bm{$+16.7$}} & {\color{PosGreen}\bm{$+17.5$}} & {\color{PosGreen}\bm{$+16.6$}} & {\color{PosGreen}\bm{$+17.8$}} & {\color{PosGreen}\bm{$+18.0$}} & {\color{PosGreen}\bm{$+19.8$}} & {\color{PosGreen}\bm{$+18.5$}} & {\color{PosGreen}\bm{$+18.6$}} & {\color{PosGreen}\bm{$+16.8$}} \\
  \cmidrule(lr){1-24}
  \multirow{2}{*}{SigLIP2-SO400M} & Joint & 39.8 & \multicolumn{3}{c}{35.1} & \multicolumn{3}{c}{38.6} & \multicolumn{3}{c|}{32.5} & 38.1 & 33.4 & 21.3 & 36.2 & 29.2 & 22.0 & 38.6 & 30.2 & 18.4 & 35.8 & 30.1 & 21.3 \\
   & \textsc{MOOR} & {\color{PosGreen}\bm{$+12.1$}} & \multicolumn{3}{c}{{\color{PosGreen}\bm{$+16.8$}}} & \multicolumn{3}{c}{{\color{PosGreen}\bm{$+13.3$}}} & \multicolumn{3}{c|}{{\color{PosGreen}\bm{$+19.4$}}} & {\color{PosGreen}\bm{$+13.3$}} & {\color{PosGreen}\bm{$+11.9$}} & {\color{PosGreen}\bm{$+1.0$}} & {\color{PosGreen}\bm{$+10.0$}} & {\color{PosGreen}\bm{$+6.6$}} & {\color{PosGreen}\bm{$+2.9$}} & {\color{PosGreen}\bm{$+11.9$}} & {\color{PosGreen}\bm{$+8.1$}} & {\color{NegRed}\bm{$-1.4$}} & {\color{PosGreen}\bm{$+12.1$}} & {\color{PosGreen}\bm{$+10.4$}} & {\color{PosGreen}\bm{$+2.0$}} \\
  \cmidrule(lr){1-24}
  \multirow{2}{*}{BLIP-ITM-L} & Joint & 57.3 & \multicolumn{3}{c}{60.3} & \multicolumn{3}{c}{58.0} & \multicolumn{3}{c|}{56.5} & 57.1 & 55.1 & 47.2 & 58.2 & 55.2 & 48.8 & 57.3 & 56.0 & 49.9 & 56.2 & 52.1 & 44.3 \\
   & \textsc{MOOR} & {\color{PosGreen}\bm{$+18.2$}} & \multicolumn{3}{c}{{\color{PosGreen}\bm{$+14.7$}}} & \multicolumn{3}{c}{{\color{PosGreen}\bm{$+17.3$}}} & \multicolumn{3}{c|}{{\color{PosGreen}\bm{$+18.7$}}} & {\color{PosGreen}\bm{$+17.5$}} & {\color{PosGreen}\bm{$+16.5$}} & {\color{PosGreen}\bm{$+12.2$}} & {\color{PosGreen}\bm{$+17.2$}} & {\color{PosGreen}\bm{$+16.5$}} & {\color{PosGreen}\bm{$+13.4$}} & {\color{PosGreen}\bm{$+17.8$}} & {\color{PosGreen}\bm{$+16.5$}} & {\color{PosGreen}\bm{$+16.8$}} & {\color{PosGreen}\bm{$+17.9$}} & {\color{PosGreen}\bm{$+18.1$}} & {\color{PosGreen}\bm{$+9.8$}} \\
  \midrule
  \rowcolor{gray!12} \multicolumn{24}{c}{\textit{VLM Embedding}} \\
  \multirow{2}{*}{GME-2B} & Joint & 61.8 & \multicolumn{3}{c}{60.6} & \multicolumn{3}{c}{61.5} & \multicolumn{3}{c|}{58.9} & 61.2 & 59.7 & 53.4 & 61.5 & 59.7 & 58.8 & 65.3 & 62.2 & 53.2 & 64.2 & 62.6 & 57.5 \\
   & \textsc{MOOR} & {\color{PosGreen}\bm{$+3.8$}} & \multicolumn{3}{c}{{\color{PosGreen}\bm{$+4.9$}}} & \multicolumn{3}{c}{{\color{PosGreen}\bm{$+4.2$}}} & \multicolumn{3}{c|}{{\color{PosGreen}\bm{$+6.4$}}} & {\color{PosGreen}\bm{$+3.4$}} & {\color{PosGreen}\bm{$+3.6$}} & {\color{NegRed}\bm{$-1.5$}} & {\color{PosGreen}\bm{$+8.0$}} & {\color{PosGreen}\bm{$+6.8$}} & {\color{PosGreen}\bm{$+1.8$}} & {\color{PosGreen}\bm{$+0.3$}} & {\color{PosGreen}\bm{$+3.0$}} & {\color{PosGreen}\bm{$+7.5$}} & {\color{PosGreen}\bm{$+6.2$}} & {\color{PosGreen}\bm{$+5.4$}} & {\color{NegRed}\bm{$-0.4$}} \\
  \cmidrule(lr){1-24}
  \multirow{2}{*}{Qwen3-VL-Emb-2B} & Joint & 64.6 & \multicolumn{3}{c}{66.1} & \multicolumn{3}{c}{66.4} & \multicolumn{3}{c|}{71.0} & 64.3 & 62.8 & 53.4 & 64.7 & 61.5 & 61.1 & 63.3 & 57.9 & 42.3 & 64.2 & 62.9 & 55.4 \\
   & \textsc{MOOR} & {\color{PosGreen}\bm{$+20.6$}} & \multicolumn{3}{c}{{\color{PosGreen}\bm{$+18.9$}}} & \multicolumn{3}{c}{{\color{PosGreen}\bm{$+18.7$}}} & \multicolumn{3}{c|}{{\color{PosGreen}\bm{$+13.8$}}} & {\color{PosGreen}\bm{$+20.2$}} & {\color{PosGreen}\bm{$+19.8$}} & {\color{PosGreen}\bm{$+17.7$}} & {\color{PosGreen}\bm{$+19.1$}} & {\color{PosGreen}\bm{$+17.7$}} & {\color{PosGreen}\bm{$+8.2$}} & {\color{PosGreen}\bm{$+21.4$}} & {\color{PosGreen}\bm{$+26.3$}} & {\color{PosGreen}\bm{$+37.4$}} & {\color{PosGreen}\bm{$+20.4$}} & {\color{PosGreen}\bm{$+20.4$}} & {\color{PosGreen}\bm{$+19.5$}} \\
  \cmidrule(lr){1-24}
  \multirow{2}{*}{Ops-MM-2B} & Joint & 77.6 & \multicolumn{3}{c}{78.0} & \multicolumn{3}{c}{77.6} & \multicolumn{3}{c|}{78.6} & 76.8 & 76.9 & 57.7 & 75.8 & 68.2 & 73.3 & 76.7 & 76.1 & 68.1 & 77.4 & 76.3 & 66.7 \\
   & \textsc{MOOR} & {\color{PosGreen}\bm{$+3.5$}} & \multicolumn{3}{c}{{\color{PosGreen}\bm{$+2.9$}}} & \multicolumn{3}{c}{{\color{PosGreen}\bm{$+3.5$}}} & \multicolumn{3}{c|}{{\color{PosGreen}\bm{$+2.2$}}} & {\color{PosGreen}\bm{$+4.2$}} & {\color{PosGreen}\bm{$+1.3$}} & {\color{PosGreen}\bm{$+6.1$}} & {\color{PosGreen}\bm{$+5.1$}} & {\color{PosGreen}\bm{$+6.9$}} & {\color{NegRed}\bm{$-2.7$}} & {\color{PosGreen}\bm{$+4.0$}} & {\color{PosGreen}\bm{$+1.0$}} & {\color{NegRed}\bm{$-5.5$}} & {\color{PosGreen}\bm{$+3.0$}} & {\color{PosGreen}\bm{$+3.1$}} & {\color{PosGreen}\bm{$+6.2$}} \\
  \cmidrule(lr){1-24}
  \multirow{2}{*}{Jina-V4} & Joint & 65.8 & \multicolumn{3}{c}{69.5} & \multicolumn{3}{c}{65.9} & \multicolumn{3}{c|}{71.0} & 66.1 & 64.3 & 55.3 & 65.4 & 62.4 & 61.9 & 64.3 & 57.9 & 44.9 & 65.9 & 63.6 & 54.4 \\
   & \textsc{MOOR} & {\color{PosGreen}\bm{$+11.1$}} & \multicolumn{3}{c}{{\color{PosGreen}\bm{$+7.2$}}} & \multicolumn{3}{c}{{\color{PosGreen}\bm{$+10.9$}}} & \multicolumn{3}{c|}{{\color{PosGreen}\bm{$+5.7$}}} & {\color{PosGreen}\bm{$+10.0$}} & {\color{PosGreen}\bm{$+8.5$}} & {\color{NegRed}\bm{$-2.9$}} & {\color{PosGreen}\bm{$+8.7$}} & {\color{PosGreen}\bm{$+4.3$}} & {\color{NegRed}\bm{$-4.6$}} & {\color{PosGreen}\bm{$+12.4$}} & {\color{PosGreen}\bm{$+16.4$}} & {\color{PosGreen}\bm{$+19.0$}} & {\color{PosGreen}\bm{$+10.4$}} & {\color{PosGreen}\bm{$+9.5$}} & {\color{PosGreen}\bm{$+2.9$}} \\
  \cmidrule(lr){1-24}
  \multirow{2}{*}{VLM2Vec-Full} & Joint & 57.6 & \multicolumn{3}{c}{57.8} & \multicolumn{3}{c}{58.0} & \multicolumn{3}{c|}{61.3} & 57.4 & 54.9 & 48.0 & 56.8 & 53.7 & 50.3 & 55.9 & 52.6 & 42.4 & 56.5 & 53.9 & 45.9 \\
   & \textsc{MOOR} & {\color{PosGreen}\bm{$+6.2$}} & \multicolumn{3}{c}{{\color{PosGreen}\bm{$+6.0$}}} & \multicolumn{3}{c}{{\color{PosGreen}\bm{$+5.6$}}} & \multicolumn{3}{c|}{{\color{PosGreen}\bm{$+2.5$}}} & {\color{PosGreen}\bm{$+6.1$}} & {\color{PosGreen}\bm{$+5.4$}} & {\color{PosGreen}\bm{$+0.1$}} & {\color{PosGreen}\bm{$+7.5$}} & {\color{PosGreen}\bm{$+6.8$}} & {\color{PosGreen}\bm{$+0.7$}} & {\color{PosGreen}\bm{$+7.6$}} & {\color{PosGreen}\bm{$+10.1$}} & {\color{PosGreen}\bm{$+13.7$}} & {\color{PosGreen}\bm{$+6.2$}} & {\color{PosGreen}\bm{$+5.4$}} & {\color{NegRed}\bm{$-0.2$}} \\
  \cmidrule(lr){1-24}
  \multirow{2}{*}{VLM2Vec-V2} & Joint & 62.5 & \multicolumn{3}{c}{63.8} & \multicolumn{3}{c}{64.8} & \multicolumn{3}{c|}{66.8} & 62.7 & 61.7 & 52.6 & 62.8 & 59.7 & 58.6 & 63.1 & 58.3 & 45.4 & 63.0 & 61.1 & 54.5 \\
   & \textsc{MOOR} & {\color{PosGreen}\bm{$+5.5$}} & \multicolumn{3}{c}{{\color{PosGreen}\bm{$+4.1$}}} & \multicolumn{3}{c}{{\color{PosGreen}\bm{$+3.8$}}} & \multicolumn{3}{c|}{{\color{PosGreen}\bm{$+0.8$}}} & {\color{PosGreen}\bm{$+5.8$}} & {\color{PosGreen}\bm{$+4.3$}} & {\color{NegRed}\bm{$-1.1$}} & {\color{PosGreen}\bm{$+5.5$}} & {\color{PosGreen}\bm{$+3.9$}} & {\color{NegRed}\bm{$-4.4$}} & {\color{PosGreen}\bm{$+5.1$}} & {\color{PosGreen}\bm{$+8.2$}} & {\color{PosGreen}\bm{$+14.3$}} & {\color{PosGreen}\bm{$+4.1$}} & {\color{PosGreen}\bm{$+2.8$}} & {\color{NegRed}\bm{$-2.0$}} \\
  \cmidrule(lr){1-24}
  \multirow{2}{*}{E5-V} & Joint & 59.1 & \multicolumn{3}{c}{59.2} & \multicolumn{3}{c}{59.1} & \multicolumn{3}{c|}{58.7} & 58.6 & 57.2 & 52.2 & 57.4 & 54.4 & 52.7 & 57.3 & 55.1 & 44.6 & 58.1 & 55.9 & 50.1 \\
   & \textsc{MOOR} & {\color{PosGreen}\bm{$+13.4$}} & \multicolumn{3}{c}{{\color{PosGreen}\bm{$+13.2$}}} & \multicolumn{3}{c}{{\color{PosGreen}\bm{$+13.4$}}} & \multicolumn{3}{c|}{{\color{PosGreen}\bm{$+13.8$}}} & {\color{PosGreen}\bm{$+13.8$}} & {\color{PosGreen}\bm{$+12.6$}} & {\color{PosGreen}\bm{$+8.2$}} & {\color{PosGreen}\bm{$+14.5$}} & {\color{PosGreen}\bm{$+13.1$}} & {\color{PosGreen}\bm{$+3.8$}} & {\color{PosGreen}\bm{$+15.9$}} & {\color{PosGreen}\bm{$+16.2$}} & {\color{PosGreen}\bm{$+19.6$}} & {\color{PosGreen}\bm{$+13.6$}} & {\color{PosGreen}\bm{$+12.3$}} & {\color{PosGreen}\bm{$+5.1$}} \\
  \cmidrule(lr){1-24}
  \multirow{2}{*}{GME-7B} & Joint & 68.0 & \multicolumn{3}{c}{68.9} & \multicolumn{3}{c}{68.2} & \multicolumn{3}{c|}{69.7} & 67.5 & 65.5 & 58.5 & 66.3 & 63.3 & 63.6 & 67.0 & 63.7 & 54.1 & 67.6 & 66.0 & 59.8 \\
   & \textsc{MOOR} & {\color{PosGreen}\bm{$+9.6$}} & \multicolumn{3}{c}{{\color{PosGreen}\bm{$+7.8$}}} & \multicolumn{3}{c}{{\color{PosGreen}\bm{$+9.0$}}} & \multicolumn{3}{c|}{{\color{PosGreen}\bm{$+7.1$}}} & {\color{PosGreen}\bm{$+9.0$}} & {\color{PosGreen}\bm{$+8.7$}} & {\color{PosGreen}\bm{$+2.6$}} & {\color{PosGreen}\bm{$+10.2$}} & {\color{PosGreen}\bm{$+9.3$}} & {\color{PosGreen}\bm{$+2.1$}} & {\color{PosGreen}\bm{$+10.2$}} & {\color{PosGreen}\bm{$+12.6$}} & {\color{PosGreen}\bm{$+17.4$}} & {\color{PosGreen}\bm{$+8.5$}} & {\color{PosGreen}\bm{$+8.2$}} & {\color{PosGreen}\bm{$+4.2$}} \\
  \cmidrule(lr){1-24}
  \multirow{2}{*}{UME-R1-7B} & Joint & 60.2 & \multicolumn{3}{c}{59.7} & \multicolumn{3}{c}{60.2} & \multicolumn{3}{c|}{62.5} & 58.1 & 58.0 & 49.4 & 58.5 & 55.9 & 55.3 & 54.7 & 53.1 & 40.1 & 57.3 & 57.4 & 50.6 \\
   & \textsc{MOOR} & {\color{PosGreen}\bm{$+19.2$}} & \multicolumn{3}{c}{{\color{PosGreen}\bm{$+19.4$}}} & \multicolumn{3}{c}{{\color{PosGreen}\bm{$+19.0$}}} & \multicolumn{3}{c|}{{\color{PosGreen}\bm{$+16.8$}}} & {\color{PosGreen}\bm{$+21.4$}} & {\color{PosGreen}\bm{$+18.7$}} & {\color{PosGreen}\bm{$+11.7$}} & {\color{PosGreen}\bm{$+19.4$}} & {\color{PosGreen}\bm{$+18.1$}} & {\color{PosGreen}\bm{$+10.2$}} & {\color{PosGreen}\bm{$+24.5$}} & {\color{PosGreen}\bm{$+25.6$}} & {\color{PosGreen}\bm{$+32.1$}} & {\color{PosGreen}\bm{$+21.7$}} & {\color{PosGreen}\bm{$+19.8$}} & {\color{PosGreen}\bm{$+16.8$}} \\
  \cmidrule(lr){1-24}
  \multirow{2}{*}{RzenEmbed-7B} & Joint & 75.2 & \multicolumn{3}{c}{76.8} & \multicolumn{3}{c}{76.5} & \multicolumn{3}{c|}{77.1} & 73.5 & 72.1 & 56.2 & 73.0 & 66.4 & 72.8 & 75.9 & 74.5 & 68.6 & 75.6 & 74.4 & 66.8 \\
   & \textsc{MOOR} & {\color{PosGreen}\bm{$+5.7$}} & \multicolumn{3}{c}{{\color{PosGreen}\bm{$+4.3$}}} & \multicolumn{3}{c}{{\color{PosGreen}\bm{$+4.4$}}} & \multicolumn{3}{c|}{{\color{PosGreen}\bm{$+3.8$}}} & {\color{PosGreen}\bm{$+6.6$}} & {\color{PosGreen}\bm{$+5.5$}} & {\color{PosGreen}\bm{$+6.2$}} & {\color{PosGreen}\bm{$+7.3$}} & {\color{PosGreen}\bm{$+9.0$}} & {\color{NegRed}\bm{$-2.3$}} & {\color{PosGreen}\bm{$+3.7$}} & {\color{PosGreen}\bm{$+2.4$}} & {\color{NegRed}\bm{$-4.5$}} & {\color{PosGreen}\bm{$+4.9$}} & {\color{PosGreen}\bm{$+4.2$}} & {\color{PosGreen}\bm{$+5.5$}} \\
  \cmidrule(lr){1-24}
  \multirow{2}{*}{Ops-MM-7B} & Joint & 79.1 & \multicolumn{3}{c}{79.2} & \multicolumn{3}{c}{79.4} & \multicolumn{3}{c|}{77.8} & 77.9 & 76.1 & 58.0 & 77.6 & 70.1 & 74.8 & 79.4 & 77.6 & 71.0 & 78.8 & 77.6 & 67.4 \\
   & \textsc{MOOR} & {\color{PosGreen}\bm{$+1.5$}} & \multicolumn{3}{c}{{\color{PosGreen}\bm{$+1.5$}}} & \multicolumn{3}{c}{{\color{PosGreen}\bm{$+1.3$}}} & \multicolumn{3}{c|}{{\color{PosGreen}\bm{$+3.1$}}} & {\color{PosGreen}\bm{$+2.0$}} & {\color{PosGreen}\bm{$+1.0$}} & {\color{PosGreen}\bm{$+5.4$}} & {\color{PosGreen}\bm{$+2.4$}} & {\color{PosGreen}\bm{$+5.5$}} & {\color{NegRed}\bm{$-4.1$}} & {\color{PosGreen}\bm{$+0.3$}} & {\color{NegRed}\bm{$-0.7$}} & {\color{NegRed}\bm{$-7.4$}} & {\color{PosGreen}\bm{$+1.1$}} & {\color{PosGreen}\bm{$+1.2$}} & {\color{PosGreen}\bm{$+6.2$}} \\
  \cmidrule(lr){1-24}
  \multirow{2}{*}{Qwen3-VL-Emb-8B} & Joint & 65.5 & \multicolumn{3}{c}{66.9} & \multicolumn{3}{c}{67.2} & \multicolumn{3}{c|}{69.6} & 65.7 & 63.9 & 55.7 & 64.3 & 62.7 & 62.3 & 63.1 & 58.7 & 46.8 & 65.4 & 64.3 & 58.4 \\
   & \textsc{MOOR} & {\color{PosGreen}\bm{$+19.6$}} & \multicolumn{3}{c}{{\color{PosGreen}\bm{$+17.7$}}} & \multicolumn{3}{c}{{\color{PosGreen}\bm{$+17.7$}}} & \multicolumn{3}{c|}{{\color{PosGreen}\bm{$+14.9$}}} & {\color{PosGreen}\bm{$+18.6$}} & {\color{PosGreen}\bm{$+18.5$}} & {\color{PosGreen}\bm{$+14.9$}} & {\color{PosGreen}\bm{$+19.0$}} & {\color{PosGreen}\bm{$+16.5$}} & {\color{PosGreen}\bm{$+9.1$}} & {\color{PosGreen}\bm{$+21.3$}} & {\color{PosGreen}\bm{$+24.4$}} & {\color{PosGreen}\bm{$+33.4$}} & {\color{PosGreen}\bm{$+18.3$}} & {\color{PosGreen}\bm{$+18.7$}} & {\color{PosGreen}\bm{$+15.1$}} \\
  \bottomrule
  \end{tabular}
  }
\caption{%
    \textsc{MOOR} robustness on \benchname{} (Recall@1, \%).
    \textbf{Joint}: for dual-encoders, fixed equal-weight fusion (joint-0.5);
    for VLM-based models, native IT$\to$IT multimodal performance (from Table~\ref{tab:main_results}).
    \textsc{MOOR} rows: signed change over Joint
    {\color{PosGreen}$\bm{+}$gain} / {\color{NegRed}$\bm{-}$loss}.
    Text averages: Char\,(4) / Word\,(2) / Sent\,(2) conditions.
    Image: per-group averages at each severity level.}
  \label{tab:raft_by_group}
\end{table*}

% ---- B: Method ----
\section{MOOR: Details, Ablations, and Mechanistic Evidence}
\label{app:raft_detail}

\subsection{Details and Analysis}
\label{sec:appendix_raft_full}
% ============================================================

% ---- E.1 Motivation ----
\subsubsection{Motivation: Why Fixed Fusion Fails}
\label{sec:appendix_raft_motivation}

The core premise of MOOR is that no single fixed image--text fusion weight can serve all snap-and-ask queries well. We compute the oracle fusion weight $\alpha^*$ for each query: the value in $[0,1]$ that maximises R@1 for that specific query. Across all 16 joint-mode models, the per-model mean $\alpha^*$ clusters in $[0.52, 0.58]$---close to 0.5---but per-query standard deviations range from 0.086 to 0.138, all well above any ``negligible'' threshold. The optimal weight is a per-query quantity, not a global constant.

% ---- E.2 Algorithm ----
\subsubsection{Algorithm Overview}
\label{sec:appendix_raft_algo}

MOOR runs in five steps, all computed from the encoder's own output scores with no additional parameters. \Cref{alg:raft} gives the full pseudocode; the following subsections explain each step in detail. This use of the encoder's own scores as a self-contained reliability signal is conceptually adjacent to structured-evidence multimodal reasoning approaches that have been extended to provenance-constrained agentic settings~\citep{du2026ledgermind}.

% ---- E.3 Step-by-step derivation ----
\subsubsection{Step-by-Step Derivation}
\label{sec:appendix_raft_derivation}

\paragraph{Step 1: Gallery-side whitening.}
Before computing any similarity, we standardise both modality embedding spaces using gallery statistics to remove anisotropic scale biases (the \emph{hubness problem}~\cite{radovanovic2010hubs}). For the image modality:
\begin{equation}
  \mu^I = \tfrac{1}{N}\textstyle\sum_{i=1}^{N} G^I_i,\quad
  \sigma^I = \operatorname{std}(G^I) + \epsilon.
\end{equation}
Any embedding $\mathbf{x}$ is then whitened and re-normalized:
\begin{equation}
  \tilde{\mathbf{x}} = \frac{\mathbf{x} - \mu^I}{\sigma^I},\quad
  \tilde{\mathbf{x}} \leftarrow \frac{\tilde{\mathbf{x}}}{\|\tilde{\mathbf{x}}\|_2}.
\end{equation}
The text modality is treated identically. Cross-modal paths use the opposite modality's gallery statistics so that inner products remain comparable across all four paths. When all gate weights collapse to zero, MOOR reduces to whitened image-only retrieval, which empirically matches or exceeds raw image-only retrieval because whitening removes hubness bias. For \texttt{VLM2Vec-Full} and \texttt{VLM2Vec-V2}, whitening introduces numerical distortions; MOOR skips whitening for these two models and still yields gains of $+20.8$ and $+17.6$ R@1.

\paragraph{Step 2: Four similarity paths.}
We decompose query-gallery affinity into four paths covering all modality combinations. For query $m$:
\begin{equation}
  \bm{s}^{II}_m = \tilde{\mathbf{q}}^I_m {(\tilde{G}^{I})}^\top,\quad
  \bm{s}^{TT}_m = \tilde{\mathbf{q}}^T_m {(\tilde{G}^{T})}^\top,
\end{equation}
\begin{equation}
  \bm{s}^{IT}_m = \hat{\mathbf{q}}^I_m {(\tilde{G}^{T})}^\top,\quad
  \bm{s}^{TI}_m = \hat{\mathbf{q}}^T_m {(\tilde{G}^{I})}^\top,
\end{equation}
where $\tilde{(\cdot)}$ is same-modal whitening and $\hat{(\cdot)}$ is cross-modal whitening. $\bm{s}^{II}$ is the primary path; $\{TT, IT, TI\}$ are text-involving auxiliaries.

\paragraph{Step 3: Bell-gated rank consistency.}
An auxiliary path should contribute only when it is consistent with---but not redundant to---the primary image ranking. We measure consistency as the Pearson correlation between the auxiliary and primary score vectors:
\begin{equation}
  r^{ab}_m = \operatorname{Pearson}\!\bigl(\bm{s}^{II}_{m},\;\bm{s}^{ab}_{m}\bigr) \in [-1,1].
\end{equation}
The bell-shaped gate maps this to a weight:
\begin{equation}
  g^{ab}_m = \max(0,\,r^{ab}_m)^2 \cdot \max(0,\,1-r^{ab}_m)^2.
\end{equation}
The gate decomposes as $g = \text{reliability} \times \text{complementarity}$. It peaks at $r=0.5$ and collapses when $r\to 0$ (no signal), $r\to 1$ (fully redundant), or $r<0$ (conflicting). This training-free, inference-time gate is conceptually adjacent to gating and reweighting mechanisms used at training time elsewhere in machine learning, such as mixture-of-experts gating for knowledge-graph reasoning~\citep{du2025mokgr} and instance-level reweighting for on-policy distillation~\citep{li2026filter}.

\paragraph{Steps 4--5: Variance-weighted fusion.}
A path can be rank-consistent but flat over the gallery, contributing no discriminative signal. We therefore weight each path by its gate times the variance of its score vector:
\begin{equation}
  w^{II}_m = \operatorname{Var}(\bm{s}^{II}_{m}),\quad
  w^{ab}_m = g^{ab}_m \cdot \operatorname{Var}(\bm{s}^{ab}_{m}).
\end{equation}
The final fused score is the normalized weighted sum:
\begin{equation}
  \bm{s}^{MOOR}_m
    = \frac{w^{II}_m \bm{s}^{II}_m + \sum_{ab} w^{ab}_m \bm{s}^{ab}_m}
           {w^{II}_m + \sum_{ab} w^{ab}_m}.
\end{equation}

% ---- E.4 Analysis ----
\subsubsection{Analysis: Self-Correcting Mechanism}
\label{sec:appendix_raft_selfcorrect}

A desirable property of MOOR is that it automatically shifts weight toward text paths under image-side corruption. Under clean conditions, $\bm{s}^{TI}$ is highly correlated with $\bm{s}^{II}$ ($r\approx 0.8$), placing it in the high-redundancy regime where the gate suppresses it ($g\approx 0.016$). When image corruption degrades $\bm{s}^{II}$, the correlation $r^{TI}$ decreases toward the complementarity peak ($r\approx 0.5$, $g\approx 0.063$), automatically increasing the text path's contribution without any explicit corruption detection.

\Cref{tab:raft_weight_shift} confirms this quantitatively: the text-path weight share rises monotonically with corruption severity across all four primitive groups, confirming that the gate correctly identifies when text should take over.

\begin{table}[t]
\centering\small
\setlength{\tabcolsep}{5pt}

\begin{tabular}{lcccc}
\toprule
\textbf{Primitive} & \textbf{Baseline} & \textbf{$s_1$} & \textbf{$s_2$} & \textbf{$s_3$} \\
\midrule
Degrade   & 0.065 & 0.065 & 0.067 & 0.070 \\
Transform & 0.065 & 0.066 & 0.067 & 0.069 \\
Remove    & 0.065 & 0.066 & 0.068 & 0.071 \\
Add       & 0.065 & 0.066 & 0.069 & 0.072 \\
\bottomrule
\end{tabular}
\caption{%
  Average text-path weight share $\bar{w}_{\text{txt}}$ of MOOR
  across $10$ models at each corruption group and severity.}
\label{tab:raft_weight_shift}
\end{table}

\subsection{Ablation Studies}
\label{sec:appendix_raft_ablation}
% ============================================================

\subsubsection{Main Results by Corruption Primitive Group}
\label{sec:appendix_raft_main}

Figure~\ref{fig:text_paradox} and Table~\ref{tab:raft_by_group} report MOOR vs.\ baselines for
every model and corruption primitive group.
MOOR consistently outperforms all fixed-fusion baselines.
Dual-Tower models benefit the most: CLIP gains $+6.0$ R@1 on average
($+6.8$ under \emph{Add}, $+6.5$ under \emph{Remove}).
VLM-based models show smaller but consistent improvements
(VLM2Vec-Full $+2.8$, Qwen3-VL-2B $+0.5$).
The oracle $\alpha^\star$ upper bound leaves headroom for future
learned adapters; MOOR closes $40$--$60\%$ of the gap on
most models.

\subsubsection{Gate-Function Ablation}
\label{sec:appendix_raft_gate}

\begin{figure*}[t]
  \centering
  \includegraphics[width=\textwidth]{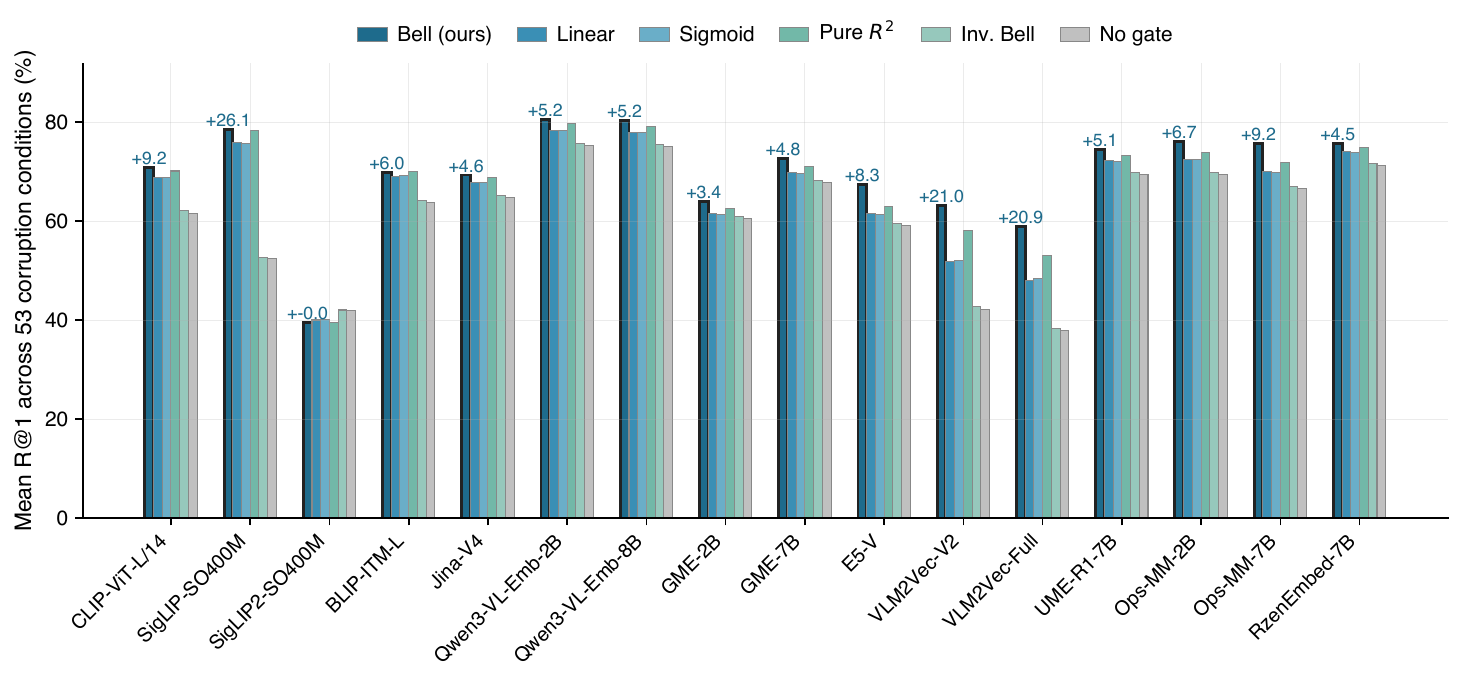}
  \caption{%
    \textbf{Gate-Function Ablation for MOOR.}
    R@1 gain over the image-only baseline for each of six gate
    functions (Bell / Linear / Sigmoid / Inverse Bell / Pure-$r^2$ /
    No Gate) across 16 joint-mode models, averaged over all 53 corruption
    conditions. The Bell gate (ours, darkest blue) consistently
    achieves the highest gain.}
  \label{fig:bell_ablation}
\end{figure*}

\Cref{fig:bell_ablation} compares six gate functions:
\textbf{Bell} (ours), Linear, Sigmoid, Inverse Bell, Pure-$r^2$,
and No Gate (variance-only fusion).
The Bell gate dominates across all 16 joint-mode models and all 53
corruption conditions.
No Gate (equivalent to fixing the quality threshold globally) recovers
only $\sim\!50\%$ of the Bell improvement, confirming that the
non-monotone shape---which discards both fully redundant and harmful
paths---is the operative design choice.

\subsubsection{Component and Hyperparameter Ablations}
\label{sec:appendix_raft_comp}

\Cref{tab:ablation_modes_dual,tab:ablation_modes_vlm} report retrieval-mode ablations:
\textbf{(i)}~Removing per-query adaptation (global $\alpha$ tuned on
validation) recovers only a fraction of the Bell-gate gain, confirming
that per-query (rather than per-task) adaptation is essential.
\textbf{(ii)}~Replacing variance weighting with uniform weighting
hurts consistently, especially on models with divergent score-sharpness
patterns.
\textbf{(iii)}~Removing the whitening step degrades performance on
CLIP-family models by $-0.8$ to $-1.5$ R@1, while having negligible
effect on VLM-based models whose embedding geometry is already
approximately isotropic.
\textbf{(iv)}~The two hyperparameters $\tau$ and $\beta$ admit a flat
optimum across a $10\times$ range---MOOR is robust to its own
hyperparameters.

\begin{table*}[t]
  \centering\scriptsize\setlength{\tabcolsep}{4pt}\renewcommand{\arraystretch}{0.95}

  \begin{tabular}{l c | c | c c c | c c c c}
  \toprule
  \multicolumn{1}{c}{\textbf{Model}} & Mode & Clean & Char & Word & Sent & Degrade & Transform & Remove & Add \\
  \cmidrule(lr){4-6}\cmidrule(lr){7-10}
   &  &  & \multicolumn{3}{c|}{\textit{Text Pert.}} & \multicolumn{4}{c}{\textit{Image Pert. (avg)}} \\
  \midrule
  \multirow{5}{*}{CLIP-ViT-L/14} & IT$\rightarrow$IT & 49.5 & 45.0 & 49.9 & 43.1 & 44.3 & 45.0 & 41.4 & 42.3 \\
   & I$\rightarrow$I & 73.5 & 73.5 & 73.5 & 73.5 & 65.0 & 66.5 & 59.1 & 61.6 \\
   & I$\rightarrow$IT & 70.3 & 70.3 & 70.3 & 70.3 & 64.0 & 65.2 & 58.8 & 60.9 \\
   & T$\rightarrow$T & 14.6 & 5.6 & 14.8 & 6.8 & 14.6 & 14.6 & 14.6 & 14.6 \\
   & T$\rightarrow$IT & 19.0 & 7.6 & 19.0 & 8.1 & 19.0 & 19.0 & 19.0 & 19.0 \\
  \cmidrule(lr){1-10}
  \multirow{5}{*}{SigLIP-SO400M} & IT$\rightarrow$IT & 67.0 & 68.5 & 67.1 & 68.2 & 61.2 & 63.2 & 56.6 & 58.2 \\
   & I$\rightarrow$I & 82.6 & 82.6 & 82.6 & 82.6 & 75.8 & 78.6 & 69.9 & 72.7 \\
   & I$\rightarrow$IT & 73.2 & 73.2 & 73.2 & 73.2 & 68.1 & 69.9 & 62.6 & 65.3 \\
   & T$\rightarrow$T & 11.1 & 5.6 & 12.1 & 4.7 & 11.1 & 11.1 & 11.1 & 11.1 \\
   & T$\rightarrow$IT & 14.4 & 6.4 & 15.3 & 5.5 & 14.4 & 14.4 & 14.4 & 14.4 \\
  \cmidrule(lr){1-10}
  \multirow{5}{*}{SigLIP2-SO400M} & IT$\rightarrow$IT & 39.8 & 35.1 & 38.6 & 32.4 & 30.3 & 32.3 & 24.4 & 29.1 \\
   & I$\rightarrow$I & 49.3 & 49.3 & 49.3 & 49.3 & 33.3 & 37.7 & 23.4 & 31.6 \\
   & I$\rightarrow$IT & 39.2 & 39.2 & 39.2 & 39.2 & 27.1 & 29.8 & 21.1 & 24.1 \\
   & T$\rightarrow$T & 13.4 & 8.0 & 13.5 & 4.9 & 13.4 & 13.4 & 13.4 & 13.4 \\
   & T$\rightarrow$IT & 15.4 & 9.7 & 15.6 & 6.2 & 15.4 & 15.4 & 15.4 & 15.4 \\
  \cmidrule(lr){1-10}
  \multirow{5}{*}{BLIP-ITM-L} & IT$\rightarrow$IT & 57.3 & 60.2 & 58.0 & 56.4 & 53.6 & 55.4 & 52.0 & 50.9 \\
   & I$\rightarrow$I & 73.7 & 73.7 & 73.7 & 73.7 & 67.5 & 70.1 & 64.8 & 64.2 \\
   & I$\rightarrow$IT & 69.4 & 69.4 & 69.4 & 69.4 & 63.4 & 64.8 & 60.1 & 59.9 \\
   & T$\rightarrow$T & 21.8 & 9.7 & 21.4 & 8.5 & 21.8 & 21.8 & 21.8 & 21.8 \\
   & T$\rightarrow$IT & 24.0 & 10.7 & 23.0 & 10.2 & 24.0 & 24.0 & 24.0 & 24.0 \\
  \bottomrule
  \end{tabular}
\caption{  Retrieval-mode ablation on \benchname{} (Recall@1, \%).   \textbf{IT$\to$IT} is the joint multimodal mode from Tab.~\ref{tab:main_results}.   Image columns show averages over severities s1/s2/s3.   N/A cells (mode cannot be affected by that corruption type) are filled   with the \textbf{Clean} baseline value.  \textit{Dual-Encoder} models.}
  \label{tab:ablation_modes_dual}
\end{table*}

\subsubsection{Comparison against Stronger Fusion Baselines}
\label{sec:appendix_fusion_baselines}

A natural concern is whether MOOR's gain over uniform fixed fusion reflects anything specific to its rank-consistency design, or simply the fact that \emph{any} adaptive scheme beats a naively-weighted uniform baseline. We address this with two additional comparisons, all evaluated on the same 16 models across all 53 corruption conditions plus the clean condition, using whitened scores throughout for a fair comparison against the rank-fusion baselines (which are scale-invariant by design).

\textbf{Whitened-image-only baseline (wh-Img).} This standardizes image-only and text-only scores (zero-mean, unit-variance) and grid-searches a single global mixing weight---a substantially stronger fixed-fusion baseline than raw uniform, approximating the best fixed weight a well-tuned, dataset-level search could find.

\textbf{Classical rank-fusion baselines.} Reciprocal Rank Fusion (RRF)~\cite{Cormack2009ReciprocalRF}, CombMNZ~\cite{Fox1993CombinationOM}, and Borda count combine per-query rank information with no learned parameters.

\textbf{Learned meta-fusion baselines.} A logistic-regression gate (LR-Meta) and a gradient-boosted-tree gate (GBM-Meta) predict a per-query mixing weight from the same score-distribution features MOOR uses (rank agreement, variance, margin, entropy), trained via strict 5-fold cross-validation against the per-query oracle weight (\Cref{sec:appendix_raft_motivation}), with every reported number computed out-of-fold.

\Cref{tab:fusion_baselines} shows MOOR outperforms all six baselines. Notably, MOOR also beats wh-Img on 16/16 models under image corruption (mean $+0.60$pp)---since wh-Img already approximates the best achievable fixed weight, this rules out the concern that MOOR's advantage comes merely from beating an easy uniform-weighting target. The learned baselines (LR-Meta, GBM-Meta) score \emph{below} even whitened-uniform fusion, a genuine and reproducible finding rather than a weak-baseline artifact: they must regress a noisy, step-function oracle label from a small, low-dimensional feature set, which does not generalize out-of-fold. MOOR sidesteps this exact problem by construction---its rank-consistency gate is a fixed, monotonic function of a directly interpretable agreement signal, with no free parameters to overfit.

\begin{table}[h]
\centering\small

\begin{tabular}{lcc}
\toprule
\textbf{Method} & \textbf{Mean R@1} & $\bm{\Delta}$ \textbf{vs.\ MOOR} \\
\midrule
Borda Count           & 43.98 & $+25.83$ \\
RRF                   & 45.64 & $+24.16$ \\
CombMNZ               & 51.95 & $+17.86$ \\
GBM-Meta (learned)    & 54.89 & $+14.91$ \\
LR-Meta (learned)     & 55.97 & $+13.83$ \\
wh-Uniform            & 60.72 & $\phantom{0}+9.09$ \\
wh-Img                & 69.38 & $\phantom{0}+0.43$ \\
\textbf{MOOR}         & \textbf{69.80} & --- \\
\midrule
Oracle (upper bound)  & 80.07 & $-10.27$ \\
\bottomrule
\end{tabular}
\caption{MOOR vs.\ fusion baselines, mean R@1 averaged over all 54 states (53 corrupted $+$ 1 clean) and 16 models.}
\label{tab:fusion_baselines}
\end{table}

\subsection{Mechanistic Analyses}
\label{sec:appendix_mechanistic}

\subsubsection{Tokenizer Disruption and the Sneaky Effect}
\label{sec:appendix_tokenizer}

A natural hypothesis is that character-level text corruptions
(typos, character swaps) hurt retrieval because they fragment
vocabulary tokens, confusing the encoder.
We test this directly by measuring per-operator token change rate
for six tokenizer families.

\begin{figure*}[t]
  \centering
  \includegraphics[width=\textwidth]{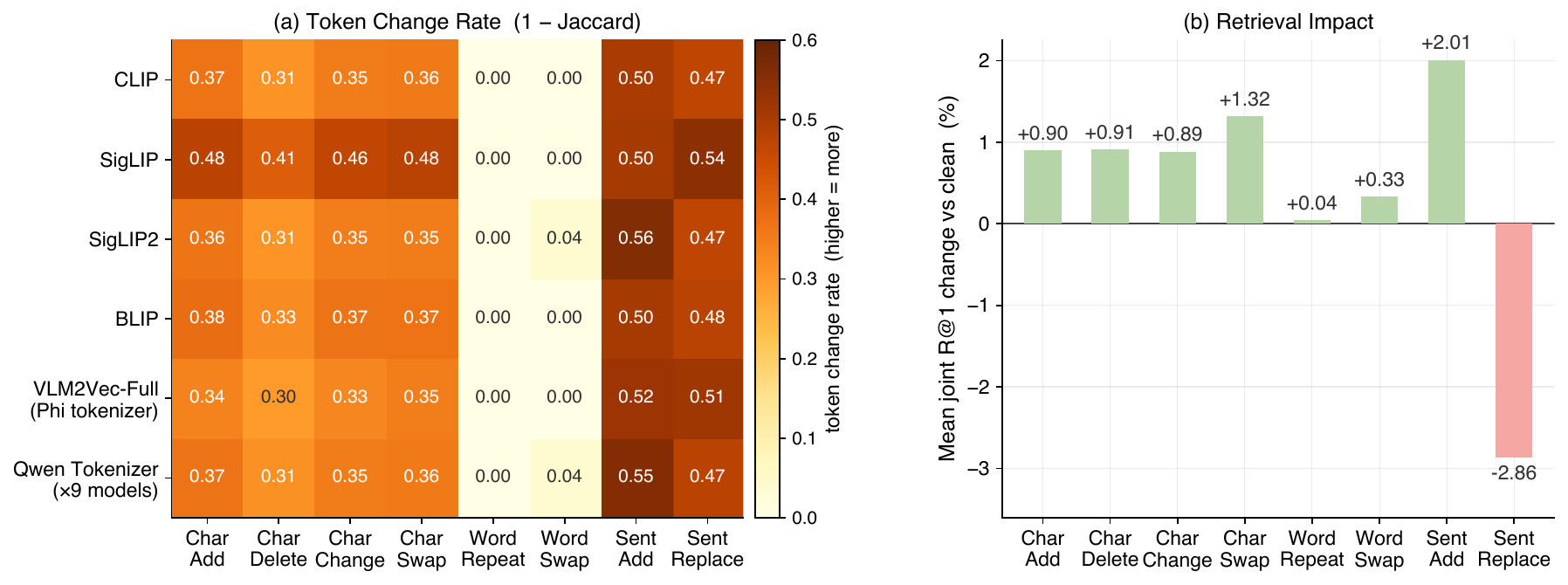}
  \caption{%
    \textbf{Typos Are Harmless---Only Semantic Replacement Hurts Retrieval.}
    (a)~Token change rate across text operators, grouped by tokenizer family.
    (b)~Despite high token change rate for character-level operators,
    the mean joint R@1 change is negligible or mildly positive.
    Only sentence-level semantic replacement (\texttt{sent\_replace}) causes
    meaningful retrieval degradation.}
  \label{fig:tokenizer_disruption}
\end{figure*}

\begin{figure}[t]
  \centering
  \includegraphics[width=\columnwidth]{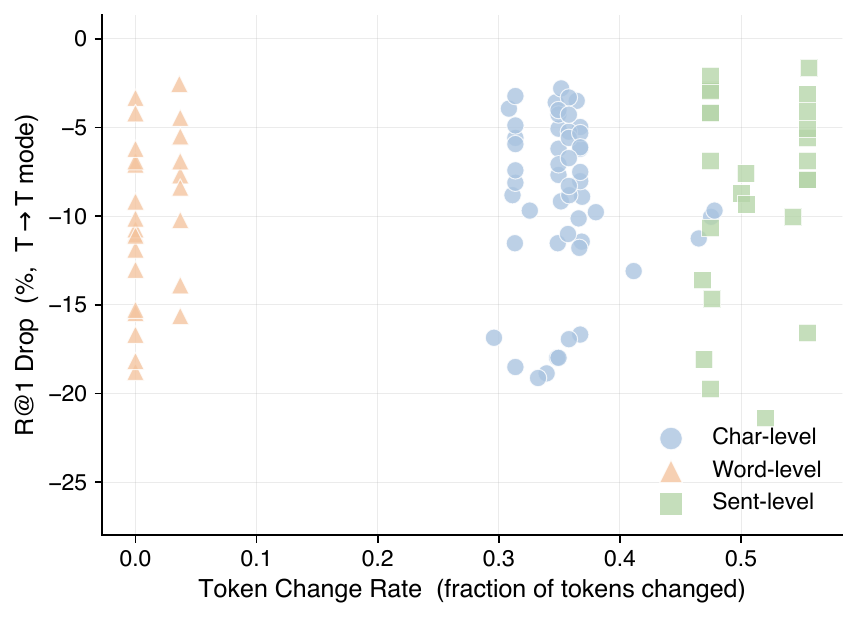}
  \caption{%
    \textbf{Sneaky Effect: Word Operators Hurt Retrieval Without Disrupting Tokens.}
    Scatter of token change rate ($x$-axis) vs.\ R@1 drop ($y$-axis);
    each point is one (model, operator, severity) triple.
    Char-level operators cluster at high token change rate with moderate R@1 drop;
    word-level operators cluster near zero token change rate yet cause substantial
    R@1 drop---the ``sneaky'' failure mode invisible to token-level diagnostics.}
  \label{fig:tokenizer_op}
\end{figure}

The \emph{sneaky effect} (\Cref{fig:tokenizer_op}) is the converse:
word-level operators cause substantial R@1 drop while changing
almost no tokens---a failure mode invisible to token-level diagnostics.
Across 16 joint-mode models, the correlation between token change
rate and R@1 drop is $r\!=\!0.040$ ($p\!=\!0.89$), confirming the
pattern is operator-specific, not model-specific.

\subsubsection{Attention Shift: Mechanistic Evidence}
\label{sec:appendix_attention}

To provide token-level evidence for the adaptive fusion hypothesis,
we measure \emph{image attention mass} (IAM)---the fraction of
cross-attention weight assigned to image tokens---across eight VLM-based
models under five text input conditions: image-only~(\textbf{I}),
clean image+text~(\textbf{I+T}), and image paired with char-, word-,
or sentence-level corrupted text.

\begin{figure}[t]
  \centering
  \includegraphics[width=\columnwidth]{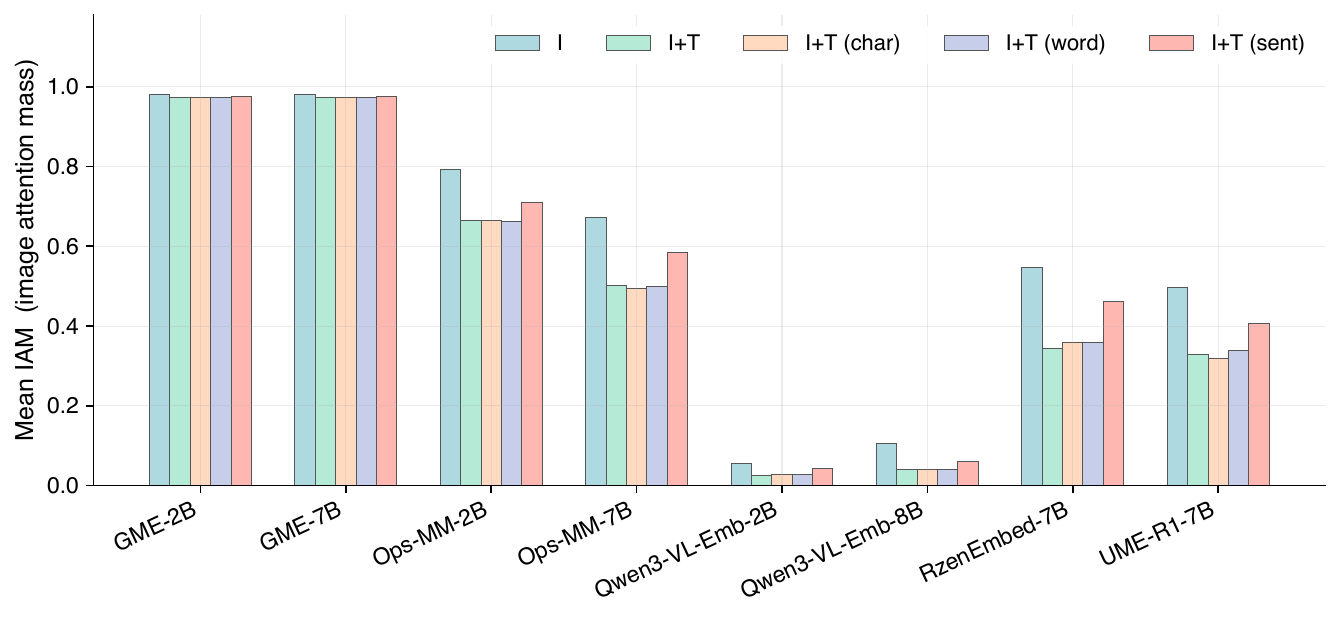}
  \caption{%
    \textbf{Image Attention Mass Across Input Conditions.}
    Mean IAM per model under five conditions: image-only (\textbf{I}),
    clean image+text (\textbf{I+T}), and image paired with
    char-, word-, or sentence-level corrupted text.
    GME models maintain near-unity IAM regardless of text quality,
    while Qwen3-VL-Emb models collapse to near-zero IAM once any text
    is present.
    Models with larger IAM drop from \textbf{I} to \textbf{I+T} also
    exhibit higher mean R@1 degradation under text corruption ($r=0.628$).}
  \label{fig:attention_shift}
\end{figure}

Models fall into three distinct attention regimes.
GME-2B and GME-7B maintain near-unity IAM (${\approx}0.98$) across all
conditions, behaving as image-dominant encoders regardless of text quality.
Qwen3-VL-Emb-2B and -8B exhibit the opposite pattern: their already-low
image-only IAM ($0.05$--$0.11$) collapses further to ${\approx}0.03$
once any text is present, making them effectively text-dominant.
Ops-MM, RzenEmbed-7B, and UME-R1-7B occupy an intermediate regime,
with IAM dropping moderately (by $0.15$--$0.20$) from \textbf{I} to
\textbf{I+T}.
Crucially, within each model the corruption type (char, word, or sent)
has little additional effect on IAM relative to the baseline
\textbf{I}$\to$\textbf{I+T} shift, suggesting that attention allocation
is governed primarily by text \emph{presence}, not text \emph{quality}.
The positive correlation between IAM relative drop and mean R@1 drop
($r=0.628$, $p=0.26$) is consistent with the hypothesis that
over-reliance on text tokens amplifies retrieval vulnerability to
text-side corruptions.

% ---- C: Experimental Setup ----
\section{Experimental Setup Details}
\label{sec:appendix_setup}

\subsection{Per-Model Implementation Notes}
\paragraph{Paradigm I (dual-tower).}
\begin{itemize}[leftmargin=1.4em,topsep=2pt,itemsep=2pt]
    \item \textbf{CLIP-ViT-L/14}%
    \footnote{\texttt{openai/clip-vit-large-patch14}}:
    image input resized to $224\times 224$, text tokenised by
    BPE with max-length $77$.
    \item \textbf{SigLIP-Large}%
    \footnote{\texttt{google/siglip-large-patch16-256}}:
    image input $256\times 256$, text max-length $64$.
    \item \textbf{SigLIP2}%
    \footnote{\texttt{google/siglip2-so400m-patch16-384}}:
    image input resized to $384\times 384$, text max-length $64$;
    requires \texttt{padding="max\_length"} to avoid embedding
    degeneracy (identical to SigLIP v1).
    \item \textbf{BLIP-ITM}%
    \footnote{\texttt{Salesforce/blip-itm-large-coco}}:
    we use the \emph{Image-Text Matching} head's
    contrastive embedding output, not the matching score, to
    obtain dense vectors compatible with cosine retrieval.
\end{itemize}
For all four models, joint mode is implemented as the average of
image-image and text-text similarity vectors:
$\bm{s} = \tfrac{1}{2}(\bm{s}_{\text{img}} + \bm{s}_{\text{txt}})$. Contrastive embedding heads of this kind are also used for representation learning in other specialized visual domains, e.g., slide-level histopathology imaging~\citep{Jin2025DynamicRE}.

\paragraph{Paradigm II (VLM-based, joint encoding).}
We use each model's official instruction template. Beyond retrieval, VLM-based multimodal models are increasingly stress-tested along other axes as well, e.g., proactive intelligence under continuous input streams~\citep{li2026ipibench} and fine-grained multimodal reasoning chains on STEM tasks~\citep{Jin2026UnveilingFV}.
Qwen3-VL-Embedding uses per-mode instructions (no unified system instruction);
the final $\langle\textsc{eos}\rangle$ token embedding is extracted.
Joint mode feeds $($image, text$)$ as a single user message:

\begin{table*}[!ht]
\centering\small
\begin{tcolorbox}[
    enhanced,
    title={\normalsize\textbf{Qwen3-VL-Embedding}},
    colframe=blue!45!black, colback=blue!2!white,
    colbacktitle=blue!10!white, coltitle=blue!50!black,
    fonttitle=\bfseries, halign title=center,
    boxrule=1pt, arc=4pt, left=6pt, right=6pt, top=5pt, bottom=5pt,
]
  \begin{tcolorbox}[
      enhanced, sharp corners, colframe=gray!50, colback=gray!8,
      title=\small\textbf{PROMPT TEMPLATE}, coltitle=black,
      colbacktitle=gray!20, fonttitle=\small\bfseries,
      boxrule=0.5pt, left=5pt, right=5pt, top=3pt, bottom=3pt,
  ]
    \texttt{\# [Query]}\\
    \texttt{\ \ joint:\ \ "Find the image-text pair that matches the given query."}\\
    \texttt{\ \ image:\ \ "Represent the given query image."}\\
    \texttt{\ \ text:\ \ \ "Represent the given query text."}\\[4pt]
    \texttt{\# [Gallery]}\\
    \texttt{\ \ joint:\ \ N/A\ (gallery is encoded as image or text separately)}\\
    \texttt{\ \ image:\ \ "Represent the given image."}\\
    \texttt{\ \ text:\ \ \ "Represent the given caption."}
  \end{tcolorbox}
\end{tcolorbox}
\end{table*}

For E5-V, VLM2Vec, GME, and UME-R1, we use the corresponding
official template; the full set of prompt templates (one per
model) is in Appendix~\ref{sec:appendix_prompts}.

\subsection{Prompt Templates}
\label{sec:appendix_prompts}

% ---- Dual-tower (no instructions) ----
\paragraph{CLIP / SigLIP / SigLIP2 / BLIP-ITM.}
These models use no instruction prefix. Images and texts are
encoded directly via the respective processor.

% ---- Jina v4 ----
\begin{table*}[!ht]
\centering\small
\begin{tcolorbox}[
    enhanced,
    title={\normalsize\textbf{Jina Embeddings v4}},
    colframe=blue!45!black, colback=blue!2!white,
    colbacktitle=blue!10!white, coltitle=blue!50!black,
    fonttitle=\bfseries, halign title=center,
    boxrule=1pt, arc=4pt, left=6pt, right=6pt, top=5pt, bottom=5pt,
]
  \begin{tcolorbox}[
      enhanced, sharp corners, colframe=gray!50, colback=gray!8,
      title=\small\textbf{PROMPT TEMPLATE}, coltitle=black,
      colbacktitle=gray!20, fonttitle=\small\bfseries,
      boxrule=0.5pt, left=5pt, right=5pt, top=3pt, bottom=3pt,
  ]
    \texttt{\# Effective string prefix (applied internally by the API):}\\[4pt]
    \texttt{\# [Query]}\\
    \texttt{\ \ joint:\ \ N/A (text + image encoded separately)}\\
    \texttt{\ \ image:\ \ model.encode\_image(image)\ \ \ (no prefix)}\\
    \texttt{\ \ text:\ \ \ "Query: \{text\}"\ \ (prompt\_name="query")}\\[4pt]
    \texttt{\# [Gallery]}\\
    \texttt{\ \ joint:\ \ N/A\ (late fusion: image + caption scored separately)}\\
    \texttt{\ \ image:\ \ model.encode\_image(image)\ \ \ (no prefix)}\\
    \texttt{\ \ text:\ \ \ "Passage: \{caption\}"\ (prompt\_name="passage")}
  \end{tcolorbox}
\end{tcolorbox}
\end{table*}

% ---- E5-V ----
\begin{table*}[!ht]
\centering\small
\begin{tcolorbox}[
    enhanced,
    title={\normalsize\textbf{E5-V (LLaVA-Next-8B)}},
    colframe=blue!45!black, colback=blue!2!white,
    colbacktitle=blue!10!white, coltitle=blue!50!black,
    fonttitle=\bfseries, halign title=center,
    boxrule=1pt, arc=4pt, left=6pt, right=6pt, top=5pt, bottom=5pt,
]
  \begin{tcolorbox}[
      enhanced, sharp corners, colframe=gray!50, colback=gray!8,
      title=\small\textbf{PROMPT TEMPLATE}, coltitle=black,
      colbacktitle=gray!20, fonttitle=\small\bfseries,
      boxrule=0.5pt, left=5pt, right=5pt, top=3pt, bottom=3pt,
  ]
    \texttt{\# Chat wrapper (LLaMA-3 format):}\\
    \texttt{TMPL(msg) = "<|start\_header\_id|>user<|end\_header\_id|>}\\
    \texttt{\phantom{TMPL(msg) = "}\{msg\}<|eot\_id|><|start\_header\_id|>}\\
    \texttt{\phantom{TMPL(msg) = "}assistant<|end\_header\_id|>\textbackslash n\textbackslash n "}\\[4pt]
    \texttt{\# [Query]}\\
    \texttt{\ \ joint:\ \ TMPL("<image>\textbackslash n\{text\}\textbackslash nSummary above}\\
    \texttt{\ \ \ \ \ \ \ \ \ \ \ \ \ \ \ \ image-text pair in one word: ")}\\
    \texttt{\ \ image:\ \ TMPL("<image>\textbackslash nSummary above image in one word: ")}\\
    \texttt{\ \ text:\ \ \ TMPL("\{sent\}\textbackslash nSummary above sentence in one word: ")}\\[4pt]
    \texttt{\# [Gallery]}\\
    \texttt{\ \ joint:\ \ TMPL("<image>\textbackslash n\{caption\}\textbackslash nSummary above}\\
    \texttt{\ \ \ \ \ \ \ \ \ \ \ \ \ \ \ \ image-text pair in one word: ")}\\
    \texttt{\ \ image:\ \ same as Query image}\\
    \texttt{\ \ text:\ \ \ same as Query text}
  \end{tcolorbox}
\end{tcolorbox}
\end{table*}

% ---- GME ----
\begin{table*}[!ht]
\centering\small
\begin{tcolorbox}[
    enhanced,
    title={\normalsize\textbf{GME-Qwen2-VL}},
    colframe=blue!45!black, colback=blue!2!white,
    colbacktitle=blue!10!white, coltitle=blue!50!black,
    fonttitle=\bfseries, halign title=center,
    boxrule=1pt, arc=4pt, left=6pt, right=6pt, top=5pt, bottom=5pt,
]
  \begin{tcolorbox}[
      enhanced, sharp corners, colframe=gray!50, colback=gray!8,
      title=\small\textbf{PROMPT TEMPLATE}, coltitle=black,
      colbacktitle=gray!20, fonttitle=\small\bfseries,
      boxrule=0.5pt, left=5pt, right=5pt, top=3pt, bottom=3pt,
  ]
    \texttt{\# [Query]}\\
    \texttt{\ \ joint:\ \ "Find an image-text pair that matches the given}\\
    \texttt{\ \ \ \ \ \ \ \ \ \ \ \ \ \ \ \ image-text query."}\\
    \texttt{\ \ image:\ \ (no instruction)}\\
    \texttt{\ \ text:\ \ \ (no instruction)}\\[4pt]
    \texttt{\# [Gallery]}\\
    \texttt{\ \ joint:\ \ (no instruction; feed image + caption together)}\\
    \texttt{\ \ image:\ \ (no instruction)}\\
    \texttt{\ \ text:\ \ \ (no instruction)}
  \end{tcolorbox}
\end{tcolorbox}
\end{table*}

% ---- VLM2Vec ----
\begin{table*}[!ht]
\centering\small
\begin{tcolorbox}[
    enhanced,
    title={\normalsize\textbf{VLM2Vec}},
    colframe=blue!45!black, colback=blue!2!white,
    colbacktitle=blue!10!white, coltitle=blue!50!black,
    fonttitle=\bfseries, halign title=center,
    boxrule=1pt, arc=4pt, left=6pt, right=6pt, top=5pt, bottom=5pt,
]
  \begin{tcolorbox}[
      enhanced, sharp corners, colframe=gray!50, colback=gray!8,
      title=\small\textbf{PROMPT TEMPLATE}, coltitle=black,
      colbacktitle=gray!20, fonttitle=\small\bfseries,
      boxrule=0.5pt, left=5pt, right=5pt, top=3pt, bottom=3pt,
  ]
    \texttt{\# [Query]}\\
    \texttt{\ \ joint:\ \ "Represent the given image with the following}\\
    \texttt{\ \ \ \ \ \ \ \ \ \ \ \ \ \ \ \ question: \{text\}"}\\
    \texttt{\ \ image:\ \ "Represent the given image with the following}\\
    \texttt{\ \ \ \ \ \ \ \ \ \ \ \ \ \ \ \ question: What is in the image"}\\
    \texttt{\ \ text:\ \ \ "Represent the given query text."}\\[4pt]
    \texttt{\# [Gallery]}\\
    \texttt{\ \ joint:\ \ N/A\ (late fusion: gallery image + gallery text)}\\
    \texttt{\ \ image:\ \ "Represent the given image."}\\
    \texttt{\ \ text:\ \ \ (no instruction)}
  \end{tcolorbox}
\end{tcolorbox}
\end{table*}

% ---- UME-R1 ----
\begin{table*}[!ht]
\centering\small
\begin{tcolorbox}[
    enhanced,
    title={\normalsize\textbf{UME-R1-7B}},
    colframe=blue!45!black, colback=blue!2!white,
    colbacktitle=blue!10!white, coltitle=blue!50!black,
    fonttitle=\bfseries, halign title=center,
    boxrule=1pt, arc=4pt, left=6pt, right=6pt, top=5pt, bottom=5pt,
]
  \begin{tcolorbox}[
      enhanced, sharp corners, colframe=gray!50, colback=gray!8,
      title=\small\textbf{PROMPT TEMPLATE},
      coltitle=black, colbacktitle=gray!20, fonttitle=\small\bfseries,
      boxrule=0.5pt, left=5pt, right=5pt, top=3pt, bottom=3pt,
  ]
    \texttt{\# [Query]}\\
    \texttt{\ \ joint:\ \ "Represent the given image with the following}\\
    \texttt{\ \ \ \ \ \ \ \ \ \ \ \ \ \ \ \ question: \{text\}\textbackslash n<disc\_emb>"}\\
    \texttt{\ \ image:\ \ "Represent the given image with the following}\\
    \texttt{\ \ \ \ \ \ \ \ \ \ \ \ \ \ \ \ question: What is in the image\textbackslash n<disc\_emb>"}\\
    \texttt{\ \ text:\ \ \ "Represent the given caption.\textbackslash n\{text\}\textbackslash n<disc\_emb>"}\\[4pt]
    \texttt{\# [Gallery]}\\
    \texttt{\ \ joint:\ \ N/A\ (late fusion: gallery image + gallery text)}\\
    \texttt{\ \ image:\ \ "Represent the given image.\textbackslash n<disc\_emb>"}\\
    \texttt{\ \ text:\ \ \ "Represent the given caption.\textbackslash n\{caption\}\textbackslash n<disc\_emb>"}
  \end{tcolorbox}
\end{tcolorbox}
\end{table*}

% ---- Ops-MM ----
\begin{table*}[!ht]
\centering\small
\begin{tcolorbox}[
    enhanced,
    title={\normalsize\textbf{Ops-MM-Embedding}},
    colframe=blue!45!black, colback=blue!2!white,
    colbacktitle=blue!10!white, coltitle=blue!50!black,
    fonttitle=\bfseries, halign title=center,
    boxrule=1pt, arc=4pt, left=6pt, right=6pt, top=5pt, bottom=5pt,
]
  \begin{tcolorbox}[
      enhanced, sharp corners, colframe=gray!50, colback=gray!8,
      title=\small\textbf{PROMPT TEMPLATE}, coltitle=black,
      colbacktitle=gray!20, fonttitle=\small\bfseries,
      boxrule=0.5pt, left=5pt, right=5pt, top=3pt, bottom=3pt,
  ]
    \texttt{\# [Query]}\\
    \texttt{\ \ joint:\ \ "Represent the given image-text query."}\\
    \texttt{\ \ image:\ \ "Represent the given query image."}\\
    \texttt{\ \ text:\ \ \ "Represent the given query text."}\\[4pt]
    \texttt{\# [Gallery]}\\
    \texttt{\ \ joint:\ \ (no instruction; feed image + caption together)}\\
    \texttt{\ \ image:\ \ (no instruction)}\\
    \texttt{\ \ text:\ \ \ (no instruction)}
  \end{tcolorbox}
\end{tcolorbox}
\end{table*}

% ---- RzenEmbed ----
\begin{table*}[!ht]
\centering\small
\begin{tcolorbox}[
    enhanced,
    title={\normalsize\textbf{RzenEmbed}},
    colframe=blue!45!black, colback=blue!2!white,
    colbacktitle=blue!10!white, coltitle=blue!50!black,
    fonttitle=\bfseries, halign title=center,
    boxrule=1pt, arc=4pt, left=6pt, right=6pt, top=5pt, bottom=5pt,
]
  \begin{tcolorbox}[
      enhanced, sharp corners, colframe=gray!50, colback=gray!8,
      title=\small\textbf{PROMPT TEMPLATE}, coltitle=black,
      colbacktitle=gray!20, fonttitle=\small\bfseries,
      boxrule=0.5pt, left=5pt, right=5pt, top=3pt, bottom=3pt,
  ]
    \texttt{\# [Query]}\\
    \texttt{\ \ joint:\ \ "Represent the given image-text query."}\\
    \texttt{\ \ image:\ \ "Represent the given query image."}\\
    \texttt{\ \ text:\ \ \ "Represent the given query text."}\\[4pt]
    \texttt{\# [Gallery]}\\
    \texttt{\ \ joint:\ \ "Represent the given image."\ (image part; caption appended)}\\
    \texttt{\ \ image:\ \ "Represent the given image."}\\
    \texttt{\ \ text:\ \ \ (no instruction)}
  \end{tcolorbox}
\end{tcolorbox}
\end{table*}

\subsection{Score Normalisation and Aggregation}
For Paradigm I models the cosine similarity is bounded in $[-1,1]$; for Paradigm II.a models the dot-product can be unbounded. MOOR uses per-query score variance for path weighting, so no inter-model rescaling is needed for fusion. R@$K$ is computed per query, treating any of the (mean $4.77$) GT items as a hit, then macro-averaged.

\subsection{Reproducibility}
\label{sec:appendix_errorbars}
All corruptions are deterministic given a fixed operator and severity ($\texttt{seed} = 42 + \text{hash}$); gallery order is fixed. Reported numbers are on the single official \benchname{} test split.

% ---- D: Additional Results & Analysis ----
\section{Additional Results and Analysis}
\label{app:additional_results}

\subsection{Additional Experimental Results}
\label{sec:appendix_more_results}

\subsubsection{Empty-Text Query Ablation}
\label{sec:appendix_text_dropped}

\begin{figure}[t]
  \centering
  \includegraphics[width=\columnwidth]{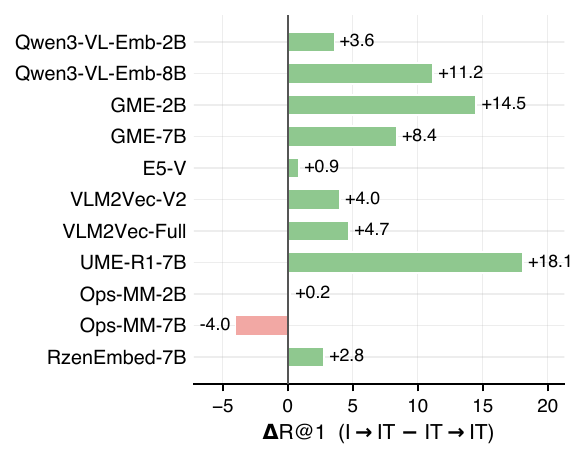}
  \caption{%
    \textbf{Effect of removing the query text on clean IT$\to$IT retrieval.}
    For each VLM embedding model, we replace the user question with an empty
    string while keeping the snapped image and the IT gallery fixed.
    Bars report $\Delta$R@1 between image-only query retrieval
    (I$\to$IT) and full query retrieval (IT$\to$IT).
    Positive values indicate that dropping the text \emph{improves} retrieval,
    consistent with \emph{coarse-text drag}: coarse user questions are
    over-weighted under fixed fusion and dilute a strong visual ranking.
    Ops-MM-7B is the only model for which the query text helps on clean inputs.}
  \label{fig:text_dropped}
\end{figure}

\Cref{fig:text_dropped} reports the R@1 change when the user question is replaced with an empty string under clean \textbf{IT$\rightarrow$IT} retrieval. Positive values indicate that removing the text improves retrieval, providing a direct check of coarse-text drag (\Cref{sec:exp_text_artifacts}).

\subsubsection{Per-Severity Super-Additivity Breakdown}
\label{sec:appendix_superadd_sev}

\Cref{tab:super_additivity} in the main text reports the super-additivity
gap $\delta = \text{Obs.} - \text{Exp.}$ averaged over the three severity
levels $s_1$/$s_2$/$s_3$.
Positive $\delta$ indicates super-additivity (joint drop exceeds the additive prediction); negative $\delta$ indicates sub-additivity.
\Cref{tab:super_additivity_appendix} provides the full per-severity
breakdown, showing $\delta$ at each severity level for all six joint
conditions and all 16 models.
Key observations: (i)~for most super-additive models (e.g., GME-2B/7B),
$\delta$ remains large across all severities; (ii)~for sub-additive
models (Ops-MM, RzenEmbed-7B, VLM2Vec), the sub-additivity tends to
grow stronger at higher severity ($s_3$), suggesting a compensatory
mechanism that activates under extreme dual-modality corruption.

\begin{table*}[t]
\centering

\resizebox{\linewidth}{!}{%
\setlength{\tabcolsep}{3pt}\renewcommand{\arraystretch}{1.05}\scriptsize
\begin{tabular}{l c | ccc | ccc | ccc | ccc | ccc | ccc | ccc}
\toprule
\textbf{Model} & \textbf{Base.} & \multicolumn{3}{c|}{\textbf{C+C}} & \multicolumn{3}{c|}{\textbf{D+D}} & \multicolumn{3}{c|}{\textbf{L+W}} & \multicolumn{3}{c|}{\textbf{M+S}} & \multicolumn{3}{c|}{\textbf{E+W}} & \multicolumn{3}{c|}{\textbf{R+S}} & \multicolumn{3}{c}{\textbf{Mean}} \\
\cmidrule(lr){3-20}\cmidrule(lr){21-23}
 & $s_1$ & $s_2$ & $s_3$ & $s_1$ & $s_2$ & $s_3$ & $s_1$ & $s_2$ & $s_3$ & $s_1$ & $s_2$ & $s_3$ & $s_1$ & $s_2$ & $s_3$ & $s_1$ & $s_2$ & $s_3$ & $s_1$ & $s_2$ & $s_3$ \\
\midrule
\rowcolor{gray!10} \multicolumn{23}{c}{\textit{Dual-Encoder}} \\
CLIP\mbox{-}ViT\mbox{-}L/14 & 63.8 & \textbf{20.4} & \textbf{22.6} & \textbf{24.9} & \textbf{22.5} & \textbf{25.3} & \textbf{26.3} & \textbf{16.5} & \textbf{17.1} & \textbf{16.6} & \textbf{16.0} & \textbf{15.8} & \textbf{14.7} & \textbf{15.9} & \textbf{18.7} & \textbf{15.5} & \textbf{31.1} & \textbf{32.4} & \textbf{32.2} & \textbf{20.4} & \textbf{22.0} & \textbf{21.7} \\
SigLIP\mbox{-}SO400M & 68.1 & \textbf{5.1} & \textbf{5.5} & \textbf{6.0} & \textbf{4.3} & \textbf{2.0} & \textbf{5.5} & \textbf{2.4} & \textbf{2.6} & \textbf{5.2} & \textbf{5.0} & \textbf{4.5} & \textbf{8.4} & \textbf{0.6} & \textbf{1.7} & \textbf{2.4} & \textbf{6.9} & \textbf{8.3} & \textbf{6.1} & \textbf{4.1} & \textbf{4.1} & \textbf{5.6} \\
SigLIP2\mbox{-}SO400M & 49.7 & \textbf{12.8} & \textbf{12.1} & \textbf{14.1} & \textbf{10.2} & \textbf{9.1} & \textbf{4.4} & \textbf{11.8} & \textbf{13.5} & \textbf{12.2} & \textbf{11.2} & \textbf{12.3} & \textbf{15.4} & \textbf{8.9} & \textbf{8.6} & \textbf{8.3} & \textbf{8.8} & \textbf{8.3} & \textbf{10.3} & \textbf{10.6} & \textbf{10.7} & \textbf{10.8} \\
BLIP\mbox{-}ITM\mbox{-}L & 64.4 & \textbf{5.4} & \textbf{6.9} & \textbf{5.5} & \textbf{4.6} & \textbf{6.1} & \textbf{11.3} & \textbf{5.8} & \textbf{5.4} & \textbf{5.8} & \textbf{5.0} & \textbf{5.3} & \textbf{7.1} & \textbf{7.5} & \textbf{6.7} & \textbf{5.5} & \textbf{11.1} & \textbf{15.6} & \textbf{20.2} & \textbf{6.6} & \textbf{7.7} & \textbf{9.3} \\
\rowcolor{gray!10} \multicolumn{23}{c}{\textit{VLM Embedding}} \\
GME\mbox{-}2B & 60.4 & \textbf{47.6} & \textbf{47.7} & \textbf{46.0} & \textbf{56.5} & \textbf{53.0} & \textbf{46.7} & \textbf{45.8} & \textbf{46.2} & \textbf{28.6} & \textbf{51.8} & \textbf{46.3} & \textbf{33.7} & \textbf{41.5} & \textbf{40.4} & \textbf{35.6} & -5.8 & -3.0 & -5.2 & \textbf{39.6} & \textbf{38.4} & \textbf{30.9} \\
Qwen3\mbox{-}VL\mbox{-}Emb\mbox{-}2B & 77.8 & \textbf{12.4} & \textbf{12.9} & \textbf{14.5} & \textbf{11.9} & \textbf{15.9} & \textbf{16.1} & \textbf{10.5} & \textbf{9.8} & \textbf{12.3} & \textbf{3.6} & \textbf{4.6} & \textbf{7.0} & \textbf{10.6} & \textbf{9.9} & \textbf{10.5} & \textbf{15.1} & \textbf{13.9} & \textbf{14.2} & \textbf{10.7} & \textbf{11.1} & \textbf{12.4} \\
Ops\mbox{-}MM\mbox{-}2B & 77.6 & \textbf{1.0} & \textbf{1.2} & \textbf{0.2} & -1.1 & -0.5 & -3.4 & -0.2 & -0.8 & -13.8 & -0.8 & -2.0 & -11.2 & \textbf{0.8} & -3.4 & -5.0 & -1.6 & -1.0 & -0.6 & -0.3 & -1.1 & -5.6 \\
Jina\mbox{-}V4 & 66.9 & \textbf{0.8} & -1.0 & \textbf{1.4} & \textbf{1.8} & \textbf{3.7} & \textbf{3.1} & \textbf{0.6} & -0.3 & -0.3 & -1.5 & \textbf{0.2} & \textbf{3.1} & \textbf{0.5} & -0.3 & -0.2 & -1.2 & \textbf{4.0} & -0.0 & \textbf{0.2} & \textbf{1.0} & \textbf{1.2} \\
VLM2Vec\mbox{-}Full & 43.1 & -15.0 & -14.2 & -10.4 & -13.8 & -11.2 & -2.0 & -13.8 & -14.8 & -10.7 & -18.6 & -17.7 & -9.6 & -13.4 & -11.4 & -8.4 & -37.6 & -32.9 & -29.2 & -18.7 & -17.0 & -11.7 \\
VLM2Vec\mbox{-}V2 & 50.4 & -11.8 & -12.5 & -11.8 & -12.2 & -11.1 & -0.2 & -10.5 & -11.7 & -6.2 & -12.4 & -11.3 & -2.3 & -13.2 & -13.4 & -11.0 & -21.1 & -19.3 & -19.6 & -13.5 & -13.2 & -8.5 \\
E5\mbox{-}V & 59.8 & \textbf{0.7} & \textbf{1.2} & \textbf{4.2} & \textbf{1.4} & \textbf{2.8} & \textbf{3.0} & -0.0 & \textbf{0.3} & -0.8 & \textbf{7.3} & \textbf{8.9} & \textbf{11.2} & \textbf{0.1} & \textbf{1.9} & \textbf{0.6} & \textbf{2.0} & \textbf{5.5} & \textbf{7.2} & \textbf{1.9} & \textbf{3.4} & \textbf{4.2} \\
GME\mbox{-}7B & 69.8 & \textbf{48.1} & \textbf{48.4} & \textbf{46.5} & \textbf{50.1} & \textbf{48.5} & \textbf{38.7} & \textbf{45.8} & \textbf{44.0} & \textbf{24.9} & \textbf{51.7} & \textbf{46.0} & \textbf{30.9} & \textbf{44.1} & \textbf{40.0} & \textbf{35.4} & \textbf{3.2} & \textbf{4.5} & \textbf{6.2} & \textbf{40.5} & \textbf{38.6} & \textbf{30.4} \\
UME\mbox{-}R1\mbox{-}7B & 71.0 & \textbf{11.6} & \textbf{11.7} & \textbf{12.6} & \textbf{11.3} & \textbf{12.4} & \textbf{13.0} & \textbf{9.1} & \textbf{10.3} & \textbf{7.9} & \textbf{9.6} & \textbf{7.5} & \textbf{9.8} & \textbf{9.5} & \textbf{7.2} & \textbf{7.5} & \textbf{9.9} & \textbf{12.5} & \textbf{13.7} & \textbf{10.2} & \textbf{10.3} & \textbf{10.7} \\
RzenEmbed\mbox{-}7B & 75.2 & \textbf{0.2} & -0.7 & -2.2 & -0.8 & \textbf{0.5} & -4.2 & \textbf{0.3} & -1.2 & -18.0 & \textbf{0.7} & -0.7 & -8.0 & \textbf{0.8} & -2.1 & -5.6 & -0.4 & \textbf{0.1} & \textbf{2.2} & \textbf{0.1} & -0.7 & -6.0 \\
Ops\mbox{-}MM\mbox{-}7B & 79.1 & \textbf{0.6} & \textbf{0.1} & -0.2 & \textbf{0.4} & -0.3 & -6.0 & -0.8 & -0.9 & -10.1 & \textbf{0.4} & \textbf{0.2} & -9.0 & \textbf{0.7} & -0.9 & -2.4 & \textbf{0.5} & \textbf{0.6} & \textbf{0.7} & \textbf{0.3} & -0.2 & -4.5 \\
Qwen3\mbox{-}VL\mbox{-}Emb\mbox{-}8B & 77.5 & \textbf{11.8} & \textbf{13.2} & \textbf{14.3} & \textbf{13.8} & \textbf{14.4} & \textbf{15.2} & \textbf{12.3} & \textbf{12.3} & \textbf{8.6} & \textbf{10.2} & \textbf{9.1} & \textbf{10.8} & \textbf{12.0} & \textbf{12.0} & \textbf{10.6} & \textbf{15.2} & \textbf{15.7} & \textbf{13.0} & \textbf{12.6} & \textbf{12.8} & \textbf{12.1} \\
\midrule
\textit{Mean} &  & \textbf{9.5} & \textbf{9.7} & \textbf{10.4} & \textbf{10.1} & \textbf{10.7} & \textbf{10.5} & \textbf{8.5} & \textbf{8.2} & \textbf{3.9} & \textbf{8.7} & \textbf{8.1} & \textbf{7.0} & \textbf{7.9} & \textbf{7.2} & \textbf{6.2} & \textbf{2.3} & \textbf{4.1} & \textbf{4.5} & \textbf{7.8} & \textbf{8.0} & \textbf{7.1} \\
\bottomrule
\end{tabular}}
\caption{%
  Appendix: Per-severity super-additivity gap $\delta = \text{Obs.} - \text{Exp.}$
  (R@1 drop in pp) for six joint corruption conditions at severity levels
  $s_1$, $s_2$, $s_3$. Positive (bold) = super-additive; negative = sub-additive.
  Condition abbreviations follow \Cref{tab:super_additivity}.}
\label{tab:super_additivity_appendix}
\end{table*}

\subsubsection{Coarse-Text Drag: Full Results and Causal Intervention}
\label{sec:appendix_drag_full}

\Cref{tab:coarse_text_drag_full} reports the drag ($\Delta$R@1 $=$ IT$\to$IT $-$ I$\to$I under clean input) for all 16 models. All 16 models show negative drag (range $-3.50$ to $-23.93$), confirming that fixed fusion over-weighting coarse text is a model-independent property rather than an artifact of any single architecture.

\begin{table}[h]
\centering\small

\begin{tabular}{lccc}
\toprule
\textbf{Model} & \textbf{I$\to$I} & \textbf{IT$\to$IT} & \textbf{Drag} \\
\midrule
CLIP ViT-L/14   & 73.45 & 49.52 & $-23.93$ \\
BLIP ITM-L      & 73.71 & 57.29 & $-16.42$ \\
Ops-MM-7B       & 80.52 & 69.61 & $-10.91$ \\
RzenEmbed       & 80.61 & 75.55 & $-5.06$ \\
\midrule
Mean (16 models) & --- & --- & $-11.21$ \\
\bottomrule
\end{tabular}
\caption{Coarse-text drag (clean IT$\to$IT $-$ I$\to$I R@1) for representative models; full 16-model table follows the same pattern.}
\label{tab:coarse_text_drag_full}
\end{table}

\paragraph{Causal intervention.} To directly test whether coarse \emph{granularity}, rather than the text channel itself, drives the drag, we replace the coarse question (``What flower is this?'') with entity-label text (``A tulip.'') on GME-7B. This reduces the drag from $-7.68$ to $-0.17$ R@1 points---a $97.8\%$ elimination---confirming the granularity-based mechanism.

\subsubsection{Text-Essential Subset}
\label{sec:appendix_text_essential}

To further test whether the text-underuse conclusion is calibration-scoped rather than a claim that text is universally unhelpful, we construct a Tier-1 \emph{text-essential} subset ($n=69/1{,}145$): queries with discriminative text modifiers (color, material, quantity, shape, position), at least three same-category hard negatives, and image-only retrieval failure on a reference model. \Cref{tab:text_essential} shows that coarse-text drag is reduced for $14$ of $16$ models on this subset (mean improvement $+1.28$ R@1 pp relative to the full-set drag), with the largest gains on strong dual-tower models and one model (GME-2B) reaching zero drag.

\begin{table}[h]
\centering\small

\begin{tabular}{lccc}
\toprule
\textbf{Model} & \makecell{\textbf{Full-set}\\\textbf{drag}} & \makecell{\textbf{Tier-1}\\\textbf{drag}} & $\bm{\Delta}$ \\
\midrule
SigLIP-SO400M & $-15.63$ & $-1.45$ & $+14.18$ \\
BLIP ITM-L    & $-16.42$ & $-4.34$ & $+12.08$ \\
CLIP ViT-L/14 & $-23.93$ & $-10.15$ & $+13.78$ \\
Jina-Emb-v4   & $-10.13$ & $-2.90$ & $+7.23$ \\
GME-2B        & $-3.58$  & $0.00$  & $+3.58$ \\
\midrule
Mean (16 models) & $-10.06$ & $-8.78$ & $+1.28$ \\
\bottomrule
\end{tabular}
\caption{Full-set vs.\ text-essential-subset (Tier-1) coarse-text drag.}
\label{tab:text_essential}
\end{table}

This supports the calibration-scoped reading: text is genuinely useful when it carries discriminative signal, but fixed fusion over-weights it under the coarse, category-level phrasing that defines snap-and-ask. The two exceptions are large-backbone VLMs with stronger text encoding, where discriminative modifiers also appear in same-category hard negatives, partially offsetting the benefit.

\subsection{Extended Benchmark Analyses}
\label{sec:appendix_extended_analyses}

\subsubsection{Query Stability under Joint Corruption}
\label{sec:appendix_drag_rate}

\begin{figure*}[t]
  \centering
  \includegraphics[width=\textwidth]{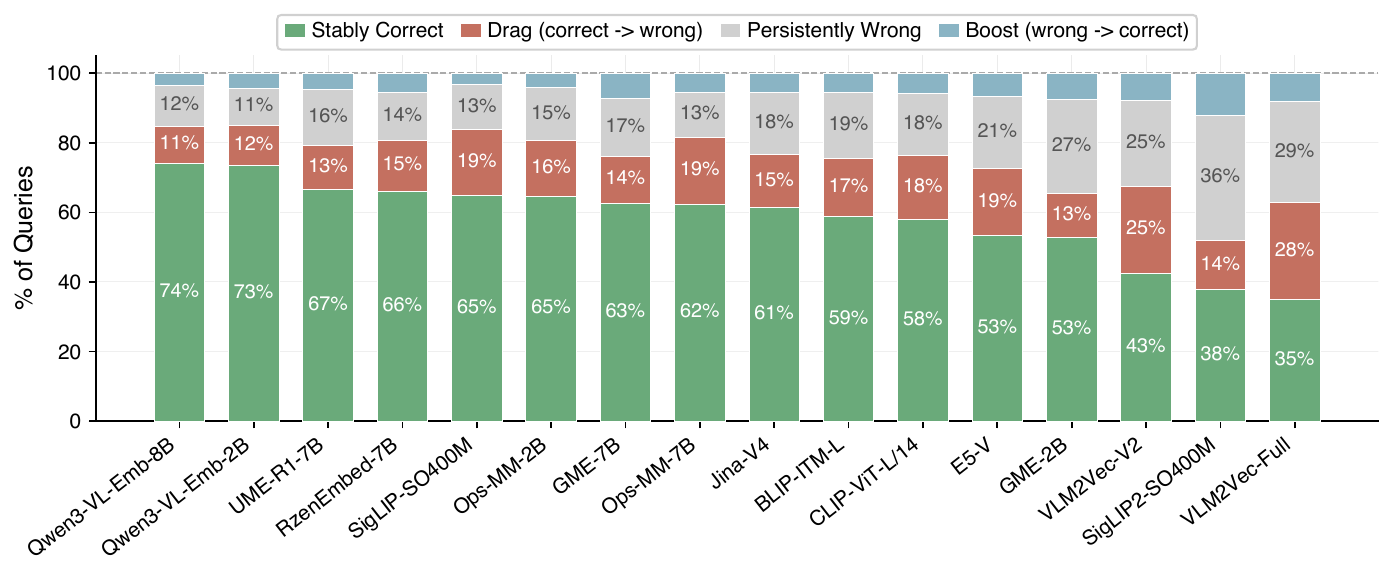}
  \caption{%
    \textbf{Query-level stability under joint corruption (IT$\to$IT).}
    Stacked bars show the fraction of queries in each of four classes
    per model (16 models, sorted by stably-correct rate):
    stably correct (green), drag from correct to wrong (red),
    persistently wrong (grey), and boost from wrong to correct (blue).
    Percentages are shown inside segments $\geq 5\%$; small boost
    segments are omitted for readability.}
  \label{fig:a13_drag_rate}
\end{figure*}

\Cref{fig:a13_drag_rate} decomposes per-query outcomes under joint
corruption into four buckets---stably correct, dragged from correct to
incorrect, persistently wrong, and boosted from incorrect to correct---
for each model, sorted by the fraction of queries that remain correct.

\subsubsection{Image-Side vs.\ Text-Side Corruption Asymmetry}
\label{sec:appendix_asymmetry}

\begin{figure*}[t]
  \centering
  \includegraphics[width=\textwidth]{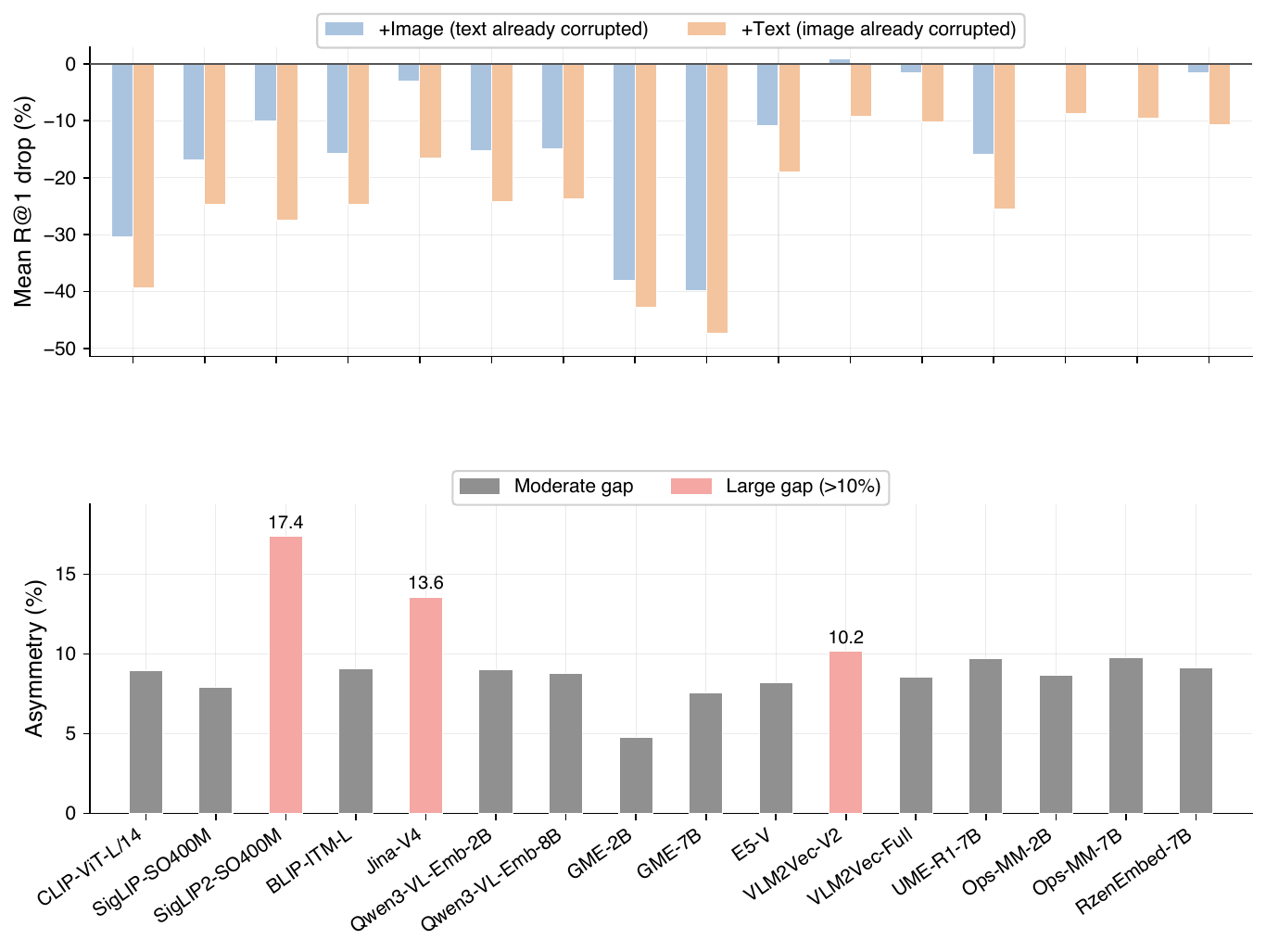}
  \caption{%
    \textbf{Asymmetry of Image-Side vs.\ Text-Side Corruption.}
    (a)~Extra R@1 drop from the second corruption.
    (b)~Text-side extra drop minus image-side extra drop (percentage points).}
  \label{fig:a22_asymmetry}
\end{figure*}

\Cref{fig:a22_asymmetry} compares the \emph{extra} R@1 drop incurred when
a second modality is corrupted, beyond the first.
Most models suffer more from text-side than image-side corruptions on this
metric, explaining why text-channel paradoxes and coarse-text drag are
central failure modes.

\subsubsection{Domain Difficulty}
\label{sec:appendix_domain_diff}

\begin{figure}[t]
  \centering
  \includegraphics[width=\columnwidth]{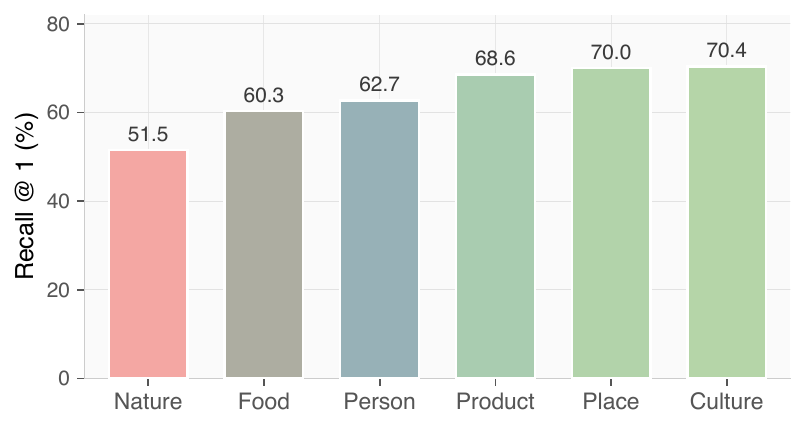}
  \caption{%
    \textbf{Domain difficulty in IT$\to$IT retrieval.}
    Mean Recall@1 under clean joint IT queries, aggregated across
    models and queries per domain (bars sorted hard$\to$easy).
    Color encodes difficulty (warmer = lower Recall@1).}
  \label{fig:domain_difficulty}
\end{figure}

\Cref{fig:domain_difficulty} aggregates clean joint IT retrieval
(HR@1) over all models and queries within each of the six image
domains in \benchname{}.
Difficulty is far from uniform: \texttt{nature} is the hardest
($51.5\%$ mean HR@1), while \texttt{culture} and \texttt{place}
exceed $70\%$, confirming that the benchmark spans heterogeneous
visual semantics.

\subsubsection{Per-Model Corruption Sensitivity}
\label{sec:appendix_model_sensitivity}

\begin{figure*}[t]
  \centering
  \includegraphics[width=\textwidth]{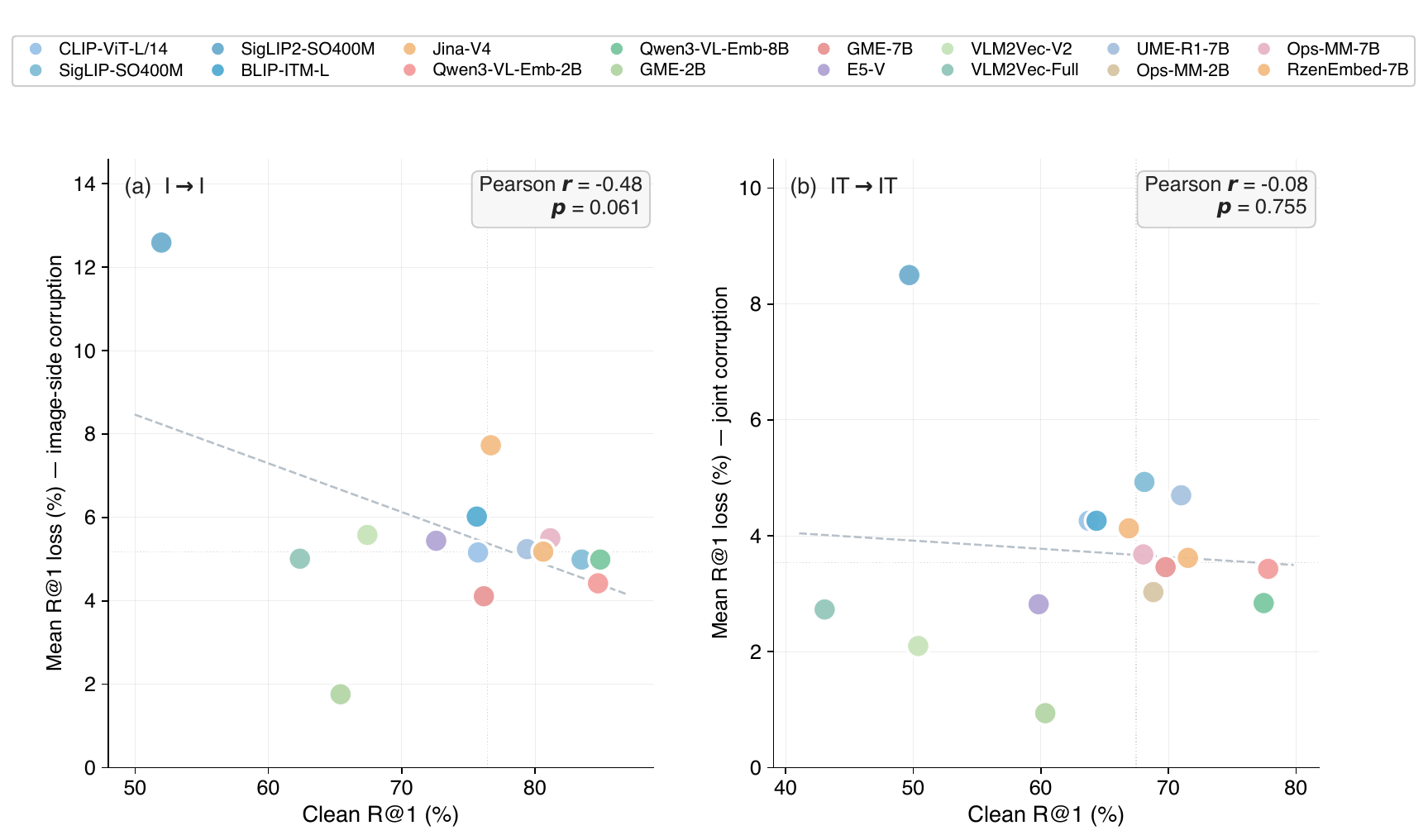}
  \caption{%
    \textbf{Clean R@1 vs.\ corruption loss under two retrieval modes.}
    (a)~I$\to$I: clean R@1 vs.\ mean R@1 loss under image-side
    corruptions.
    (b)~IT$\to$IT: clean R@1 vs.\ mean R@1 loss under joint
    corruptions.
    Each colored point is one model (legend above); dashed line
    shows the Pearson trend; statistics are in the upper-right box.
    Weak correlation in (b) shows that clean IT$\to$IT accuracy
    alone does not predict joint-corruption robustness.}
  \label{fig:a31_robustness}
\end{figure*}

\Cref{fig:a31_robustness} contrasts each model's clean joint R@1
(open markers) with its mean R@1 averaged over all joint corruption
conditions (filled markers).
High clean scores do not imply robustness---e.g., Qwen3-VL-Emb-8B
ranks among the strongest on clean inputs yet still loses
substantially under joint corruption.

% \subsubsection{LLM Query Refinement}
% \label{sec:appendix_query_refinement}

% \Cref{fig:query_refinement} compares three query formulations:
% \emph{Original} (coarse category-level), \emph{Sharper Q}
% (entity-discriminating rewrite), and \emph{Descriptive Q}
% (full entity description).
% Gains are small and inconsistent for most VLMs
% ($|\Delta|\!\lesssim\!2$~pp), while several Dual-Tower encoders
% show larger swings (e.g., SigLIP-SO400M $+6.7$~pp under Sharper Q),
% confirming that LLM rewrites are not a reliable fix for Coarse-Text
% Drag.

% \begin{figure}[t]
%   \centering
%   \includegraphics[width=\columnwidth]{fig9_query_refinement_l3l4}
%   \caption{%
%     \textbf{LLM Query Refinement Across All Models.}
%     $\Delta$R@1 (Sharper / Descriptive minus original query) for all
%     $16$ models.}
%   \label{fig:query_refinement}
% \end{figure}

\subsubsection{Sensitivity and Robustness Tables}
\label{sec:appendix_sensitivity_tables}

\begin{table}[t]
  \centering
  \small

  \begin{tabular}{lcc}
    \toprule
    \textbf{Primitive} & \textbf{R@1 drop (pp)} & \textbf{GT rank displacement} \\
    \midrule
    {Remove}    & $-26.8$ & $7.84$ positions \\
    {Add}       & $-17.9$ & $4.97$ positions \\
    {Degrade}   & $-14.5$ & $3.47$ positions \\
    {Transform} & $\phantom{0}{-7.9}$ & $1.38$ positions \\
    \bottomrule
  \end{tabular}
\caption{%
    \textbf{Retrieval impact at severe severity ($s_3$)} by image primitive
    (model-average over 4 models).}
  \label{tab:failure_quant}
\end{table}

  \begin{table*}[t]
  \centering\scriptsize\setlength{\tabcolsep}{6pt}\renewcommand{\arraystretch}{0.95}

  \begin{tabular}{l c | c | c c c | c c c c}
  \toprule
  \multicolumn{1}{c}{\textbf{Model}} & Mode & Clean & Char & Word & Sent & Degrade & Transform & Remove & Add \\
  \cmidrule(lr){4-6}\cmidrule(lr){7-10}
   &  &  & \multicolumn{3}{c|}{\textit{Text Pert.}} & \multicolumn{4}{c}{\textit{Image Pert. (avg)}} \\
  \midrule
  \multirow{5}{*}{GME-2B} & IT$\rightarrow$IT & 61.8 & 60.6 & 61.5 & 58.9 & 58.1 & 60.0 & 60.2 & 61.4 \\
   & I$\rightarrow$I & 65.4 & 65.4 & 65.4 & 65.4 & 60.1 & 64.1 & 64.6 & 62.5 \\
   & I$\rightarrow$IT & 53.5 & 53.5 & 53.5 & 53.5 & 49.2 & 52.3 & 53.2 & 52.8 \\
   & T$\rightarrow$T & 22.5 & 15.1 & 22.2 & 9.0 & 22.5 & 22.5 & 22.5 & 22.5 \\
   & T$\rightarrow$IT & 24.1 & 15.7 & 23.8 & 8.8 & 24.1 & 24.1 & 24.1 & 24.1 \\
  \cmidrule(lr){1-10}
  \multirow{5}{*}{Qwen3-VL-Emb-2B} & IT$\rightarrow$IT & 64.6 & 66.1 & 66.4 & 71.0 & 60.2 & 62.4 & 54.5 & 60.8 \\
   & I$\rightarrow$I & 84.7 & 84.7 & 84.7 & 84.7 & 80.0 & 82.0 & 69.2 & 80.4 \\
   & I$\rightarrow$IT & 81.4 & 81.4 & 81.4 & 81.4 & 76.6 & 78.4 & 66.5 & 76.8 \\
   & T$\rightarrow$T & 21.3 & 13.2 & 20.7 & 8.5 & 21.3 & 21.3 & 21.3 & 21.3 \\
   & T$\rightarrow$IT & 16.2 & 8.5 & 14.5 & 5.5 & 16.2 & 16.2 & 16.2 & 16.2 \\
  \cmidrule(lr){1-10}
  \multirow{5}{*}{Ops-MM-2B} & IT$\rightarrow$IT & 77.5 & 78.0 & 77.6 & 78.6 & 71.7 & 74.8 & 73.1 & 70.5 \\
   & I$\rightarrow$I & 79.6 & 79.6 & 79.6 & 79.6 & 70.4 & 76.5 & 74.5 & 68.4 \\
   & I$\rightarrow$IT & 48.3 & 48.3 & 48.3 & 48.3 & 41.4 & 44.5 & 42.8 & 40.1 \\
   & T$\rightarrow$T & 22.3 & 15.7 & 21.4 & 9.7 & 22.3 & 22.3 & 22.3 & 22.3 \\
   & T$\rightarrow$IT & 22.9 & 15.2 & 22.9 & 8.2 & 22.9 & 22.9 & 22.9 & 22.9 \\
  \cmidrule(lr){1-10}
  \multirow{5}{*}{Jina-V4} & IT$\rightarrow$IT & 65.8 & 69.5 & 65.9 & 71.0 & 61.9 & 63.2 & 55.7 & 61.3 \\
   & I$\rightarrow$I & 76.0 & 76.0 & 76.0 & 76.0 & 67.6 & 71.3 & 56.2 & 67.8 \\
   & I$\rightarrow$IT & 68.7 & 68.7 & 68.7 & 68.7 & 60.1 & 63.7 & 51.8 & 60.6 \\
   & T$\rightarrow$T & 21.2 & 15.4 & 21.5 & 9.7 & 21.2 & 21.2 & 21.2 & 21.2 \\
   & T$\rightarrow$IT & 24.9 & 17.6 & 24.2 & 10.7 & 24.9 & 24.9 & 24.9 & 24.9 \\
  \cmidrule(lr){1-10}
  \multirow{5}{*}{VLM2Vec-Full} & IT$\rightarrow$IT & 57.6 & 57.8 & 58.0 & 61.3 & 53.4 & 53.6 & 50.3 & 52.1 \\
   & I$\rightarrow$I & 62.4 & 62.4 & 62.4 & 62.4 & 56.8 & 59.5 & 54.0 & 54.4 \\
   & I$\rightarrow$IT & 57.6 & 57.6 & 57.6 & 57.6 & 53.4 & 53.6 & 50.3 & 52.1 \\
   & T$\rightarrow$T & 20.6 & 12.6 & 21.6 & 6.4 & 20.6 & 20.6 & 20.6 & 20.6 \\
   & T$\rightarrow$IT & 57.6 & 57.8 & 58.0 & 61.3 & 57.6 & 57.6 & 57.6 & 57.6 \\
  \cmidrule(lr){1-10}
  \multirow{5}{*}{VLM2Vec-V2} & IT$\rightarrow$IT & 62.5 & 63.8 & 64.8 & 66.8 & 59.0 & 60.4 & 55.6 & 59.5 \\
   & I$\rightarrow$I & 67.4 & 67.4 & 67.4 & 67.4 & 61.5 & 64.1 & 55.4 & 59.6 \\
   & I$\rightarrow$IT & 62.5 & 62.5 & 62.5 & 62.5 & 59.0 & 60.4 & 55.6 & 59.5 \\
   & T$\rightarrow$T & 20.0 & 12.8 & 20.0 & 7.5 & 20.0 & 20.0 & 20.0 & 20.0 \\
   & T$\rightarrow$IT & 62.5 & 63.8 & 64.8 & 66.8 & 62.5 & 62.5 & 62.5 & 62.5 \\
  \cmidrule(lr){1-10}
  \multirow{5}{*}{E5-V} & IT$\rightarrow$IT & 59.1 & 59.2 & 59.1 & 58.7 & 56.0 & 54.8 & 52.3 & 54.7 \\
   & I$\rightarrow$I & 71.8 & 71.8 & 71.8 & 71.8 & 67.5 & 66.4 & 61.4 & 63.7 \\
   & I$\rightarrow$IT & 60.0 & 60.0 & 60.0 & 60.0 & 55.5 & 55.9 & 50.9 & 52.8 \\
   & T$\rightarrow$T & 22.1 & 16.4 & 21.9 & 8.6 & 22.1 & 22.1 & 22.1 & 22.1 \\
   & T$\rightarrow$IT & 24.0 & 17.6 & 23.9 & 8.6 & 24.0 & 24.0 & 24.0 & 24.0 \\
  \cmidrule(lr){1-10}
  \multirow{5}{*}{GME-7B} & IT$\rightarrow$IT & 68.0 & 68.9 & 68.2 & 69.7 & 63.8 & 64.4 & 61.6 & 64.5 \\
   & I$\rightarrow$I & 76.3 & 76.3 & 76.3 & 76.3 & 70.4 & 72.8 & 66.7 & 69.3 \\
   & I$\rightarrow$IT & 65.3 & 65.3 & 65.3 & 65.3 & 59.3 & 60.7 & 55.3 & 58.8 \\
   & T$\rightarrow$T & 24.9 & 18.1 & 24.3 & 9.9 & 24.9 & 24.9 & 24.9 & 24.9 \\
   & T$\rightarrow$IT & 27.6 & 20.2 & 26.3 & 10.9 & 27.6 & 27.6 & 27.6 & 27.6 \\
  \cmidrule(lr){1-10}
  \multirow{5}{*}{UME-R1-7B} & IT$\rightarrow$IT & 60.2 & 59.7 & 60.2 & 62.5 & 55.2 & 56.6 & 49.3 & 55.1 \\
   & I$\rightarrow$I & 79.0 & 79.0 & 79.0 & 79.0 & 72.7 & 75.2 & 65.6 & 73.0 \\
   & I$\rightarrow$IT & 60.2 & 60.2 & 60.2 & 60.2 & 55.2 & 56.6 & 49.3 & 55.1 \\
   & T$\rightarrow$T & 22.6 & 16.4 & 23.0 & 10.3 & 22.6 & 22.6 & 22.6 & 22.6 \\
   & T$\rightarrow$IT & 60.2 & 59.7 & 60.2 & 62.5 & 60.2 & 60.2 & 60.2 & 60.2 \\
  \cmidrule(lr){1-10}
  \multirow{5}{*}{RzenEmbed-7B} & IT$\rightarrow$IT & 75.2 & 76.8 & 76.5 & 77.1 & 69.0 & 72.8 & 72.3 & 70.2 \\
   & I$\rightarrow$I & 80.6 & 80.6 & 80.6 & 80.6 & 69.1 & 77.2 & 75.3 & 69.0 \\
   & I$\rightarrow$IT & 50.4 & 50.4 & 50.4 & 50.4 & 42.6 & 47.4 & 44.7 & 41.4 \\
   & T$\rightarrow$T & 23.8 & 18.3 & 23.6 & 9.7 & 23.8 & 23.8 & 23.8 & 23.8 \\
   & T$\rightarrow$IT & 23.7 & 18.5 & 23.7 & 9.2 & 23.7 & 23.7 & 23.7 & 23.7 \\
  \cmidrule(lr){1-10}
  \multirow{5}{*}{Ops-MM-7B} & IT$\rightarrow$IT & 79.1 & 79.3 & 79.4 & 77.7 & 72.1 & 76.0 & 75.2 & 72.1 \\
   & I$\rightarrow$I & 80.5 & 80.5 & 80.5 & 80.5 & 69.4 & 76.2 & 73.4 & 69.2 \\
   & I$\rightarrow$IT & 49.8 & 49.8 & 49.8 & 49.8 & 42.6 & 46.6 & 44.7 & 42.5 \\
   & T$\rightarrow$T & 23.2 & 17.9 & 22.9 & 10.4 & 23.2 & 23.2 & 23.2 & 23.2 \\
   & T$\rightarrow$IT & 24.4 & 17.6 & 23.2 & 8.8 & 24.4 & 24.4 & 24.4 & 24.4 \\
  \cmidrule(lr){1-10}
  \multirow{5}{*}{Qwen3-VL-Emb-8B} & IT$\rightarrow$IT & 65.5 & 66.9 & 67.2 & 69.6 & 61.8 & 63.1 & 56.2 & 62.7 \\
   & I$\rightarrow$I & 83.8 & 83.8 & 83.8 & 83.8 & 79.4 & 81.6 & 70.4 & 79.0 \\
   & I$\rightarrow$IT & 76.7 & 76.7 & 76.7 & 76.7 & 72.2 & 73.6 & 63.3 & 72.5 \\
   & T$\rightarrow$T & 23.2 & 15.2 & 22.8 & 8.3 & 23.2 & 23.2 & 23.2 & 23.2 \\
   & T$\rightarrow$IT & 23.7 & 16.1 & 23.2 & 9.0 & 23.7 & 23.7 & 23.7 & 23.7 \\
  \bottomrule
  \end{tabular}
\caption{  Retrieval-mode ablation on \benchname{} (Recall@1, \%).   \textbf{IT$\to$IT} is the joint multimodal mode from Tab.~\ref{tab:main_results}.   Image columns show averages over three severities.   N/A cells (mode cannot be affected by that corruption type) are filled   with the clean baseline value.  \textit{VLM Embedding} models.}
  \label{tab:ablation_modes_vlm}
\end{table*}

\subsection{Failure Mode Analysis}
\label{sec:appendix_failure_analysis}

We inspect approximately $1{,}000$ retrieval failures---queries on which the correct entity is \textbf{not} ranked first---under image-only and joint corruption, drawing on two representative encoders (CLIP-ViT-L/14 and Qwen3-VL-Emb-2B).
Examination of these cases reveals four recurring failure patterns that together account for the large majority of observed errors. This manual failure-mode analysis is conceptually related to anomaly-detection-style formulations explored in adjacent domains such as industrial video understanding~\citep{Zhao2026MMVIADMM}.

% ----------------------------------------------------------
\subsubsection{Failure Patterns}
\label{sec:appendix_failure_patterns}
% ----------------------------------------------------------

\paragraph{Quality Degradation.}
The corruption removes or destroys the discriminative visual content needed to identify the target entity, leaving the model without a reliable signal.
This pattern dominates at severe severity ($s_3$) under \textsc{Remove} and \textsc{Degrade}: aggressive cropping removes the entity from the frame entirely, while extreme low-light or overexposure renders the image uniformly dark or washed out.
\textit{Example}: after severe overexposure (\texttt{overexposure}, $s_3$), the distinctive facial features and fur pattern of a cat are completely washed out; the model retrieves an unrelated gallery item.

\paragraph{Semantic Drift.}
The corrupted image remains recognizable, but fine-grained discriminative cues are weakened, causing the model to retrieve a visually similar but incorrect entity from the same semantic category.
This is the dominant pattern at mild severity ($s_1$) across all four primitives, and persists at severe severity under \textsc{Transform}, whose geometric distortions preserve entity content while misaligning spatial features.
\textit{Example}: after heavy JPEG compression (\texttt{compression}, $s_3$), a query of a specific medicine box loses fine label detail; the model retrieves a different box from the same brand family.

\paragraph{Hard-Negative Confusion.}
The model retrieves a gallery item that is genuinely visually similar to the corrupted query---a same-category hard negative---rather than the exact ground-truth instance.
Even mild degradation can trigger this pattern when the gallery contains near-duplicate items.
\textit{Example}: a mild resolution reduction (\texttt{low\_resolution}, $s_1$) on a smartphone query causes the model to retrieve the same phone model in a different color, which is indistinguishable at the reduced resolution.

\paragraph{Background Bias.}
Corruption suppresses the foreground entity signal, allowing incidental background cues to dominate the embedding.
This pattern is most common under \textsc{Add} operators (e.g.\ \texttt{mosaic}, \texttt{ui\_elements}) that occlude the foreground subject while leaving the background intact.
\textit{Example}: a severe mosaic overlay on a car key query obscures the key itself; the model matches on the background pattern shared with a wrong gallery item.

% ----------------------------------------------------------
\subsubsection{Quantitative Summary}
\label{sec:appendix_failure_quant}
% ----------------------------------------------------------

\Cref{tab:failure_quant} summarises the retrieval impact at severe severity ($s_3$) for each image primitive, averaged over four models.
\textsc{Remove} is the most destructive: the correct entity is pushed down $7.84$ ranking positions on average and R@1 drops by $26.8$ percentage points.
\textsc{Transform} is the mildest, consistent with its entity-preserving nature.
Illustrative image and text perturbation cases are provided in \Cref{sec:appendix_failure_primitive_cases,sec:appendix_failure_text_cases}.

% ----------------------------------------------------------
\subsubsection{Answerability vs.\ Robustness at High Severity}
\label{sec:appendix_answerability}
% ----------------------------------------------------------

A severe corruption can degrade an image while preserving the evidence needed to recognize the target (a genuine robustness failure), or it can remove the target evidence altogether (an answerability limit that no model, however robust, could overcome). To separate these two regimes, we ran a human-recognizability check on $750$ severity-3 images (${\approx}50$ per operator, $15$ operators): each image was binary-annotated as \emph{target identifiable} or not.

\Cref{tab:answerability} shows a sharp split by operator type. \textbf{Cliff-type} operators (\textsc{mosaic}, \textsc{watermark}, \textsc{low light}, \textsc{downscale}, \textsc{cropping}) leave the target identifiable in only $10\%$ of severity-3 images, accounting for $89\%$ of their severity-3 R@1 drop; at severity 3, these operators mostly test an answerability limit rather than model robustness. \textbf{Gradual-type} operators (\textsc{compression}, \textsc{motion blur}, \textsc{rotation}) keep $95\%$ of targets identifiable at severity 3, so their drops reflect genuine retrieval robustness failures rather than unanswerable inputs.

\begin{table}[h]
\centering\small

\begin{tabular}{p{2.5cm}cc}
\toprule
\textbf{Operator type} & \makecell{\textbf{Target}\\\textbf{ID'able}} & \makecell{\textbf{\% of $s_3$}\\\textbf{drop}} \\
\midrule
Cliff-type (mosaic, watermark, low light, downscale, cropping) & 10\% & 89\% \\
\midrule
Gradual-type (compression, motion blur, rotation) & 95\% & 6\% \\
\bottomrule
\end{tabular}
\caption{Target identifiability at severity 3, by operator type ($750$ images, $15$ operators).}
\label{tab:answerability}
\end{table}

This distinction sharpens rather than overturns the paper's primary findings, which are computed over severity 1/2 or severity-averaged conditions, where target-identifiable rates remain high across both operator types. The three-level severity design lets us tell these two failure regimes apart: a smooth, monotonic decline across $s_1\!\to\!s_3$ signals a genuine robustness failure (gradual-type), while a sharp late-stage drop signals information collapse (cliff-type)---a distinction a single fixed severity level could never reveal.

\subsection{Visualisation and Case Studies}
\label{sec:appendix_visualisation}

% Visual case studies: all image and text perturbations for Case B
% (market fish query: "What kind of fish is this?")

\providecommand{\caseimgw}{0.16\linewidth}
\providecommand{\caseorigw}{0.16\linewidth}

% ----------------------------------------------------------
\subsubsection{Image Perturbation Cases}
\label{sec:appendix_failure_primitive_cases}
% ----------------------------------------------------------

\begin{figure*}[t]
  \centering\small
  \setlength{\tabcolsep}{3pt}
  \renewcommand{\arraystretch}{1.0}
  \begin{tabular}{@{} l c c c c @{}}
    \toprule
    \textbf{Operator} & \textbf{Original} & \textbf{$s_1$} & \textbf{$s_2$} & \textbf{$s_3$} \\
    \midrule
    \texttt{low\_light} &
      \includegraphics[width=\caseimgw]{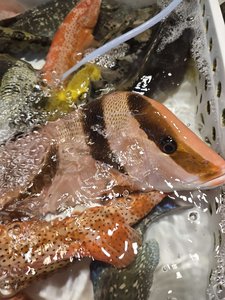} &
      \includegraphics[width=\caseimgw]{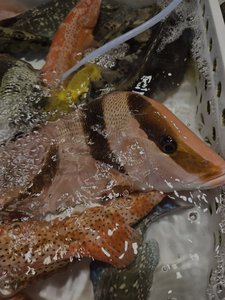} &
      \includegraphics[width=\caseimgw]{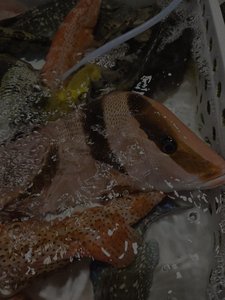} &
      \includegraphics[width=\caseimgw]{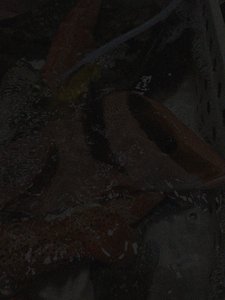} \\[2pt]
    \texttt{overexposure} &
      \includegraphics[width=\caseimgw]{query_orig/caseB_orig} &
      \includegraphics[width=\caseimgw]{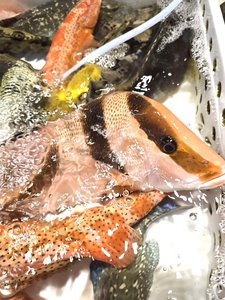} &
      \includegraphics[width=\caseimgw]{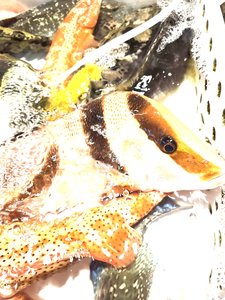} &
      \includegraphics[width=\caseimgw]{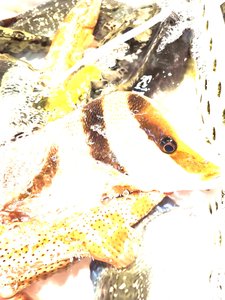} \\[2pt]
    \texttt{defocus\_blur} &
      \includegraphics[width=\caseimgw]{query_orig/caseB_orig} &
      \includegraphics[width=\caseimgw]{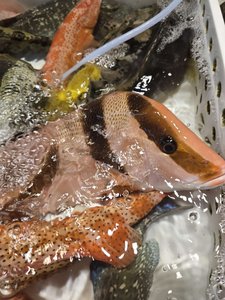} &
      \includegraphics[width=\caseimgw]{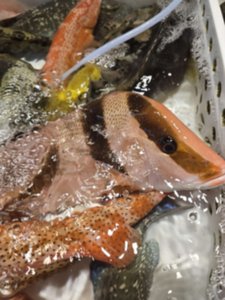} &
      \includegraphics[width=\caseimgw]{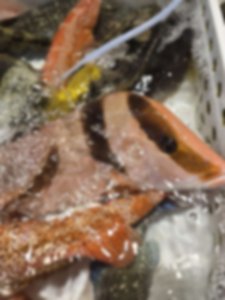} \\[2pt]
    \texttt{motion\_blur} &
      \includegraphics[width=\caseimgw]{query_orig/caseB_orig} &
      \includegraphics[width=\caseimgw]{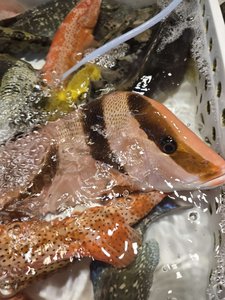} &
      \includegraphics[width=\caseimgw]{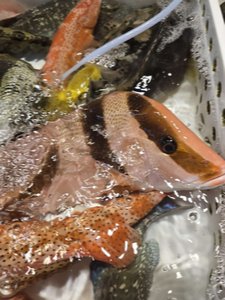} &
      \includegraphics[width=\caseimgw]{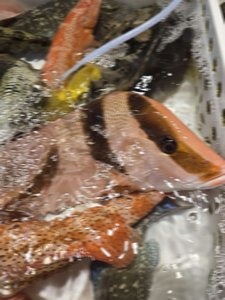} \\
    \bottomrule
  \end{tabular}
  \caption{\textbf{Degrade operators (lighting \& blur)} applied to Case~B at $s_1\!\to\!s_2\!\to\!s_3$.}
  \label{fig:case_degrade_a}
\end{figure*}

\begin{figure*}[t]
  \centering\small
  \setlength{\tabcolsep}{3pt}
  \renewcommand{\arraystretch}{1.0}
  \begin{tabular}{@{} l c c c c @{}}
    \toprule
    \textbf{Operator} & \textbf{Original} & \textbf{$s_1$} & \textbf{$s_2$} & \textbf{$s_3$} \\
    \midrule
    \texttt{compression} &
      \includegraphics[width=\caseimgw]{query_orig/caseB_orig} &
      \includegraphics[width=\caseimgw]{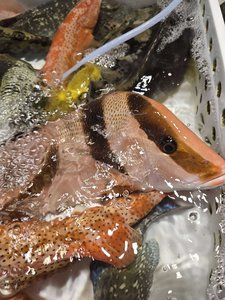} &
      \includegraphics[width=\caseimgw]{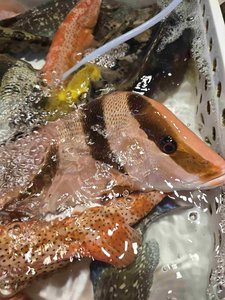} &
      \includegraphics[width=\caseimgw]{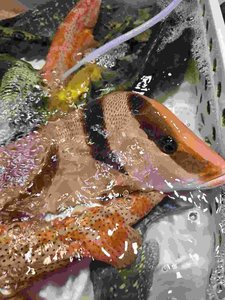} \\[2pt]
    \texttt{low\_resolution} &
      \includegraphics[width=\caseimgw]{query_orig/caseB_orig} &
      \includegraphics[width=\caseimgw]{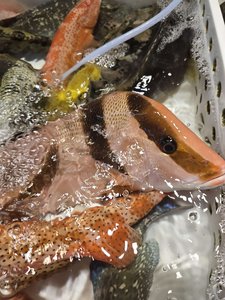} &
      \includegraphics[width=\caseimgw]{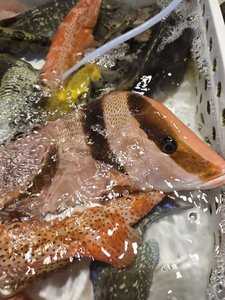} &
      \includegraphics[width=\caseimgw]{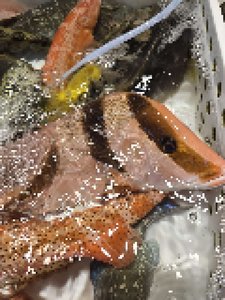} \\[2pt]
    \texttt{downscale} &
      \includegraphics[width=\caseimgw]{query_orig/caseB_orig} &
      \includegraphics[width=\caseimgw]{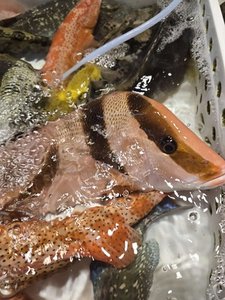} &
      \includegraphics[width=\caseimgw]{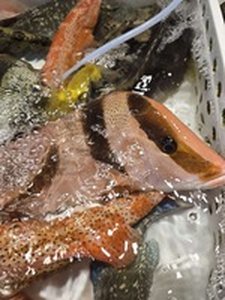} &
      \includegraphics[width=\caseimgw]{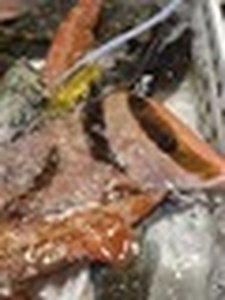} \\
    \bottomrule
  \end{tabular}
  \caption{\textbf{Degrade operators (resolution \& compression)} applied to Case~B at $s_1\!\to\!s_2\!\to\!s_3$.
    Signal-quality degradation progressively destroys discriminative visual cues.}
  \label{fig:case_degrade_b}
\end{figure*}

\begin{figure*}[t]
  \centering\small
  \setlength{\tabcolsep}{3pt}
  \renewcommand{\arraystretch}{1.0}
  \begin{tabular}{@{} l c c c c @{}}
    \toprule
    \textbf{Operator} & \textbf{Original} & \textbf{$s_1$} & \textbf{$s_2$} & \textbf{$s_3$} \\
    \midrule
    \texttt{rotation} &
      \includegraphics[width=\caseimgw]{query_orig/caseB_orig} &
      \includegraphics[width=\caseimgw]{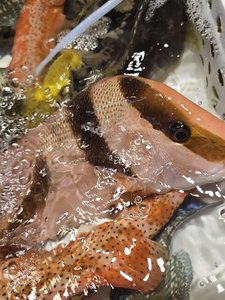} &
      \includegraphics[width=\caseimgw]{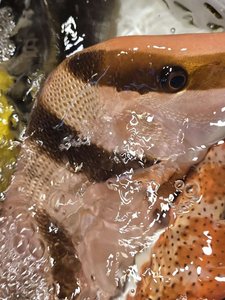} &
      \includegraphics[width=\caseimgw]{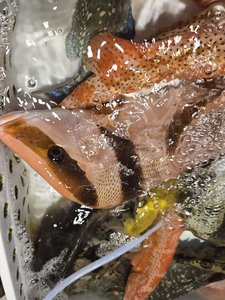} \\[2pt]
    \texttt{perspective} &
      \includegraphics[width=\caseimgw]{query_orig/caseB_orig} &
      \includegraphics[width=\caseimgw]{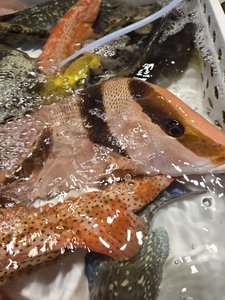} &
      \includegraphics[width=\caseimgw]{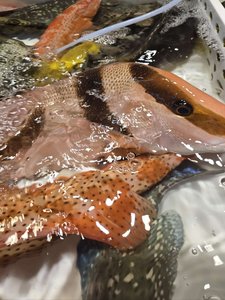} &
      \includegraphics[width=\caseimgw]{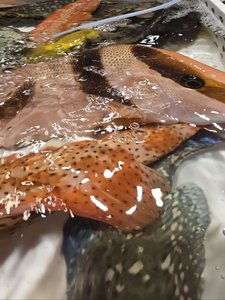} \\[2pt]
    \texttt{lens\_distortion} &
      \includegraphics[width=\caseimgw]{query_orig/caseB_orig} &
      \includegraphics[width=\caseimgw]{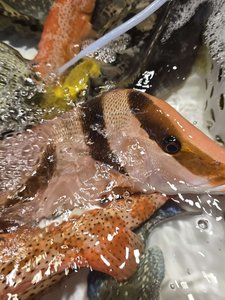} &
      \includegraphics[width=\caseimgw]{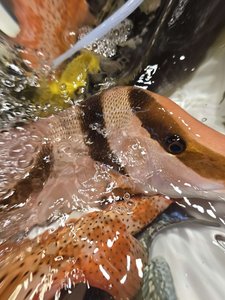} &
      \includegraphics[width=\caseimgw]{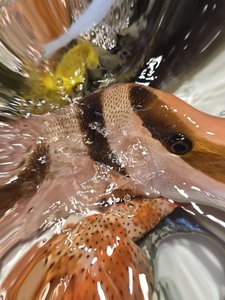} \\
    \bottomrule
  \end{tabular}
  \caption{\textbf{Transform operators} applied to Case~B at $s_1\!\to\!s_2\!\to\!s_3$.
    Geometric distortions misalign spatial features but preserve entity content,
    keeping failures in the Semantic Drift regime even at $s_3$.}
  \label{fig:case_transform}
\end{figure*}

\begin{figure*}[t]
  \centering\small
  \setlength{\tabcolsep}{3pt}
  \renewcommand{\arraystretch}{1.15}
  \begin{tabular}{@{} l c c c c @{}}
    \toprule
    \textbf{Operator} & \textbf{Original} & \textbf{$s_1$} & \textbf{$s_2$} & \textbf{$s_3$} \\
    \midrule
    \texttt{cropping} &
      \includegraphics[width=\caseimgw]{query_orig/caseB_orig} &
      \includegraphics[width=\caseimgw]{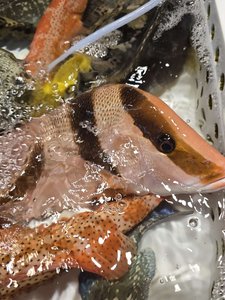} &
      \includegraphics[width=\caseimgw]{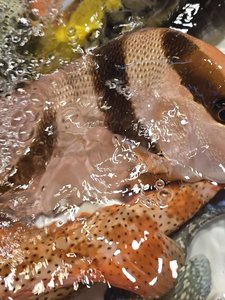} &
      \includegraphics[width=\caseimgw]{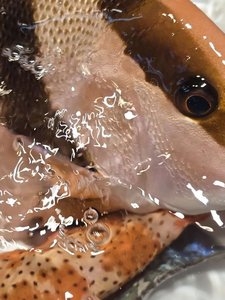} \\
    \bottomrule
  \end{tabular}
  \caption{\textbf{Remove operator} applied to Case~B at $s_1\!\to\!s_2\!\to\!s_3$.
    Aggressive cropping removes the entity from the frame, triggering Quality Degradation at $s_3$.}
  \label{fig:case_remove}
\end{figure*}

\begin{figure*}[t]
  \centering\small
  \setlength{\tabcolsep}{3pt}
  \renewcommand{\arraystretch}{1.0}
  \begin{tabular}{@{} l c c c c @{}}
    \toprule
    \textbf{Operator} & \textbf{Original} & \textbf{$s_1$} & \textbf{$s_2$} & \textbf{$s_3$} \\
    \midrule
    \texttt{mosaic} &
      \includegraphics[width=\caseimgw]{query_orig/caseB_orig} &
      \includegraphics[width=\caseimgw]{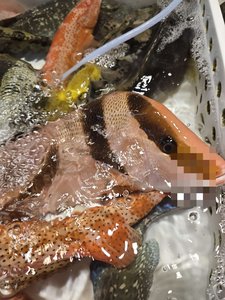} &
      \includegraphics[width=\caseimgw]{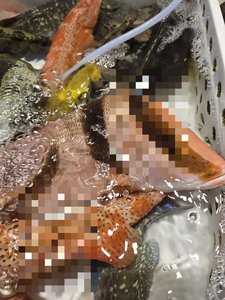} &
      \includegraphics[width=\caseimgw]{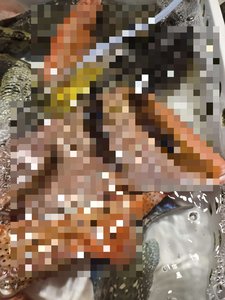} \\[2pt]
    \texttt{watermark} &
      \includegraphics[width=\caseimgw]{query_orig/caseB_orig} &
      \includegraphics[width=\caseimgw]{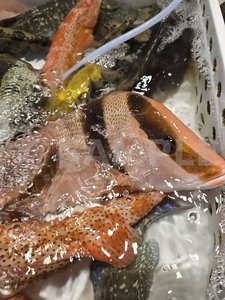} &
      \includegraphics[width=\caseimgw]{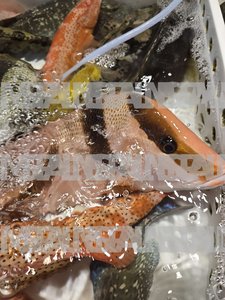} &
      \includegraphics[width=\caseimgw]{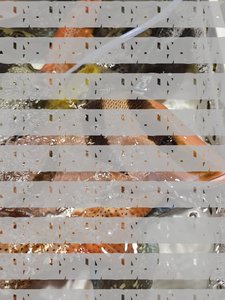} \\[2pt]
    \texttt{scribble} &
      \includegraphics[width=\caseimgw]{query_orig/caseB_orig} &
      \includegraphics[width=\caseimgw]{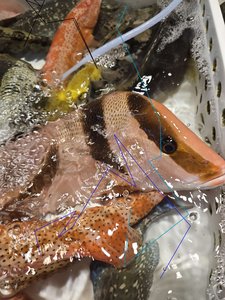} &
      \includegraphics[width=\caseimgw]{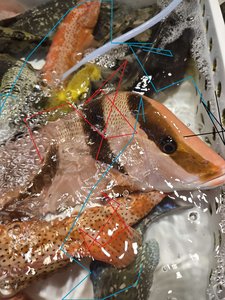} &
      \includegraphics[width=\caseimgw]{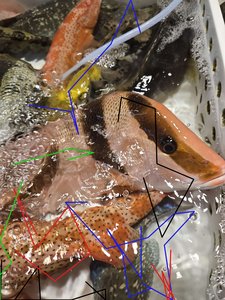} \\[2pt]
    \texttt{ui\_elements} &
      \includegraphics[width=\caseimgw]{query_orig/caseB_orig} &
      \includegraphics[width=\caseimgw]{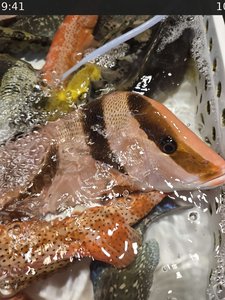} &
      \includegraphics[width=\caseimgw]{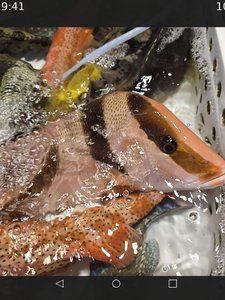} &
      \includegraphics[width=\caseimgw]{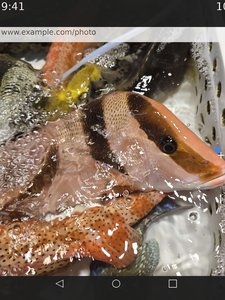} \\
    \bottomrule
  \end{tabular}
  \caption{\textbf{Add operators} applied to Case~B at $s_1\!\to\!s_2\!\to\!s_3$.
    Overlaid elements occlude local regions without globally destroying the scene;
    at $s_3$, coverage is dense enough to suppress the foreground entity and trigger Background Bias.}
  \label{fig:case_add}
\end{figure*}

\Cref{fig:case_degrade_a,fig:case_degrade_b,fig:case_transform,fig:case_remove,fig:case_add} show Case~B (market fish query) under all 15 image operators at three severity levels, grouped by primitive.
As severity rises, entity-level cues are progressively obscured---consistent with the failure patterns described in \Cref{sec:appendix_failure_patterns}.

% ---- Degrade part 1 (lighting + blur, 4 operators) ----

% ---- Degrade part 2 (resolution/compression, 3 operators) ----

% ---- Transform (3 operators) ----

% ---- Remove (1 operator) ----

% ---- Add (4 operators) ----

% ----------------------------------------------------------
\subsubsection{Text Perturbation Cases}
\label{sec:appendix_failure_text_cases}
% ----------------------------------------------------------

\Cref{fig:appendix_failure_text_operators} lists all eight text operators applied to the same query
(Case~B, entity~=~\texttt{fish}).
Character-level edits attack the entity token directly, degrading the retrieval signal at the word level.
Sentence-level \texttt{sent\_replace} removes the category keyword entirely, steering retrieval toward generic visual matches and triggering Semantic Drift.

\begin{figure*}[t]
  \centering\small
  \setlength{\tabcolsep}{5pt}
  \renewcommand{\arraystretch}{1.35}
  \begin{tabular}{@{} l l >{\raggedright\arraybackslash}p{0.62\linewidth} @{}}
    \toprule
    \textbf{Level} & \textbf{Operator} & \textbf{Perturbed query (corruption in \textbf{bold})} \\
    \midrule
    \multicolumn{3}{@{}l}{\textit{Original: What kind of fish is this?}} \\
    \midrule
    \multirow{4}{*}{\makecell[l]{Char.}}
      & \texttt{char\_add}    & What kind of \textbf{fiish} is this? \\
      & \texttt{char\_delete} & What kind of \textbf{fsh} is this? \\
      & \texttt{char\_change} & What kind of \textbf{gish} is this? \\
      & \texttt{char\_swap}   & What kind of \textbf{ifsh} is this? \\
    \midrule
    \multirow{2}{*}{\makecell[l]{Word}}
      & \texttt{word\_repeat} & What kind of \textbf{fish fish} is this? \\
      & \texttt{word\_swap}   & What kind of is \textbf{fish} this? \\
    \midrule
    \multirow{2}{*}{\makecell[l]{Sent.}}
      & \texttt{sent\_add}    & What kind of fish is this? \textbf{Traffic was bad today.} \\
      & \texttt{sent\_replace}& \textbf{What is this?} \\
    \bottomrule
  \end{tabular}
  \caption{%
    \textbf{Text perturbation cases per operator} (Case~B).
    All operators target the entity word \texttt{fish}; corrupted spans are in bold.
    Character-level noise degrades the entity token directly;
    word-level noise preserves the entity string but perturbs syntax;
    \texttt{sent\_replace} removes the category keyword and is the strongest Semantic Drift trigger.}
  \label{fig:appendix_failure_text_operators}
\end{figure*}

% ---- E: Discussion & Ethics ----
% \section{Extended Discussion and Ethics}
\section{LLM Use Statement and Ethics}
% \subsection{Future Work}
% \label{sec:appendix_limitations}
% % ============================================================
% \subsubsection{Extended Limitations}
% \paragraph{Synthetic vs.\ in-the-wild corruptions.}
% \benchname{}'s corruptions are programmatically generated for factorisability; the joint distribution of corruptions in real uploads may differ from our marginal distributions. We argue elsewhere that the primitive-level taxonomy spans the operator space, but explicitly acknowledge this caveat.

% \paragraph{Scale.} $1{,}145$ queries / $9{,}085$ gallery items is modest by web-scale standards; we deliberately trade scale for dense ground-truth annotation ($4.77$ GT/query) and tight hard-negative coverage.

% \paragraph{MOOR preconditions.} MOOR is most effective when modal signal quality varies across queries; on queries where both modalities are uniformly informative or uniformly uninformative, MOOR reverts to fixed-fusion behavior and provides no advantage by construction.

% ============================================================
% \subsubsection{Future Work}
% \begin{itemize}[leftmargin=1.4em,topsep=2pt,itemsep=2pt]
%     \item Relaxing primitive-frequency assumptions in real-world corruption mixtures.
%     \item Extension to multi-turn agentic queries (the second hop
%     after grounding).
%     \item Training-aware variants of MOOR (lightweight
%     learned gating with the same closed-form initialisation).
% \end{itemize}

% ============================================================
% \subsection{LLM Use Statement and Ethics}
\label{sec:appendix_ethics}
% ============================================================
\subsubsection{Use of Large Language Models}
LLMs were used in three clearly bounded roles in this work.
(i)~\emph{Query generation} (Appendix~\ref{sec:appendix_query}, Gemini-3-Flash): queries are generated automatically, with $100$ randomly sampled queries verified by human spot-check (Stage~5, all pass), following the broader paradigm of LLM-agent-driven data synthesis in data-limited settings~\citep{du2025graphmaster}.
(ii)~\emph{Semantic quality checking} (Appendix~\ref{sec:appendix_query}, GPT-5.4-mini in read-only \texttt{PASS}/\texttt{FAIL} mode, no rewriting): the model verifies entity-tag coarseness and absence of answer leakage.
(iii)~\emph{Gallery caption generation} (Appendix~\ref{sec:appendix_gallery}, Qwen-VL-based captioning model): entity-centric captions are generated automatically for each gallery image. LLMs were \emph{not} used to grade retrieval results or to construct ground-truth relevance labels.

\subsubsection{Use of AI Assistants in Research and Writing}
AI writing assistants were used for grammar and language polishing during manuscript preparation. All technical content, experimental results, and conclusions were reviewed and verified by the authors.

% \subsubsection{Ethical Considerations (Extended)}
% All gallery images are sourced from publicly available web data
% under a copyright filter. Personally identifying information in
% query text is removed at Stage 1 of the query pipeline
% (Appendix~\ref{sec:appendix_query}). Annotators were paid above the local
% minimum wage and informed of the research nature of the task;
% annotation guidelines explicitly exclude content involving
% violence, sexual material, or political sensitivity.
% MOOR, as a training-free fusion adapter, inherits the
% biases of whichever underlying encoder it is applied to; we
% recommend users evaluate MOOR jointly with their
% encoder of choice on bias-relevant evaluations before
% deployment.

\end{document}